\documentclass{egpubl}
\usepackage{egsr2021}
\usepackage[table]{xcolor}

\SpecialIssuePaper         

\usepackage[T1]{fontenc}
\usepackage{dfadobe}  
\usepackage{amsmath}  

\BibtexOrBiblatex
\electronicVersion
\PrintedOrElectronic
\ifpdf \usepackage[pdftex]{graphicx} \pdfcompresslevel=9
\else \usepackage[dvips]{graphicx} \fi

\graphicspath{{./fig}{./images}}
\usepackage{egweblnk} 
\usepackage{booktabs} 

\newcommand{\CR}  [1] { #1}

\def\VSL{V_{\mathrm{sel}}}

\DeclareMathOperator*{\argmax}{argmax}

\title{Point-Based Neural Rendering with Per-View Optimization}

\author[G. Kopanas, J. Philip, T. Leimk\"uhler \& G. Drettakis ]
{\parbox{\textwidth}{\centering Georgios Kopanas$^{1}$, Julien Philip$^{1,2}$, Thomas Leimk\"uhler$^{1}$, and George Drettakis$^{1}$  } \\
	{\parbox{\textwidth}{\centering $^1$ Inria, Université Côté d'Azur\\
			$^2$ Adobe Research
		}
	}
}

\begin{document}


\teaser{
\vspace{-0.75cm}
		\centering
		\includegraphics[width=\linewidth]{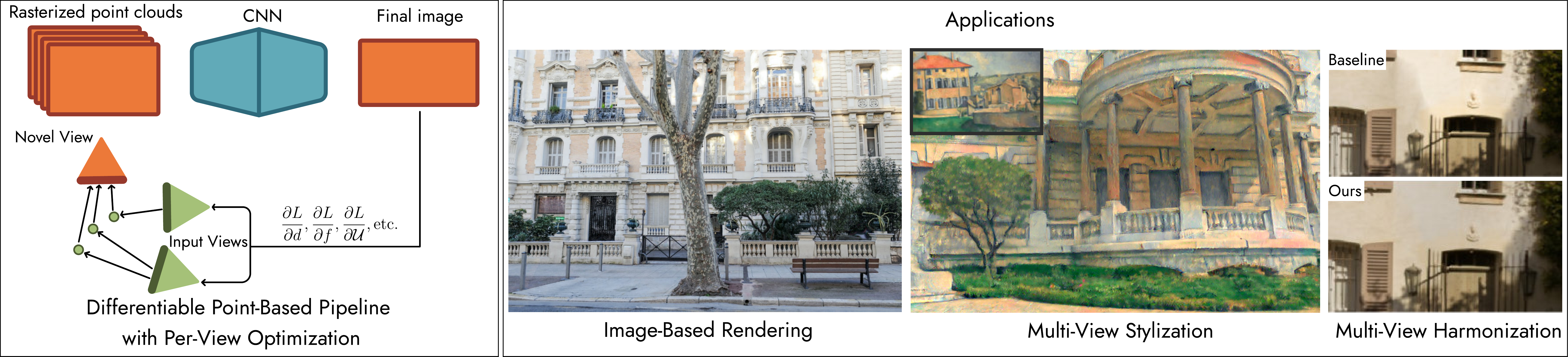}
		\caption{\CR{Left: Our differentiable point-based pipeline allows optimization of attributes such as reprojected features or depth in each input view. 
Right: We illustrate three applications of our pipeline. From left to right: High-quality image-based rendering; multi-view stylization which can be used to achieve the illusion of navigating inside a painting with temporal and multi-view consistency;  multi-view harmonization that can be used to improve image-based rendering (notice the removal of discontinuities due to exposure).}}
}	

\maketitle

\begin{abstract}
There has recently been great interest in neural rendering methods. Some approaches use 3D geometry reconstructed with Multi-View Stereo (MVS) but cannot recover from the errors of this process, while others directly learn a volumetric neural representation, but suffer from expensive training and inference.
We introduce a general approach that is initialized with MVS, but allows further optimization of scene properties in the space of input views, including depth and reprojected features, resulting in improved novel-view synthesis. A key element of our approach is our new differentiable point-based pipeline, based on bi-directional Elliptical Weighted Average splatting, a probabilistic depth test and effective camera selection. We use these elements together in our neural renderer, that outperforms all previous methods both in quality and speed in almost all scenes we tested. Our pipeline can be applied to multi-view harmonization and stylization in addition to novel-view synthesis.

\end{abstract}



\section{Introduction}

Multi-view capture of real-world scenes has become a very popular way of creating digital content, e.g., as reconstructed 3D textured meshes or as input to Image-Based Rendering (IBR). 
To improve unreliably reconstructed depth, and thus image quality, traditional IBR algorithms use \emph{per-input-view} information ~\cite{chaurasia2013depth,hedman2018deep}. 
Recently, neural rendering methods have been proposed to navigate in such captured scenes, some directly using traditional 3D reconstruction to guide rendering~\cite{riegler2020free,hedman2018deep} -- offering stability and robustness in novel view synthesis -- while others use neural networks to learn a neural radiance field representation of the underlying geometry~\cite{mildenhall2020nerf,srinivasan2020nerv}. The former suffer from the fact that the rendering cannot recover from errors in the reconstructed geometry, while the latter have been shown to be less stable than methods that use explicit geometry~\cite{zhang2020nerfplusplus,riegler2020free}. In the \emph{wide-baseline, free-viewpoint} scenarios we target, such neural radiance field methods often present artifacts, partly because they do not optimize \emph{per input view} parameters.
We present a new neural rendering pipeline that is initialized with standard 3D reconstruction, but allows subsequent \emph{per-input view} optimization of attributes such as reprojected latent features and depth, offering benefits of both approaches. Our per-view approach optimizes features and depth together, finding a good compromise between correcting depth and blur artifacts in novel-view synthesis. 

Our differentiable multi-view pipeline is based on point-based splatting, together with an efficient, high-quality neural renderer, that optimizes \emph{per-input-view} attributes.
We first introduce a fast algorithm to select the cameras from which we actually reproject points, significantly improving the speed/quality tradeoff.
Our fully differentiable algorithm then performs soft, alpha-blended splatting of points reprojected from the input images into novel views, using bi-directional Elliptical Weighted Average (EWA) filtering.

When reprojecting images from input views with inaccurate geometry, it is always hard to determine which view contains the correct information. Point splats lack the connectivity of meshes, but also make it easier to formulate a probabilistic approach to this depth testing problem, which is necessary because of the ``soft'' point splatting approach.
We introduce such an approach, that uses the distribution of depths from each reprojected input view to perform soft depth resolution. 

Taken together, all the elements of this high-quality differentiable point-splatting framework allow us to introduce a temporally consistent neural renderer that we use to first optimize per-input-view parameters such as reprojected features or depth for a given scene. For each scene we can then use our neural renderer for several multi-view imaging tasks, such as free-viewpoint IBR, multi-view harmonization and multi-view consistent image stylization.
In summary, our contributions are:
\begin{itemize}
\item A point-based multi-view imaging framework that allows per-view optimization. To do this, we introduce three new components: 1) a camera selection algorithm, for efficient rendering and training; 2) a differentiable point-splatting method with bi-directional EWA filtering and 3) a probabilistic per-view depth testing algorithm.
\item A neural renderer based on our framework that allows optimization of per-view parameters and high-quality, temporally coherent rendering.
\end{itemize}
Our results show how our framework can be used for free-viewpoint IBR, notably with improved IBR quality for difficult cases such as vegetation or thin structures compared to previous work, multi-view color harmonization and multi-view consistent image stylization. 
Our neural renderer outperforms all previous view synthesis methods we tested both in quality and speed in almost all scenes presented, in terms of quantitative measures and visual quality.

\section{Related Work}
\label{sec:related}

Our method is related to Image-Based Rendering (IBR), neural and point-based rendering, and in addition to IBR, we show applications of our approach to multi-view image harmonization and stylization. Each of these is a vast field on its own; we discuss only the most closely related representative work for each case.

\subsection{Traditional IBR}
\label{sec:rel-ibr}

Early methods in IBR~\cite{mcmillan1995plenoptic,levoy1996light} demonstrated the power of blending a set of images to allow various effects such as viewpoint changes or depth-of-field, without the need for manually building and rendering a full 3D scene. Using even approximate 3D was shown to be beneficial~\cite{gortler1996lumigraph}, and led to IBR methods that allow free-viewpoint navigation~\cite{buehler2001unstructured,heigl1999plenoptic}, by calculating \emph{blend weights} to mix input images reprojected into a novel view using approximate geometry.
The advent of powerful Structure from Motion (SfM)~\cite{snavely2006photo} and Multi-View Stereo (MVS)~\cite{goesele2007multi} which allow the automatic reconstruction of an approximate 3D geometry \emph{proxy}, made IBR a much more attractive solution for capture of multi-view datasets of real-world scenes, and subsequent free-viewpoint navigation with realistic rendering.

More recently, a new class of algorithms has been developed that exploit per-input-view -- or simply \emph{per-view} -- data to improve rendering quality. These methods show that it is easier to accurately represent depth for a single view, even if that representation is not necessarily multi-view consistent. 
This can be achieved using a superpixel decomposition \cite{chaurasia2013depth} of each input view to preserve depth edges and synthesize missing depth, or per-view meshes to improve local estimation of depth~\cite{HRDB16}. Both methods greatly improve IBR quality in the presence of inaccurate depth. We also use per-view information, but in a \emph{differentiable} pipeline, suitable for deep learning.

Specific IBR solutions have been developed for hard cases such as vegetation or thin structures~\cite{thonat2018thin}, sometimes requiring manual intervention. 
We improve rendering quality for such cases compared to other automatic methods, without user input.

Camera selection is a critical component of many IBR and neural algorithms (see below). For large datasets, it is impossible to consider all input views for projection in a novel view being synthesized due to GPU memory limitations. 
Simple solutions include the use of Unstructured Lumigraph (ULR)-style weights~\cite{buehler2001unstructured} to choose a small number of cameras, e.g., for per-pixel ULR~\cite{sibr2020}, or superpixel-based IBR~\cite{chaurasia2013depth}. 
A voxel grid in which input view meshes are ``sliced''~\cite{HRDB16} can also be used, providing an approximation valid up to the grid resolution. Riegler and Koltun~\shortcite{riegler2020free} use reprojected depth maps to maximize the overlap with a target view.
Our approach builds on the K-Maximum coverage algorithm~\cite{hochbaum1998analysis} to provide fast and accurate camera selection.

\subsection{Neural Rendering}

Recent years have seen the use of deep learning to improve IBR, and as a completely new way to perform rendering overall, resulting in the rapidly expanding field of \emph{neural rendering}~\cite{Tewari2020NeuralSTAR}. 

One class of neural rendering solutions builds on MVS proxy geometry. Hedman et al.~\cite{hedman2018deep} refine per-view geometry and train a neural network to learn the \emph{blend weights} for novel view synthesis, while Riegler and Koltun~\shortcite{riegler2020free} use a recurrent architecture to project features using MVS geometry using per-view depth maps; in follow-up work~\cite{riegler2020stable} they use the MVS mesh to project features in a stable manner, resolving some of the issues of the original method.
While we also use MVS geometry for reprojection, a fundamental limitation of these methods is that the fixed, often erroneous reconstructed geometry hinders image quality in novel view synthesis, while our approach optimizes per-view features and depth using backpropagation, improving the overall result (see Sec.~\ref{sec:ibr} for comparisons). 

Point-based rendering~\cite{gross2011point} has also been used recently for neural rendering~\cite{aliev2019neural,wiles2020synsin}. This representation is well studied in graphics and the advantages and disadvantages are well understood \cite{gross2011point}. Points come naturally from RGBD sensors, as well as SfM and early steps of many MVS algorithms; they also allow easy reprojection of learned features. In the work of Aliev et~al.~\shortcite{aliev2019neural} point positions are assumed to be correct, and the neural renderer learns to correct visual artifacts. This process is successful if the artifacts are present in the training set. In contrast, Wiles et al.~\shortcite{wiles2020synsin} use points to leverage a differentiable renderer to compute gradients that in turn drive a model to estimate depth maps from a single image subsequently used for view synthesis.
In the different context of geometric modelling, Yifan et~al.~\shortcite{Yifan:DSS:2019} present forward EWA point splatting in a differentiable pipeline. We extend this approach to be bi-directional, allowing optimization of per-view features in the 2D space of the input images.
While we use methodological components from these two methods, our multi-view datasets and per-view reprojection context is very different.

In concurrent work, Lassner and Zollhoeffer \cite{Lassner_pulsar} develop a fast differentiable point-based pipeline, but this method would need to be extended to allow alpha-blending and the bi-directional component of our pipeline to be used in our context.

Instead of using MVS, several methods start simply with the calibrated cameras from SfM and try and learn geometric representations. Initial methods started with very small numbers of views~\cite{choi2019extreme}, plane sweep~\cite{flynn2016deepstereo} or multi-plane images (MPIs)~\cite{flynn2019deepview}, with similarities in spirit to traditional optimization of per-view depth~\cite{penner2017soft}.
Mildenhall et al.~\shortcite{mildenhall2019local} learn to reduce opacity of individual MPIs in uncertain areas, allowing several of them to be blended, thus supporting somewhat larger camera movement.
These methods are limited to small-baseline capture such as light fields, in contrast to our wide-baseline scenario.

Learned volumetric representations have been recently proposed with very promising results either on video~\cite{Lombardi2019} or isolated objects~\cite{sitzmann2019deepvoxels}. It is unclear how well these approaches would fare in our target sparse wide-baseline capture of large scenes. 
Similarly, recent video-based neural rendering solutions~\cite{Attal2020,broxton2020immersive} introduce ideas with impressive results, but typically require much involved capture setups.

Learned neural radiance (``NeRF'') representations (e.g., \cite{mildenhall2020nerf,martin2020nerf}) learn density and color in an outgoing direction. This is a very powerful representation, as testified by the follow-up work for various other applications~(e.g., \cite{park2020deformable,srinivasan2020nerv,liu2020neural}). 
NeRF methods present artifacts in our wide-baseline scenes, especially for vegetation (see Sec.~\ref{sec:eval}). Even though recent, unpublished manuscripts (e.g.,~\cite{yu2021plenoctrees,hedman2021baking,reiser2021kilonerf}) present interactive NeRF solutions for small rendering resolutions, they all involve some degradation in quality, which would possibly be more pronounced in our scenes.

\subsection{Harmonization and Stylization}

In addition to IBR, we show two other applications of our differentiable multi-view imaging approach: image stylization and image harmonization.

Image stylization is an active field in Computer Graphics, and many methods have been developed to transform a photograph to look like a painting or drawing~\cite{kyprianidis2012state}. Deep learning methods have been developed for this application with widespread success (e.g.,~\cite{gatys2016image,isola2017image,deepimageprior} to name a few).
Recently, solutions allow video stylization using deep learning \cite{ruder2018artistic,Texler20-SIG}; however they usually require reliable optical flow to work well, which is not available for our wide-baseline inputs.
Image harmonization is useful in many different contexts e.g., panoramic image stitching, multi-view texturing etc.
Unknown vignetting, exposure and camera response functions can be removed in individual images~\cite{kim2008robust,goldman2010vignette}, but adapting these solutions to multi-view can be challenging.
Zhang et~al.~\shortcite{zhang2016emptying} address the challenge of recovering per-image exposure and radiance at each vertex to radiometrically calibrate the scene.
Image harmonization can also be performed using color transfer using image statistics for single~\cite{hacohen2011non} or multiple images~\cite{hacohen2013optimizing}.
These methods tend to require more overlap between images than we have in our wide baseline datasets.
Finally, Huang et~al.~\shortcite{huang20173dlite} optimize a parametric curve per image of a multi-view dataset, and harmonize image intensities.

Image harmonization can be used to improve the quality of a textured mesh produced from an MVS reconstruction. Previous methods often optimize view selection and apply image editing operators to remove seams~\cite{zhou2014color,waechter2014let}, although they also eliminate all view-dependent effects which is undesirable in IBR.

\section{Differentiable Multi-View Rendering with Per-View Optimization}

Our framework allows us to optimize \emph{per-view attributes}, such as projected features, depth or others as required by different applications (see Sec.~\ref{sec:applis}), in the context of neural rendering. This is in contrast to previous methods that focus only on optimization for the synthesis of a novel view or the global geometry.

To allow this, we need to select the cameras that will be reprojected; we first present and effective and robust camera selection algorithm. 
To allow optimization of attributes in each selected \emph{input view}, we introduce a bi-directional point splatting approach from input views to a common novel view space, and back again to the input views so that the properties to be optimized are differentiable, while providing high-quality rendering using efficient filtered splatting. 
Finally, we introduce a probabilistic depth-test to resolve visibility between projected input views.

\subsection{Camera Selection}
\label{sec:cam-sel}

Selecting a subset of cameras to use when synthesizing a novel view is an important component of many IBR and neural rendering algorithms (see Sec.~\ref{sec:rel-ibr}, ~\ref{sec:ibr}). Our goal is to develop a solution that is effective in choosing a good set of cameras by considering pixel coverage.
\CR{Previous work \cite{riegler2020free}} ranks and chooses input views based on how many novel-view pixels they cover. However, this approach does not necessarily lead to an optimal coverage of the entire novel view.

\begin{figure}[!h]
	\includegraphics[width=\linewidth]{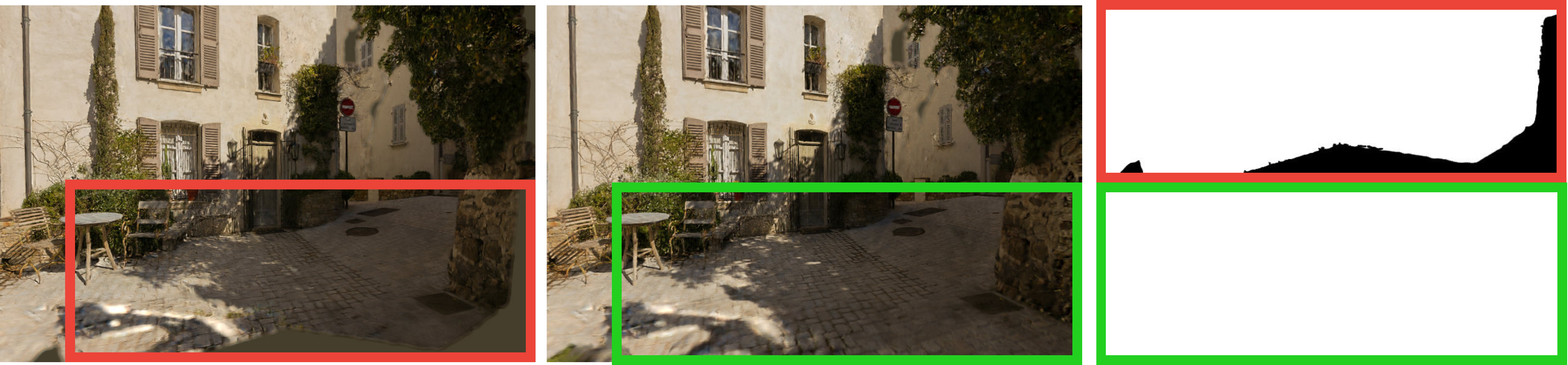}
	\caption{
		\label{fig:camera_selection}
With a limited budget of cameras, i.e., $N=4$, we see the importance of camera selection even in simple scenes like Ponche. Left is the camera selection method of \cite{riegler2020free}. In the middle is ours and on the right we show a mask with black for missing content. In the top right (red), we see that  the first method clearly fails to provide information for the whole frame, while ours (bottom, green) with the same number of cameras covers the frame.
	}

\end{figure}

Our solution is inspired by the Maximum $K$-Coverage problem from set theory to approximate the solution \cite{hochbaum1998analysis}; in this problem we are given several possibly overlapping sets and a number $K$. The goal is to select $K$ of these sets so that the union of the selected sets has maximal size. In our case the input views are the sets, and their elements are (reprojected) visible pixels.

We first estimate a downscaled score map $S_i$ in the novel-view space for each input camera. We use the proxy mesh estimated from the MVS algorithm as a basis for our re-projection, including a depth test for occlusions using this mesh. We use the mesh instead of point splatting for speed. Our algorithms chooses a sub-set of cameras that when combined maximizes the maximum score of the surface projected to the novel view. We do this by solving the following optimization problem:
\[ 
\argmax_{\VSL} \sum_{p\in P}\max_{v\in \VSL}S_i(p) \textrm{  with   } |\VSL|=k 
\]
\noindent
where $P$ are the pixels in the novel view, $\VSL$ target set of views and the visibility score $S_i(p)$ is an indicator function returning 1 if $p$ is visible in view $i$.
This algorithm has a lower bound over the ratio of the score  compared to the best solution $\frac{\mathrm{greedy}}{\mathrm{best}}<1-1/e$~\cite{hochbaum1998analysis}.

In the degenerate case where the novel view is the same as an input view, this algorithm is ambiguous. That is because the first input view will cover all the pixels in the set, hence all other views will have an equal score of zero. We can detect this case in its most general form when $\frac{V_{k+1}}{V_{k}}<\epsilon$ where $V_{k}$ is the total score after selecting the $k$-th candidate view. In this case we switch the candidate selection criterion and pick the view with the absolute maximum $\sum_{p\in P}S_i(p)$. 

The visibility function can be replaced trivially to incorporate a per-pixel score that is non-binary. We choose the ratio between the distance of the point visible from the input camera and the novel view. This is a fast approximation of the density of the points in the novel view. 
Given $M$ input cameras the complexity of the algorithm is $O(N\cdot M)$ which is tractable for interactive rendering. 

We compare our camera selection in Fig.~\ref{fig:camera_selection} to that of Riegler et~al.~\cite{riegler2020free}; this approach misses important content on the lower right corner (shown in black on the red box -- upper right).

\subsection{Bi-Directional Differentiable Point Cloud Rasterization}
\label{Differentiable-Point-Cloud-Rasterization}

As discussed in Sec.~\ref{sec:related}, using point-based rendering allows us to directly benefit from the geometry provided by MVS/SfM and from the advantages of a fully differentiable pipeline~\cite{aliev2019neural,wiles2020synsin,Yifan:DSS:2019}.
We want to allow optimization of \emph{per-view} attributes, such as latent features, depth etc. in each \emph{input image}. 

\begin{figure}[!h]
  \centering
  \includegraphics[width=\linewidth]{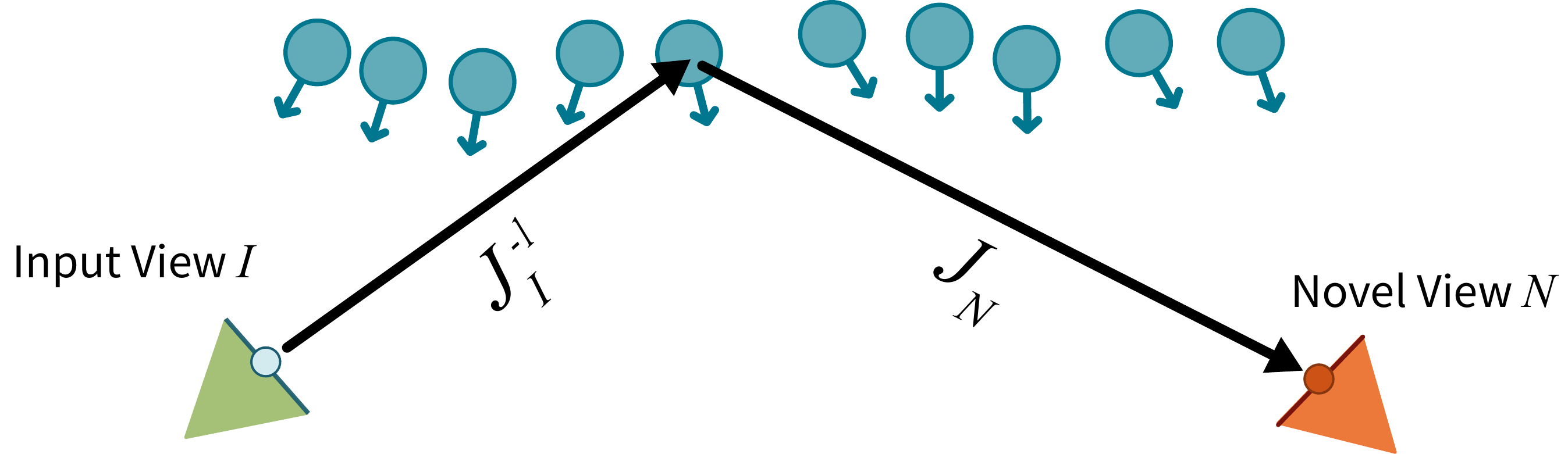}
  \caption{\label{fig:bidir-ewa}
Points (cyan, arrows denote normals) are lifted from the 2D input image (green), and then reprojected to the novel view (orange). This requires the use of a \emph{bi-directional} EWA filtering method, employing two Jacobians.}
\vspace{-.5cm}
\end{figure}

\CR{Traditional rasterization suffers from discontinuities during splatting and z-buffering}. \CR{Overcoming this and incorporating rasterization techniques in a differentiable pipeline} requires a soft splatting approach of input view pixels into the novel view; we adapt Elliptical Weighted Averaging (EWA) \cite{heckbert1989fundamentals,ren2002object,Yifan:DSS:2019} for this task.
EWA has been used for the projection of 3D points to 2D images where Ren et al.~\shortcite{ren2002object} (eq. 9-12) provide the required procedure to compute the Jacobian of the transformation from a 3D point to a regular 2D grid.
In our approach, we need to account for the combined stretch induced by two transformations:
First the lifting of a sample from a regular 2D grid to 3D, and then the projection of this point back to the novel view.
This naturally gives rise to a bi-directional procedure, where we first use the inverse Jacobian of the transformation from the input view to 3D, followed by the forward Jacobian to move from 3D to the novel view.
Since the Jacobian of an inverse transformation is the inverse of the Jacobian, we can use the method of Ren et al. in both cases and just invert the first Jacobian (Fig.~\ref{fig:bidir-ewa}).
The net result of this process is an anisotropic 2D Gaussian $\mathbf{G}$ associated with each pixel to reproject (see~\cite{Yifan:DSS:2019} for details).
This bi-directional approach allows us to extend the differentiable point-splatting method for per-view optimization. 

{\bfseries Alpha Blending.}
We use alpha-blending to avoid discontinuities at point boundaries and when points overlap~\cite{wiles2020synsin}.
We find the opacity $\alpha_{n,i}^{(v)}$ of pixel $i$ in input-view $n$ by evaluating $\mathbf{G}$ for novel-view pixel $v$.
Specifically, alpha blending is performed using front-to-back compositing:

\begin{equation}
\label{eq:front-to-back}
	c_n^{(v)} = \sum_{i \in \mathcal{N}_n^{(v)}}
	\left(c_{n,i}\alpha_{n,i}^{(v)}
	\prod_{j=1}^{i-1}(1-\alpha_{n,j}^{(v)})\right)
\end{equation}
where $c$ is a pixel attribute (e.g., color) and $\mathcal{N}_n^{(v)}$ the depth-ordered set of splats from input view $n$ that overlap novel-view pixel $v$.
We will omit the superscript $v$ in the remainder of this section to aid readability.
The alpha-blending step requires a specific gradient computation, extending Yifan et al.~\cite{Yifan:DSS:2019}:
\begin{equation*}
	\frac{\partial c_n}{\partial c_{n,i}}
	=
	\alpha_{n,i}
	\prod_{j=1}^{i-1}(1-\alpha_{n,j})
\end{equation*}
 \begin{equation*}
	\frac{\partial c_n}{\partial \alpha_{n,i}} 
	= 
	c_{n,i}
	\prod_{j=1}^{i-1}(1-\alpha_{n,j}) 
	- 
	\sum_{l=k+1}^{|\mathcal{N}_n|}
	c_{n,l} \alpha_{n,l}
	\prod_{\substack{j=1\\j\neq i}}^{l-1}
	(1-\alpha_{n,j})
\end{equation*}

For computational efficiency, we limit the support of $\mathbf{G}$ based on two criteria:
First, we stop considering more input pixels if the accumulated alpha reaches one grayscale level, since in this case any subsequent points will not alter the color.
Second, we consider the spatial decay of $\mathbf{G}$, by computing the eigenvalues of its covariance matrix.
We remove outliers (top 3\% of points with the highest variance), and choose the highest remaining variance $\sigma_\mathrm{max}$. We define the cut-off radius $r$ to be 99\% of the energy of the corresponding Gaussian: $r=\lceil 3\sigma_\mathrm{max} \rceil$, i.e., the standard ``three $\sigma$'' cutoff.
In general, we only maintain the front-most $k_d=150$ splat contributions per novel-view pixel to keep a constant memory footprint.

The EWA filter process correctly accounts for distance and orientation during our bi-directional splatting. However, there is inherent \emph{uncertainty} in the position of each input-view pixel arising from the MVS reconstruction. We model this uncertainty $\mathcal{U}$ as a multiplicative factor on the covariance matrix of $\mathbf{G}$, \CR{effectively controlling the fuzziness of the individual splats.}
Thanks to our differentiable framework, we can include these per-input view pixel uncertainties to the set of optimizable attributes. \CR{We found this to be a simple yet expressive way to let the pipeline handle and balance geometric uncertainty arising from reprojection errors.}
We initialize uncertainty with $\mathcal{U}=0.5$; we initially used a value of one (corresponding to standard EWA), but our tighter kernel improves image sharpness.
During optimization (Sec.~\ref{sec:optimization}) uncertainty is refined and adapts to reconstruction and rendering errors. 

\begin{figure}[!h]
	\includegraphics[width=\linewidth]{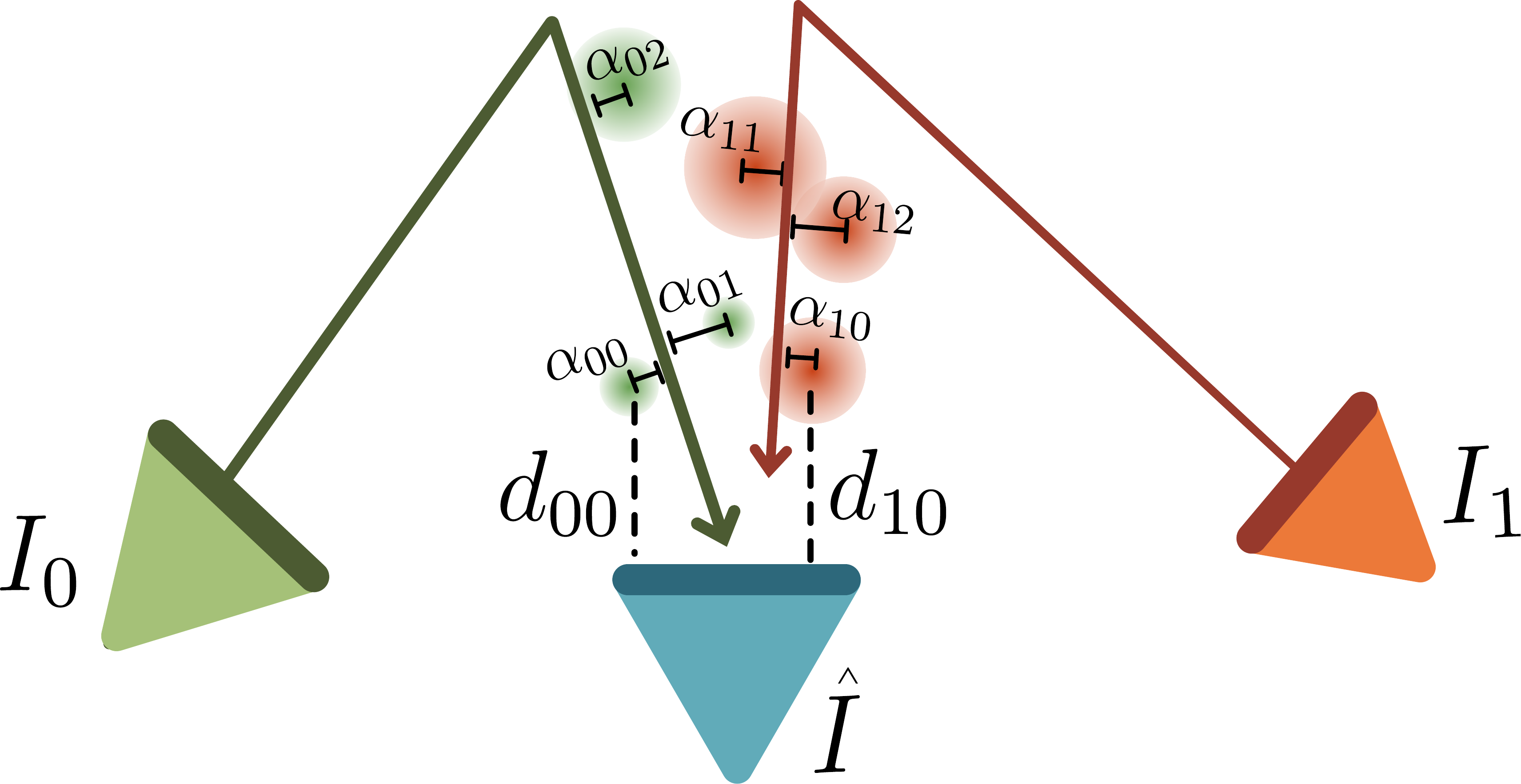}
	\caption{
		\label{fig:probabilistic_intepretation_depth}
Soft rasterization incorporates uncertainty, allowing this information to propagate to the depth test. This makes all decisions soft, which is required due to the noisy data from MVS reconstruction; hard visibility could discard valuable information.
We see how points that do not directly intersect a ray convey depth information to the novel view.}
\vspace{-.5cm}
\end{figure}

\subsection{Probabilistic Depth Testing for Point-Based Rendering}
\label{sec:depth-test}

A given pixel in the novel view will receive splats of continuous opacity from points coming from different input images (Fig.~\ref{fig:probabilistic_intepretation_depth}). Determining which view is in front of all others is a complex task, further exacerbated by the uncertainty in the depth values of the different points. Many ad hoc solutions have been proposed to this problem of resolving uncertain depth in IBR~\cite{chaurasia2013depth,HRDB16,PLRD21}. Hedman et al.~\shortcite{hedman2018deep} rely on the learned blending step to make the correct decision, which is not always reliable.
Soft Z-buffering methods (e.g., \cite{tulsiani2018layer}) are not directly applicable since they treat visibility during rasterization; we need to maintain separate information for each view, and our depth test determines which input view to use \emph{after} rasterization.

\CR{Initial experiments showed that our neural renderer} \CR{struggles to learn the depth test without any inductive bias. To introduce the depth-test explicitly, we define a distribution of depths $d_n$ for a given input view $n$. The density function of this distribution is given by the splats reprojected with soft rasterization}; we can thus opt for a probabilistic approach to this depth test that amounts to determining whether
the points reprojected from $n$ are more likely to be in front of those projected from all other views.

\noindent
We define a random variable $D_n$ representing the depths projected to a novel-view pixel.
Our goal is thus to determine probability $P(D_n<\min_{m\neq n}(D_m)) $ i.e., that $D_n$ is closer than all other $D_m$.

We derive a solution to this problem based on a mixture model. 
Intuitively, we aim to compute the probability that the re-projection of a pixel for input view $n$ on a pixel of the novel view is closer than all re-projections coming from the other input views. 

We first rewrite this as a product of probabilities that compares the depth distributions coming from two views. Using a mixture model further allows us to compare each component of the two mixtures, leading to a quadratic computation with respect to the number of points re-projected on the pixel. We choose a triangle distribution that allows for softness in the depth test and accounts for depth uncertainty, while having a finite support. We choose this distribution over Gaussians, since it simplifies and speeds up the quadratic computations.
The detailed derivation is presented in the supplemental; we state the final expression here:
\begin{align} 
P(D_n<\min_{m\neq n}(D_m)) &\approx \nonumber \\ 
\frac{2\sigma}{S} 
\sum_{i \in \mathcal{N}_n}
\beta_{n,i}\sum_{t=1}^{S}& \prod_{m\neq n}\left(
\sum_{j \in \mathcal{N}_m}
\beta_{m,j} T(s(t),d_{m,j},\sigma)\right)\;f_i(s(t),d_{n,i}),\;
\label{eq:soft-depth}
	\end{align}

\noindent
with $T$ the integral of the symmetrical triangular distribution $f$ with support $2\sigma$, $d_{m,j}$ the depth of point $j$ in view $m$ and
	\[\beta_{n,i}=\alpha_{n,i}\prod_{j=1}^{i-1}(1-\alpha_{n,j}).\]
Further, $S$ is the number of samples and
$s(t) = d_{n,i}-\sigma+\frac{t}{S+1}$.
We found that setting $S=1$ provided satisfactory results in all our experiments. This computation is performed in parallel per pixel using CUDA.

\section{Temporally Consistent Neural Rendering and Optimization}

Our bi-directional point splatting approach allows us to define a powerful neural renderer that we will use both for multi-view imaging and rendering as well as for per-view attribute optimization.

\subsection{Point-Based Neural Renderer}
\label{sec:nr}

\begin{figure}
\includegraphics[width=\linewidth]{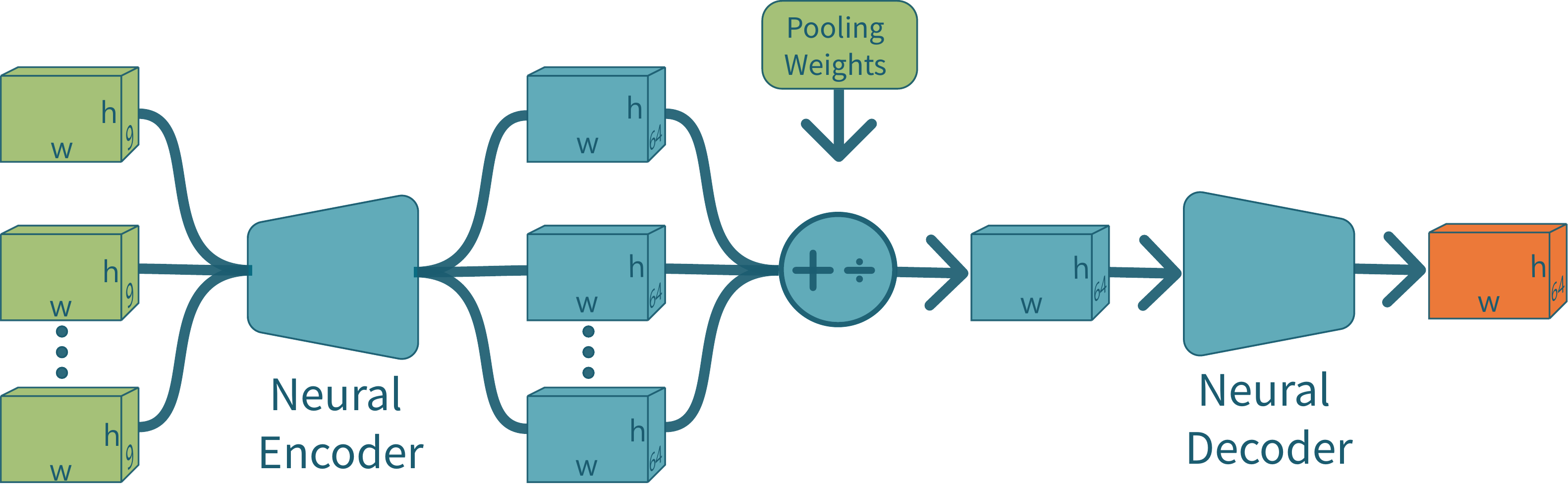}
\caption{
\label{fig:network}
\CR{Our neural network architecture takes as input the rasterized point clouds and feeds each one individually in an encoder network. The encoded rasterizations are then pooled by a set of weights that introduce inductive bias for the visibility but also allow for temporal smoothing in an interactive scenario. The pooled features are fed through a decoder to generate the final rendering.}
\vspace{-0.2cm}
}
\end{figure}

For each view we generate a set of attribute layers, namely the input-view colors and a set of optimized latent features reprojected to the novel view using our point splatting method. 
The architecture of our pipeline is visualized in Fig.~\ref{fig:network}.

The first part of the network encodes each view in a feature space of $\mathcal{F}$ dimensions for each input view with a series of convolutional residual blocks, with shared weights across all input views.
Then we apply a weighted average pooling \cite{zaheer2017deep} and produce a single $\mathcal{F}~=~64$ channel image, which is subsequently decoded by consecutive convolutional residual blocks. 
The weights for each input image consist of three terms, as detailed below.

First, camera selection (Sec.~\ref{sec:cam-sel}) can cause temporal instability when cameras appear/disappear for a selected novel view. We deal with this problem with a smooth fading strategy using temporal filtering.
Our camera selection algorithm returns a set of selected cameras that get a score of $s=1.0$ while the rest get $s=0.0$.
We temporally filter the score $s_i^t$ of view $i$ in frame $t$ as follows:
\begin{equation*}
	w_i^t = \lambda s_i^t + (1.0 - \lambda)w_i^{t-1}
\end{equation*}
where $\lambda$ controls how fast the temporal filtering adapts to changes; we set $\lambda=0.05$.
Once we update the weights, we select the $N$ highest to use for rendering. Unless stated otherwise, we used $N~=~9$ in all our experiments.
To avoid popping we further refine the weights across all views, similar to the approach used by \cite{sun2020light}, to obtain final smooth camera selection weights $w_{CS}$:
\begin{equation*}
\label{eq:w}
	w_{CS} = \frac{ \mathbf{w} - \min_i w_i}{\sum{(\mathbf{w}-\min_i(w_i))}}
\end{equation*}
ensuring that views which are coming in and out of the set have zero weight, and
$\min_i$ is the element-wise min of a vector, where $\mathbf{w}$ is the vector of all $w_i$ and the equation uses element-wise difference. This moves the $C_0$ discontinuity to $C_1$, and thus every time the selected set changes, the gradients of the weights become discontinuous instead of the actual values of the weights. 

Second, we observe that weighted average pooling can lead to artifacts in the presence of view-dependent effects. To compensate for this, we use weights based on input-view texture stretch: For each point reprojected in the novel view we quantify the amount of stretch as the ratio between the two eigen-values of the covariance of the splatting kernel $\mathbf{G}$ (Sec.~\ref{Differentiable-Point-Cloud-Rasterization}) \cite{kopf2014first}. This weight $w_{TS}$ penalizes pixels at grazing angles, increasing the influences of more front-facing views.

Third, we employ probabilistic depth weights 
$w_{\mathrm{PD}} = P(D_n<\min_{m\neq}(D_m))$ 
(Sec.~\ref{sec:depth-test}, Eq.~\ref{eq:soft-depth}) which account for point visibility.

The final weights $w$ are the product of the three weights described above: 
\begin{equation*}
\vspace{-0.1cm}
	w = w_{\mathrm{CS}} \cdot w_{\mathrm{TS}} \cdot w_{\mathrm{PD}}
\end{equation*}
Compared to previous neural renderers, this architecture shows improved temporal stability. Importantly, if we decide to change the number of views we select, we do not need to retrain. This is because the weighted average pooling normalizes the sum of the weights of all views per pixel, hence no matter how many views are added the distribution of the magnitude of the features will stay the same for the decoder.

\subsection{Optimization}
\label{sec:optimization}
We optimize the scene representation and our neural renderer jointly. 
The list of attributes we optimize in each input view reads as:
color, depth, normal, uncertainty, and a 6-component latent feature vector.
\CR{The feature vector extends the input-view color channels and allows our pipeline to encode useful additional information per input-view pixel}. \CR{The decoder uses the optimized features to improve rendering.} We initialize the six features to 0.5, and depth is initialized with the generated per-view meshes from Hedman et~al.~\cite{hedman2018deep}. The neural renderer is initialized using the method of Zhang et~al. \cite{zhang2018residual}.

We perform optimization with a leave-one-out strategy, i.e., at each iteration we randomly hold out one view to use as ground truth. 
We select the 9 views (chosen randomly from the best 13), which empirically gave the best compromise in our test scenes; please see Sec.~\ref{sec:ablations} for an ablation study on the number of views selected.
We lift all the pixels to 3D using the depth maps of their corresponding input views so we can splat them to the view we want to render. We stack all views in batches and feed them through the neural renderer. We use an $L_1$ loss
\begin{equation*}
	L = \|I-I_{gt}\|_1,
\end{equation*}

\noindent
while certain applications require additional loss terms as described in Sec.~\ref{sec:applis}.
We use an ADAM optimizer for all our parameters and we renormalize normal vectors in each iteration to avoid numerical instabilities.  We also feed the features through a sigmoid to scale them to unit range so they are in the same range as colors, before feeding them to the next module. The learning rates we use are the following: $lr_{\mathrm{CNN}} ~=~ lr_{\mathrm{normal}} ~=~lr_{\mathrm{depth}}=0.0001$,  $lr_{\mathrm{features}}=0.001$, $lr_{\mathcal{U}}=0.01$.
Unless stated otherwise, we trained all scenes for 100,000 iterations in 150x150 patches, at which point the loss no longer improved.

\begin{figure}[!ht]
\centering
\includegraphics[width=\linewidth]{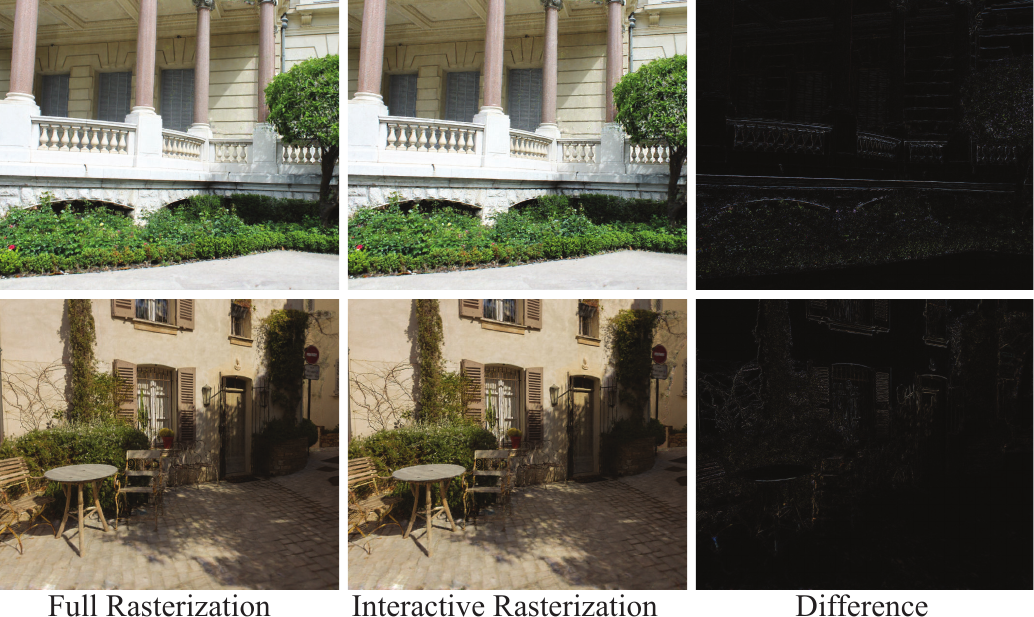}
\caption{
\label{fig:comparison_ogl}
Comparison of our full soft rasterization approach with the interactive version. Differences are subtle and do not significantly degrade visual image quality.
\vspace{-.2cm}
}
\end{figure}

\subsection{Forward Rendering}
\label{sec:forward-rendering}
Once optimized for a given scene, our neural network can be used for high-quality free-viewpoint rendering. For each novel view, we choose the input views using our camera selection approach, then project the points from these views using our splatting approach, perform the probabilistic depth test and then run our neural network to synthesize the image.

A major bottleneck of the forward pass is the point-splatting. During training, we need to use the full approach including the gradient computations in the backward pass. However, for fast rendering at runtime, we have implemented an efficient OpenGL-based approximation of the point reprojection in Sec.~\ref{Differentiable-Point-Cloud-Rasterization}.
We use $k_d=10$ global depth layers, where the relevant depth interval is determined by rasterizing the MVS mesh depth at low (1/16) resolution followed by a min-max mipmap construction.
We then splat all pixels of all input views in parallel into the depth layers, where the multiplicative accumulation of opacity (Eq.~\ref{eq:front-to-back}) utilizes hardware accelerated additive blending of $\log(1-\alpha)$.
We composite the depth layers back-to-front in parallel over all re-projected pixels of all input images. We then apply the probabilistic depth test (Sec.~\ref{sec:depth-test}) that now only needs to consider at most $k_d$ alpha-weighted depth samples. Quality degrades only marginally using this approximation, allowing interactive viewing; please see Fig.~\ref{fig:comparison_ogl} and the supplemental for further visual comparisons and statistics.
Results are computed at full solution unless otherwise stated.
Our unoptimized implementation runs at 4.5\,fps, where camera selection takes 87\,ms, point splatting 36\,ms, the depth test 38\,ms, and network evaluation 59\,ms. \CR{While we achieve interactive frame-rates, this comes at the cost of memory consumption. First, the depth layers take approximately 4\,GB of memory  for a resolution of $900\times600$ with 10 layers and 9 views. This could be compressed through standard texture compression methods. Second, the fact that we batch together all the input views and feed them through our neural renderer means that memory consumption is high for this stage as well; pytorch reported 4.5\,GB of memory usage for the above configuration.}

\begin{figure}[!ht]
\centering
\begin{tabular}{cc}
\includegraphics[width=.48\linewidth]{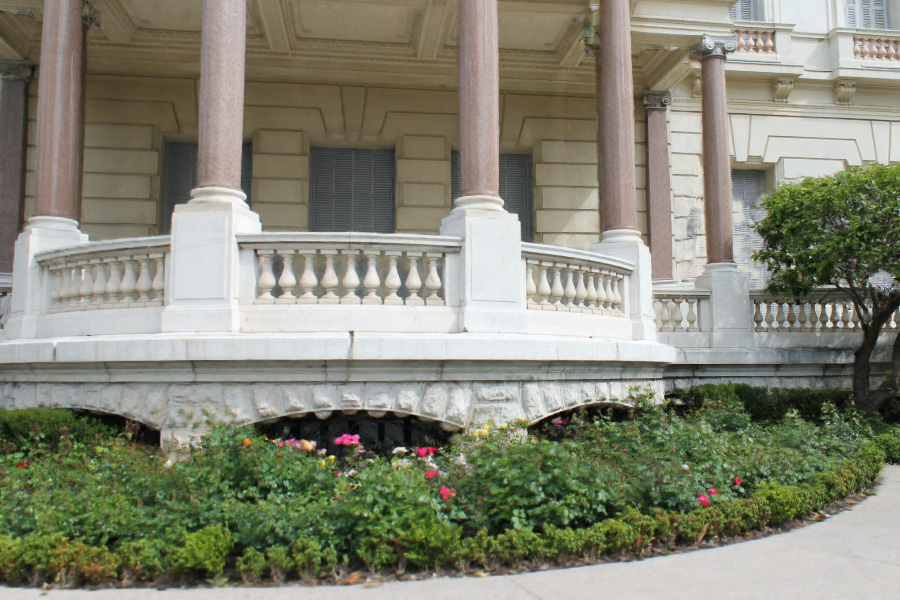}&
\includegraphics[width=.48\linewidth]{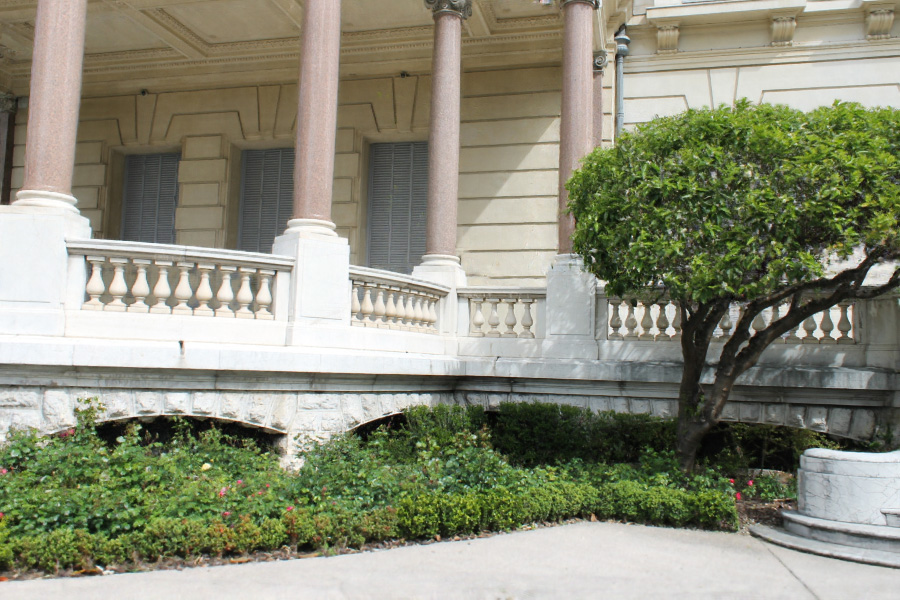} \\
\includegraphics[width=.48\linewidth]{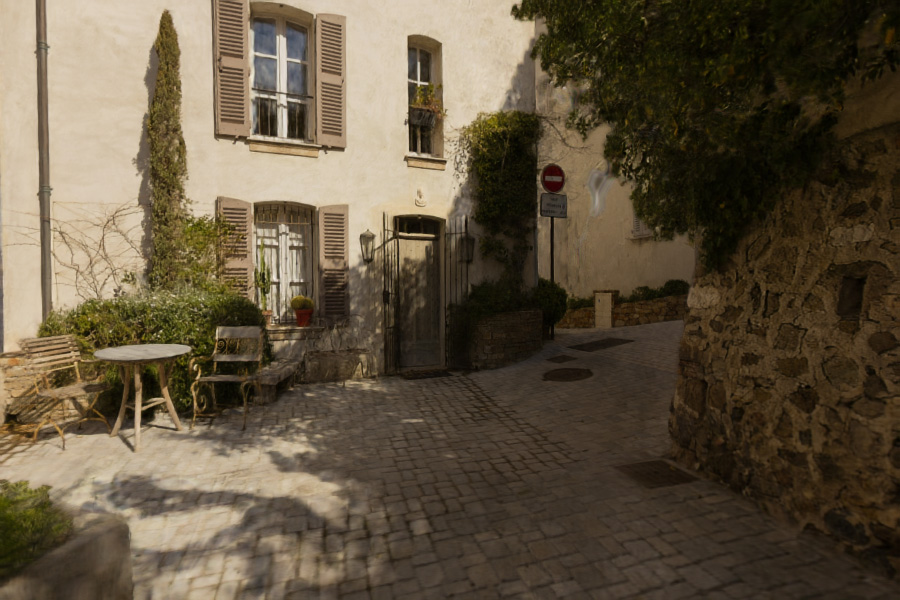} &
\includegraphics[width=.48\linewidth]{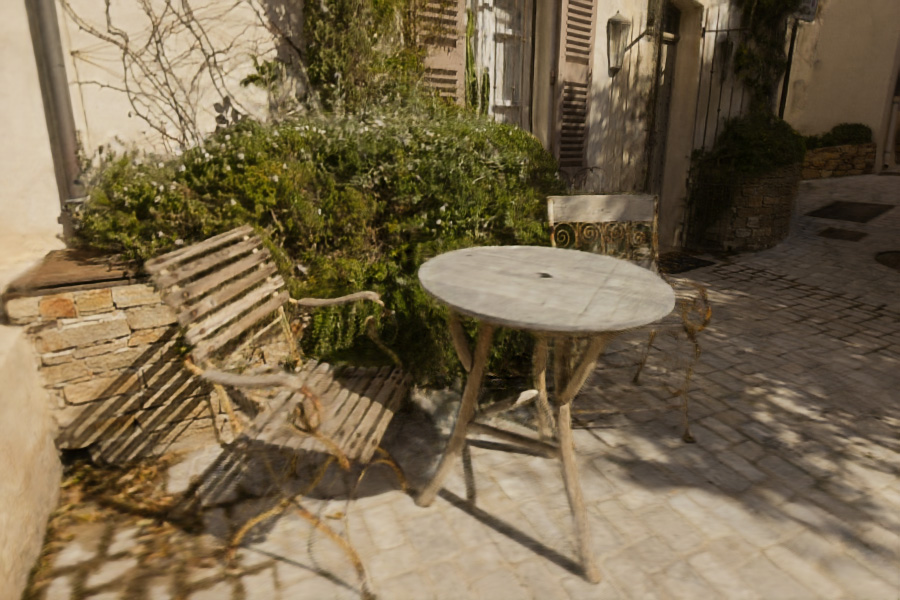}\\
\includegraphics[width=.48\linewidth]{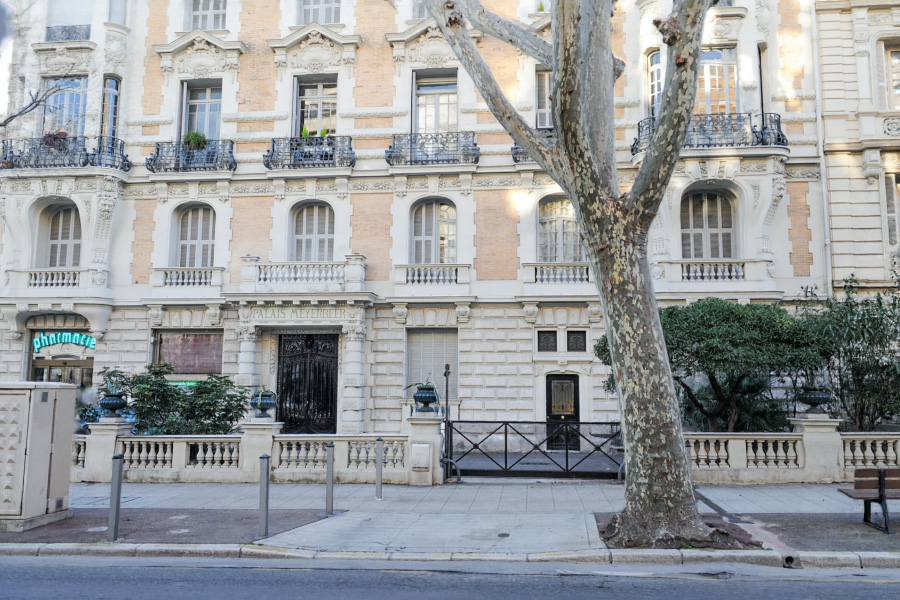} &
\includegraphics[width=.48\linewidth]{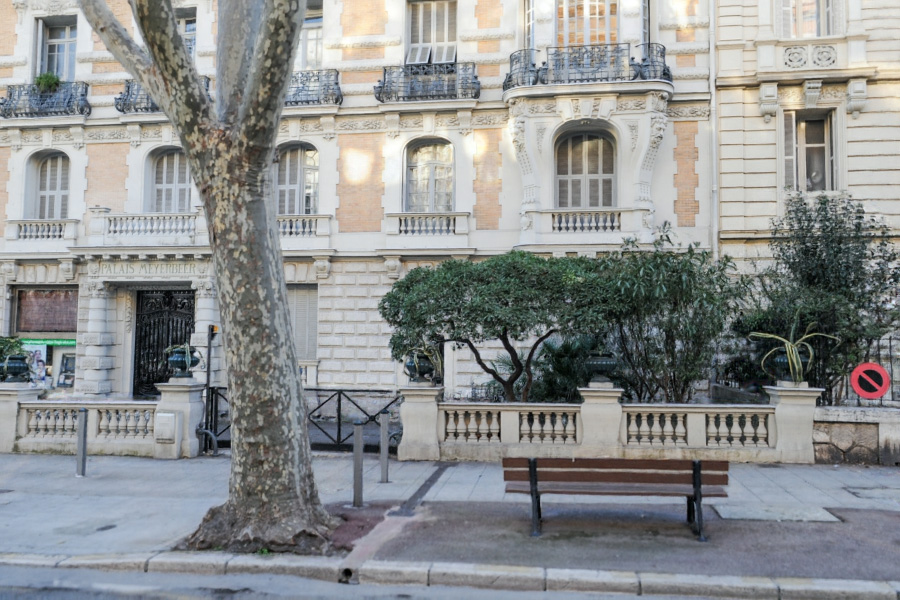} \\
\includegraphics[width=.48\linewidth]{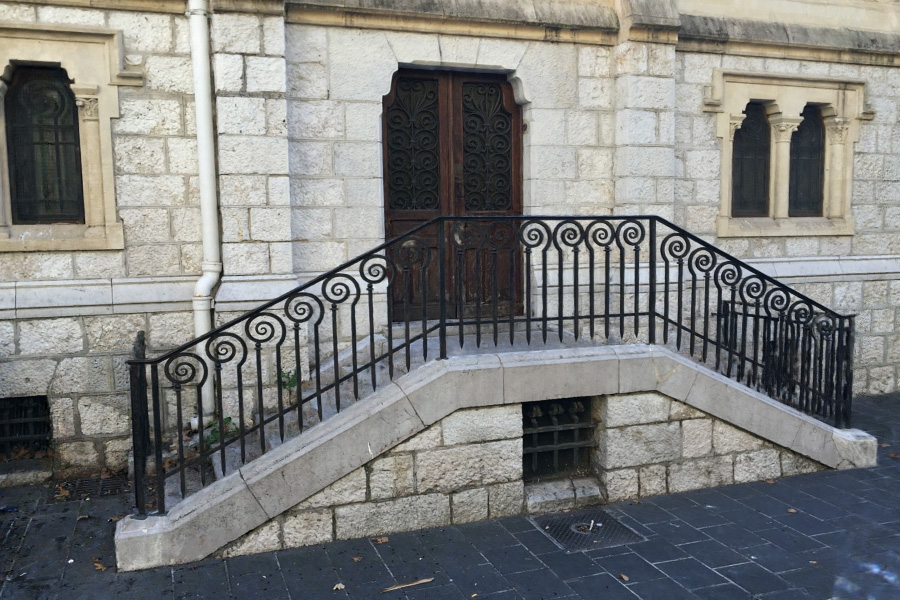} &
\includegraphics[width=.48\linewidth]{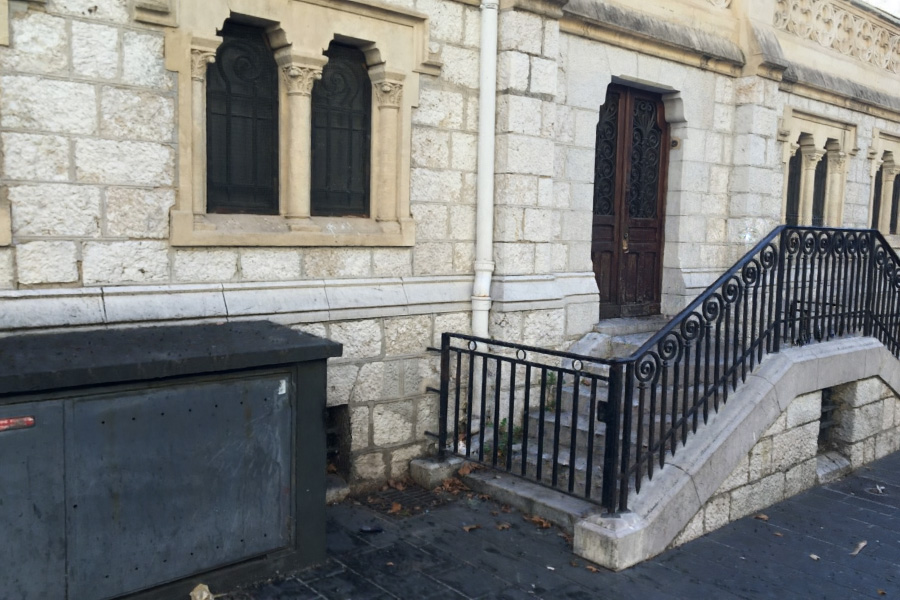}	
\end{tabular}
\caption{
\label{fig:ibr-results}
Novel views synthesized with our method for the scenes (top to bottom): Museum, Ponche, Hugo and Stairs.
}
\end{figure}

\begin{figure*}[!ht]
\includegraphics[width=\linewidth]{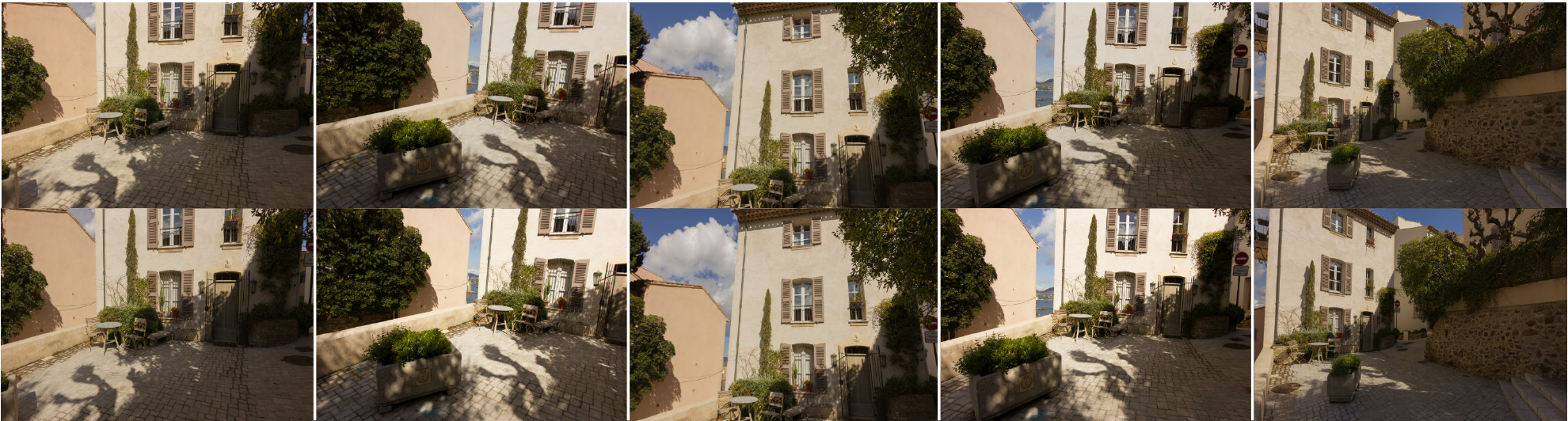}
\caption{
\label{fig:harm-results}
Multi-view capture can suffer from differences in exposure and other camera parameters between views; our algorithm optimizes for a brightness coefficient to achieve harmonization e.g., in the second and fourth image. Top row: the images adjusted by our optimization. Bottom row: original images.
}
\end{figure*}

\section{Applications and Results}
\label{sec:applis}

We illustrate our differentiable multi-view pipeline on three applications. The first is IBR, the second multi-view harmonization and the third multi-view style transfer. 

We implemented our system in PyTorch, but wrote custom CUDA kernels for the point splatting algorithm used in training and accurate rendering, since available implementations (e.g., PyTorch3D) are an order of magnitude slower and we have integrated our network into a C++ framework with OpenGL for display; our interactive version implements point splatting in OpenGL. We will release all code \CR{and data, please see \url{https://repo-sam.inria.fr/fungraph/differentiable-multi-view}, which also includes the supplemental website}.

\subsection{Image-Based Rendering}

Our IBR pipeline uses our neural renderer directly, and allows free-viewpoint navigation. \CR{We show example images in Fig.~\ref{fig:ibr-results}; three different scenes, taken from the Deep Blending datasets~ \footnote{See \url{http://www-sop.inria.fr/reves/publis/2018/HPPFDB18/datasets.html}.} and the ``Stairs'' scene from~\cite{thonat2018thin}. We also show the Truck scene from Tanks and Temples~\cite{knapitsch2017tanks} in Fig.~\ref{fig:comparisons}.} Our method achieves sharp results in regions with vegetation and can recover from some of the reconstruction artifacts due to thin structures. We encourage the reader to watch the supplemental videos to appreciate the visual quality of our method. 

All timings are reported on an RTX6000 GPU for display and RTX8000 for training. Training for 100K iterations takes 12-14h. For small scenes (Hugo and Tree) only 20K iterations are required (approximately 3h).

\subsection{Multi-view Harmonization}

We demonstrate a multi-view harmonization technique that works well for one of the most common problems in real-world multi-view captured content: 
Exposure and other camera parameters can fluctuate between images, creating multi-view inconsistencies breaking basic IBR algorithm assumptions. We model this as an additional coefficient $\mu_i$ per view that is multiplied with the color of the corresponding input image. We initialize $\mu_i=1.0$ and we allow $\mu_i$ to be optimized like any other per-view parameter. This parameter modifies the input images that are both used for ground truth and for re-projections. It is thus necessary to avoid degenerate solutions during the optimization i.e., $\mu_i=0$ for all views. 
We address this problem by adding a regularization term in the loss function:
\begin{equation}
	L_\mu = \lambda\frac{\sum_i^N{(\mu_i - 1.0)^2}}{N},
\end{equation} 
where $N$ is the number of input views and $\lambda=0.2$ is a hyper-parameter that controls the weight of the reguralization.
We also introduce a photoconsistency loss between the optimized image $I_n$ and the other reprojected images $R(I_m)$, where $M$ is a mask of the pixels containing content in $R(I_m)$.
\begin{equation}
	L_{\mathrm{photo}} = \sum_{m\neq n}(M I_n-R(I_m))^2
\end{equation}

We illustrate the effect of harmonization on 5 images taken from the Ponche scene
in Fig.~\ref{fig:harm-results}, and visualize the optimization of individual $\mu_i$ in the supplemental.

\subsection{Multi-view Style-Transfer}

For multi-view style-transfer we leverage the differentiability of our point-based reprojection to jointly optimize each input image to match a given style, while maintaining photo-consistency.
We base our stylization method on the approach of Mechrez et~al.~\shortcite{mechrez2018contextual}.
We first make the input image colors parameters of the optimization process. Then we add the following multi-view style-transfer loss to the optimization:
\begin{equation}
	L_{\mathrm{mvst}}(I_n,R(I_{m\neq n}))=L_{\mathrm{photo}} +CX_{\mathrm{style}}(I_n,S) + CX_{\mathrm{cont}}(I_n,I_n^*)
\end{equation}

The first term ensures photo-consistency. The second and third terms allow style transfer between $I_n$ and a style image $S$, while maintaining content between $I_n$ and its original version $I_n^*$ and use the contextual loss $CX()$ for style transfer as described in Mechrez et al. \shortcite{mechrez2018contextual}. We show the effect of multi-view style-transfer on 5 images taken from the Museum datasets in Fig.~\ref{fig:style-results}. Additional results and comparisons are presented in the supplemental and video.

\begin{figure}[!ht]
\includegraphics[width=\linewidth]{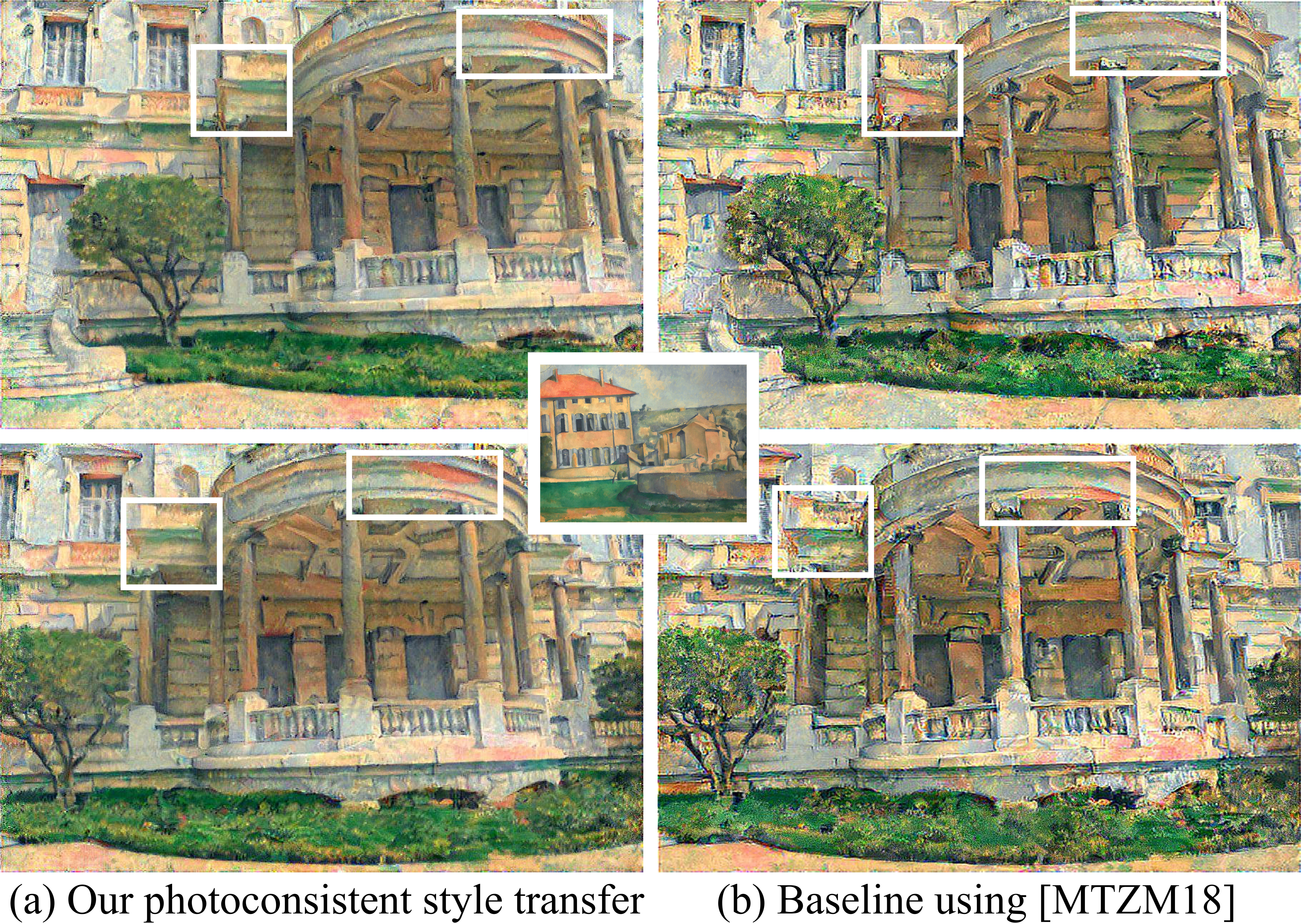}
\caption{
\label{fig:style-results}
Two different input views for each stylization method with the painting shown in the middle. Left: our photoconsistent algorithm. Right: results if we apply the method of  Mechrez et al.~\shortcite{mechrez2018contextual}. Insets in white show multi-view inconsistencies in the baseline.
\vspace{-.2cm}
}
\end{figure}
\begin{figure*}[!ht]
\centering
\setlength{\tabcolsep}{1pt}
\begin{tabular}{ccccc}
	\includegraphics[width=.19\linewidth, height=.126\linewidth]{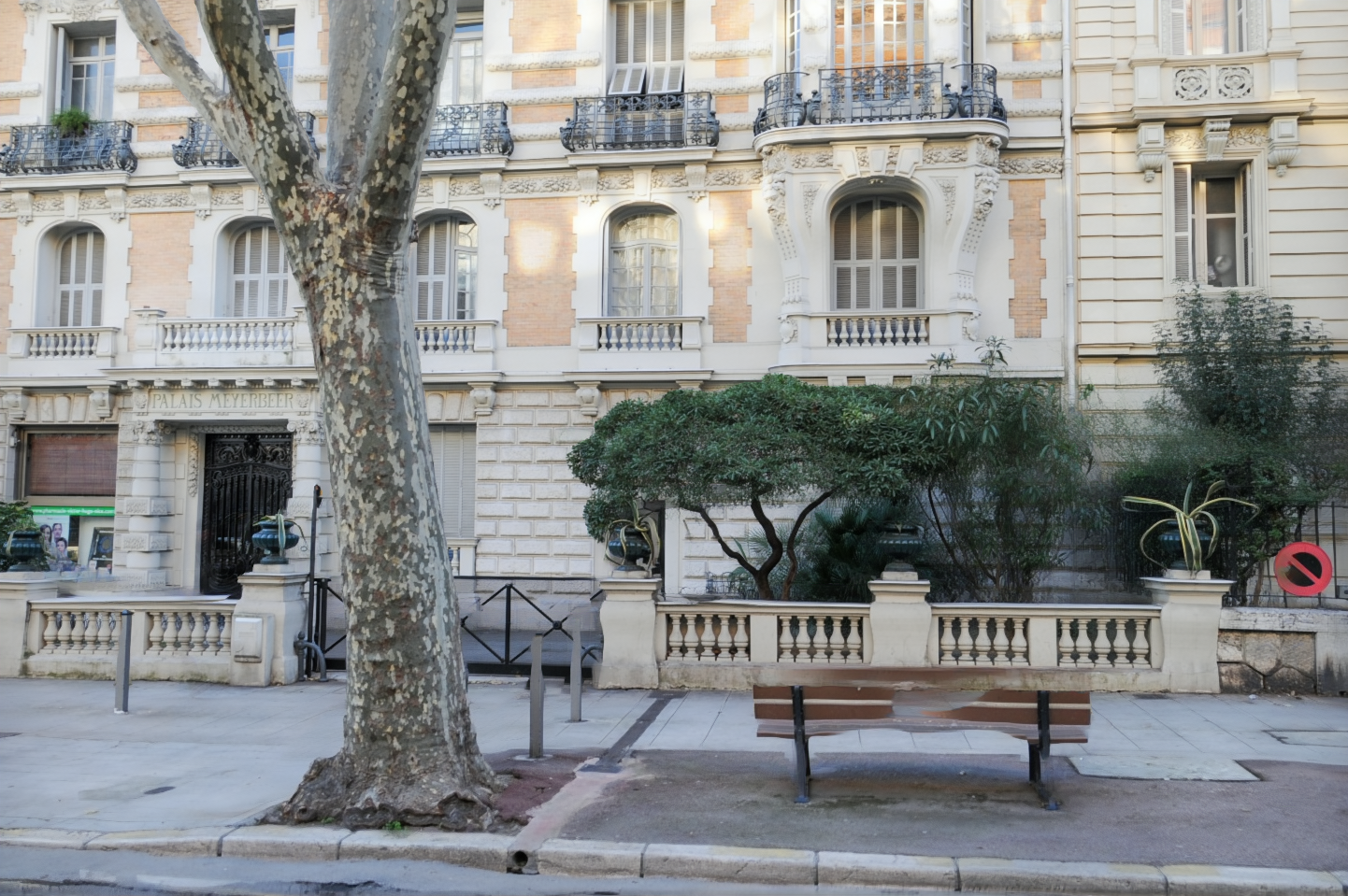} &
	\includegraphics[width=.19\linewidth, height=.126\linewidth]{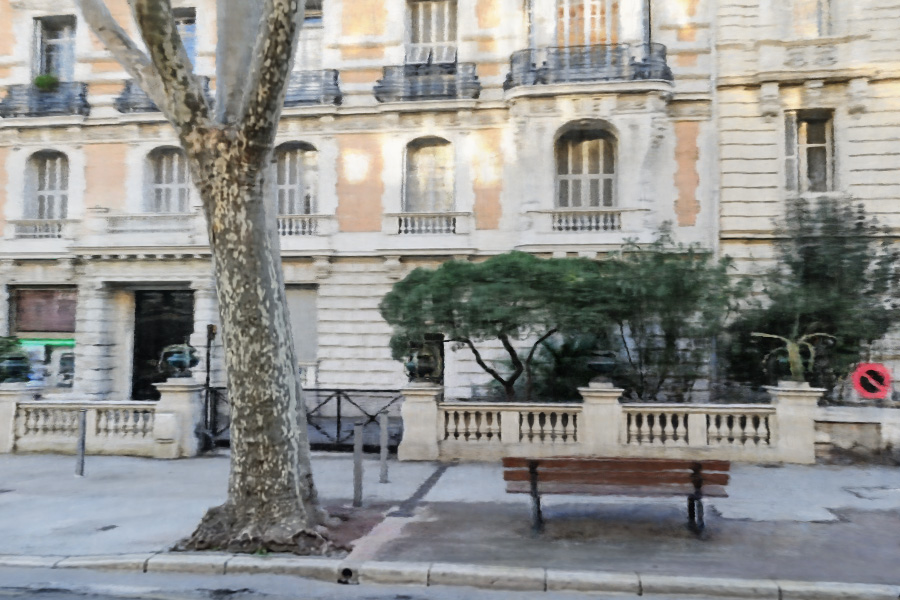}  &
	\includegraphics[width=.19\linewidth, height=.126\linewidth]{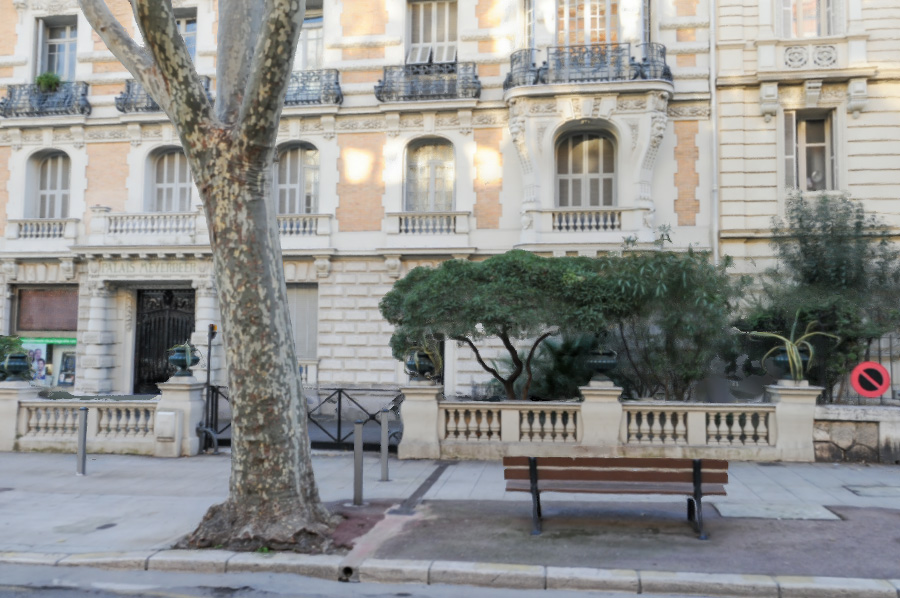} &
	\includegraphics[width=.19\linewidth, height=.126\linewidth]{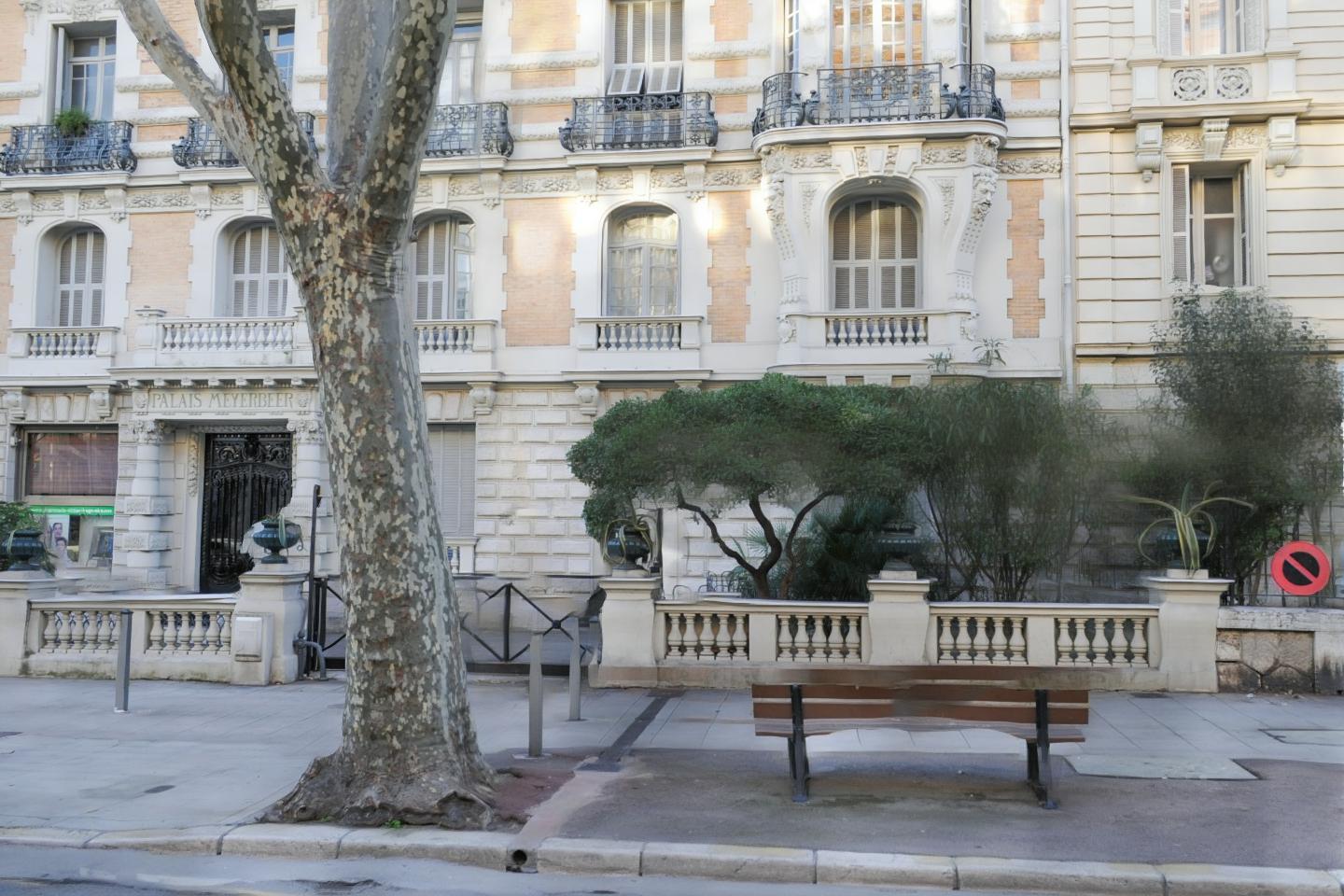}  &
	\includegraphics[width=.19\linewidth, height=.126\linewidth]{images/results/hugo/ours/00000643.jpg} \\
	
	\includegraphics[width=.19\linewidth, height=.126\linewidth]{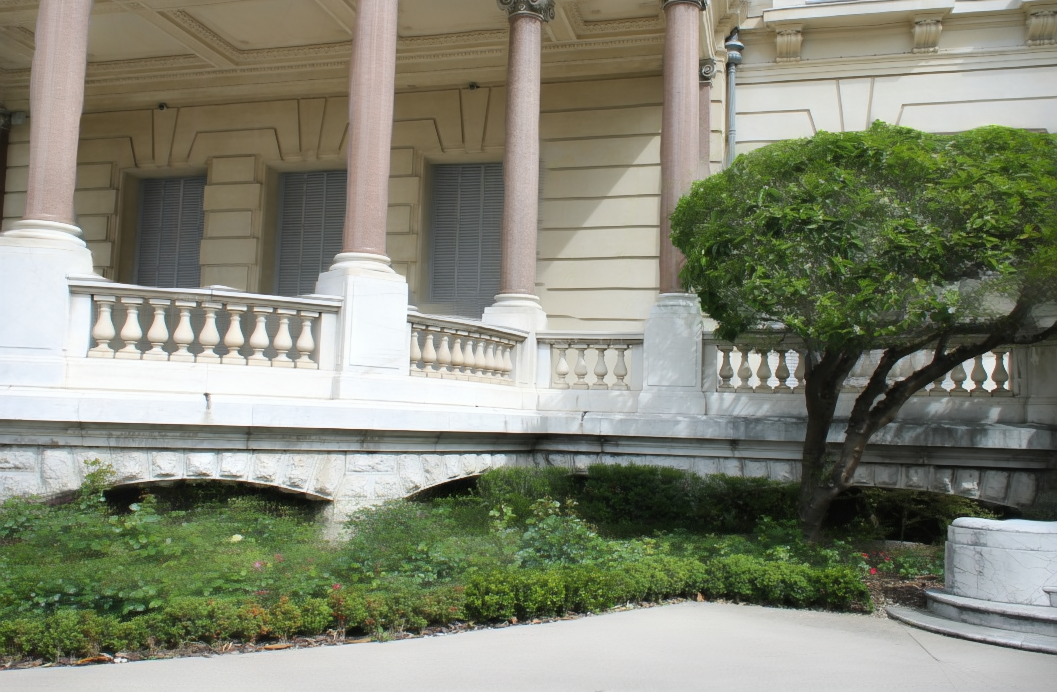} &
	\includegraphics[width=.19\linewidth, height=.126\linewidth]{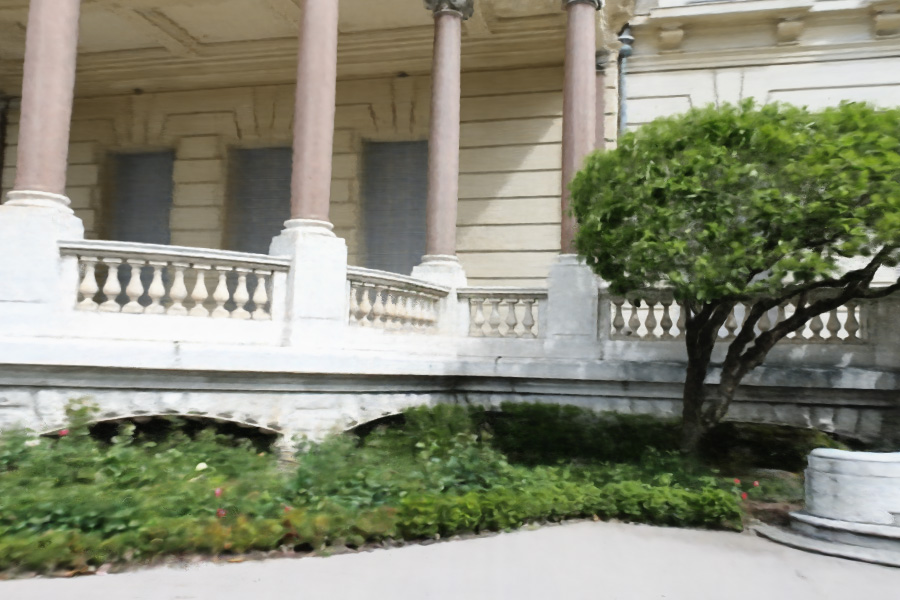}  &
	\includegraphics[width=.19\linewidth, height=.126\linewidth]{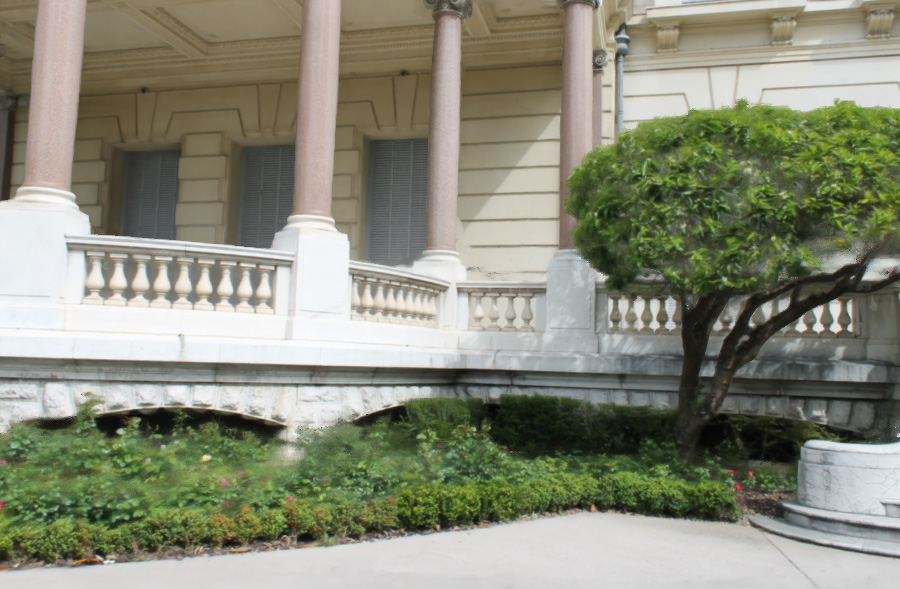} &
	\includegraphics[width=.19\linewidth, height=.126\linewidth]{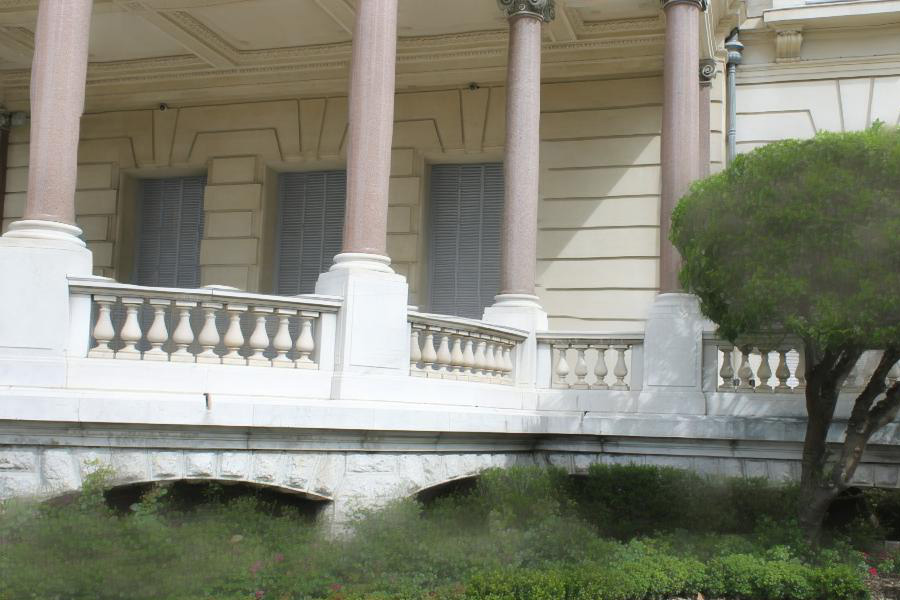} &
	\includegraphics[width=.19\linewidth, height=.126\linewidth]{images/results/museum/ours/00000230.jpg} \\
	
	\includegraphics[width=.19\linewidth, height=.126\linewidth]{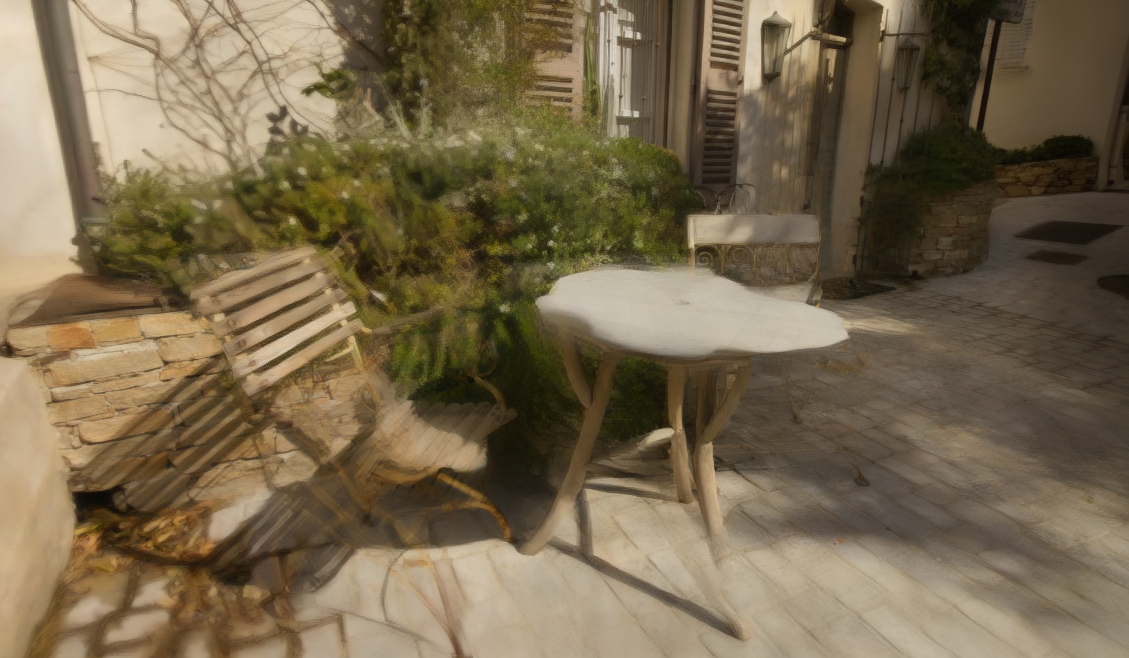} &
	\includegraphics[width=.19\linewidth, height=.126\linewidth]{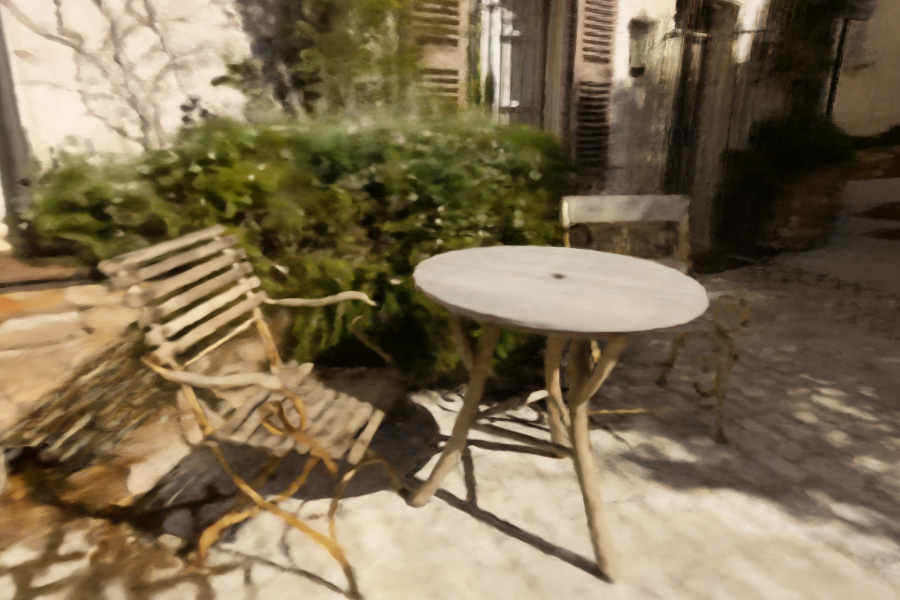}  &
	\includegraphics[width=.19\linewidth, height=.126\linewidth]{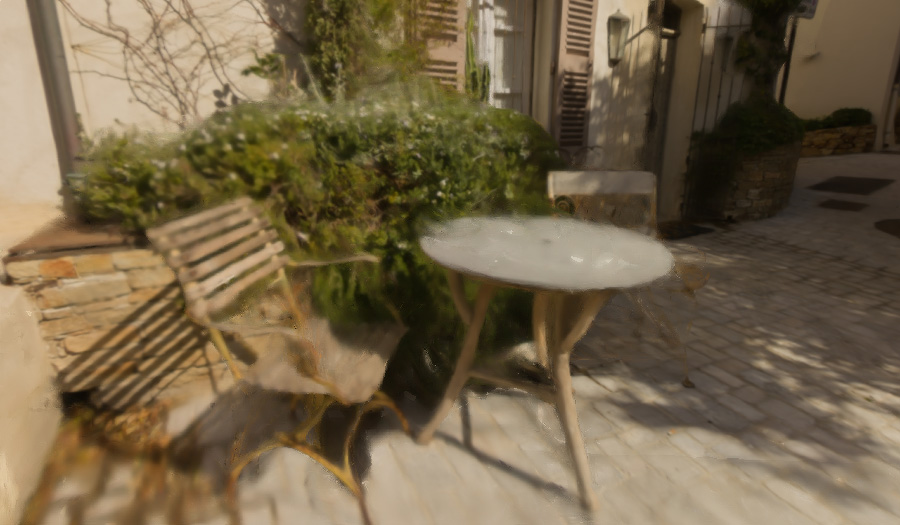} &
	\includegraphics[width=.19\linewidth, height=.126\linewidth]{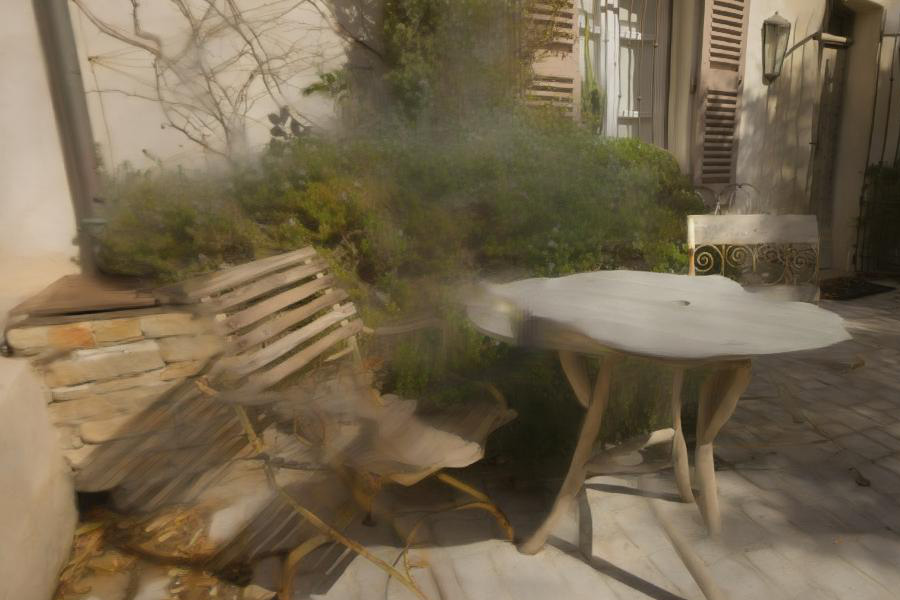}  &
	\includegraphics[width=.19\linewidth, height=.126\linewidth]{images/results/ponche/ours/00000357.jpg} \\
	
	\includegraphics[width=.19\linewidth, height=.126\linewidth]{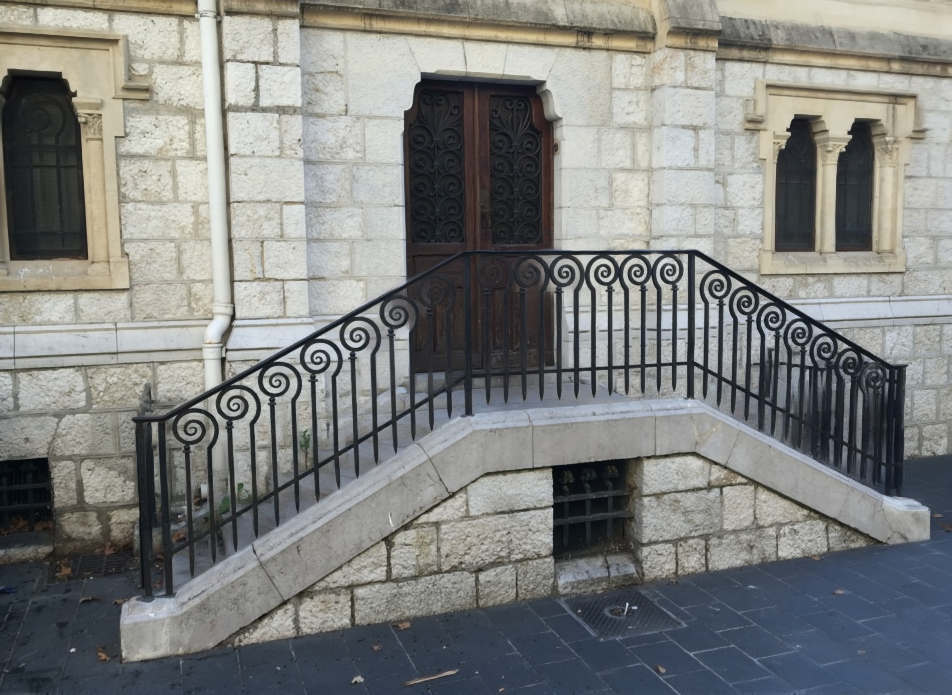} &
	\includegraphics[width=.19\linewidth, height=.126\linewidth]{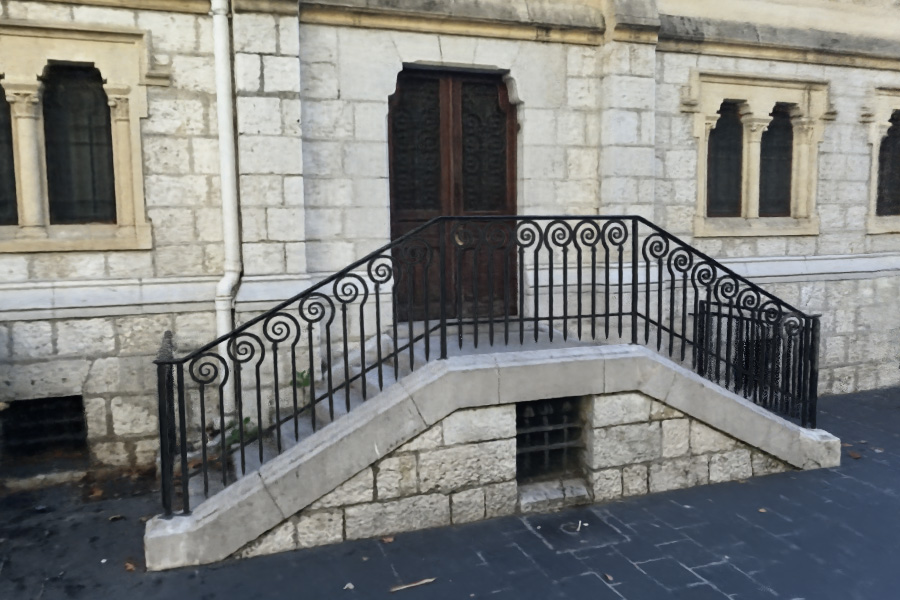}  &
	\includegraphics[width=.19\linewidth, height=.126\linewidth]{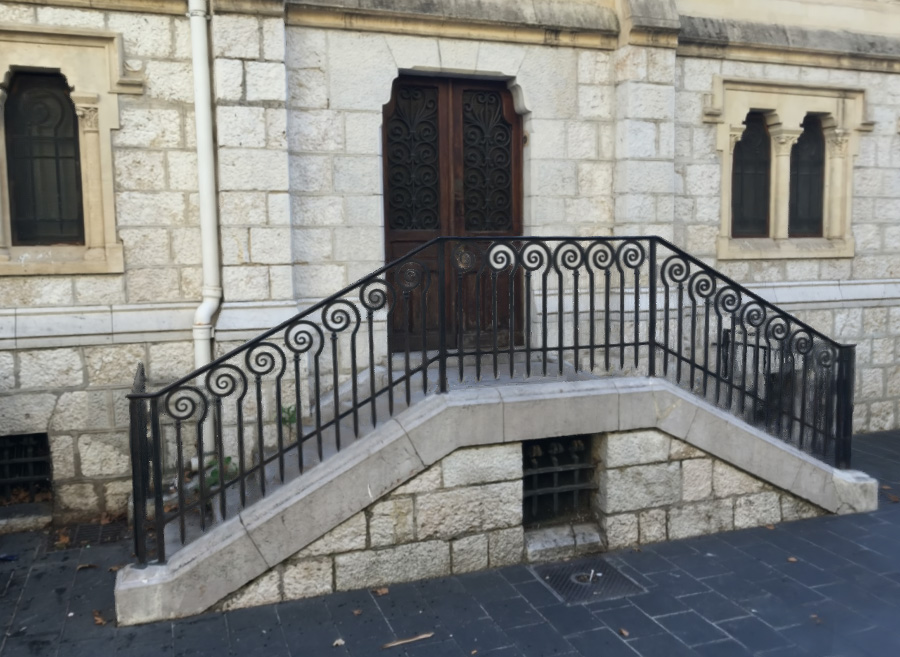}&
	\includegraphics[width=.19\linewidth, height=.126\linewidth]{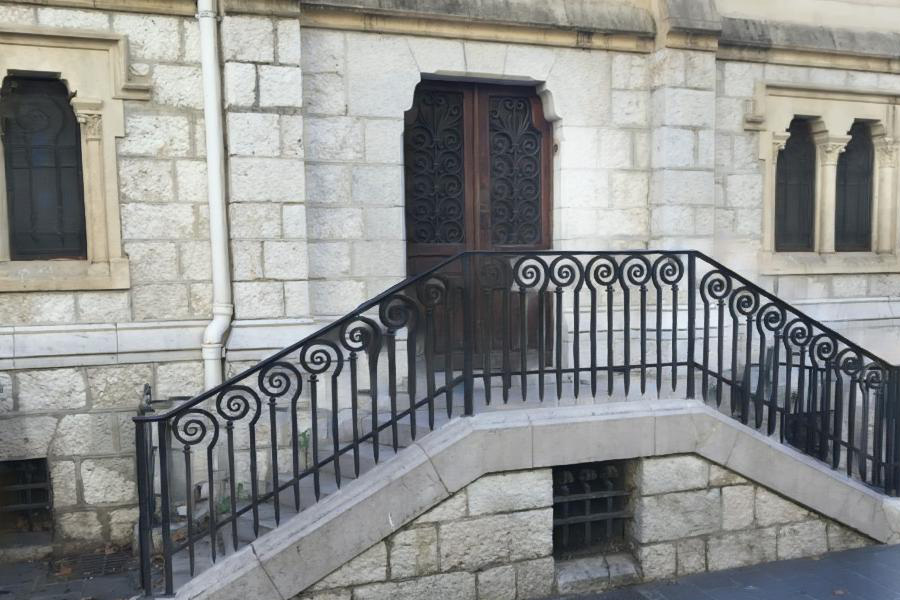} &
	\includegraphics[width=.19\linewidth, height=.126\linewidth]{images/results/stairs/ours/00000000.jpg} \\
	
	\includegraphics[width=.19\linewidth, height=.126\linewidth]{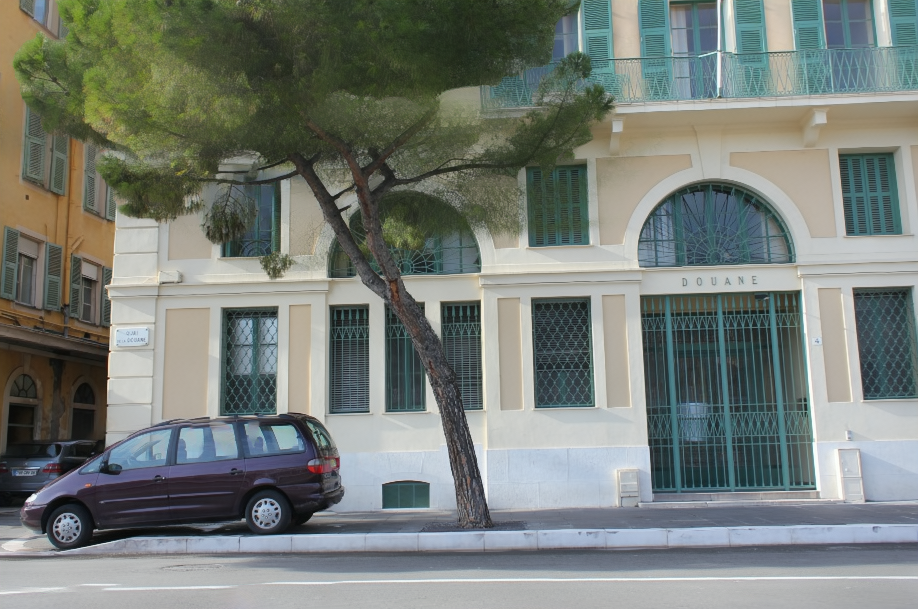} &
	\includegraphics[width=.19\linewidth, height=.126\linewidth]{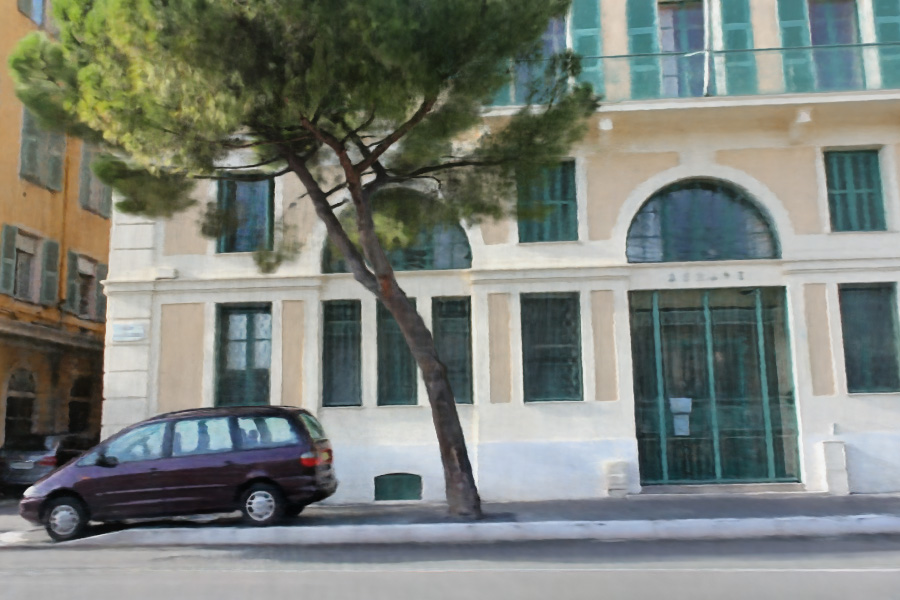}  &
	\includegraphics[width=.19\linewidth, height=.126\linewidth]{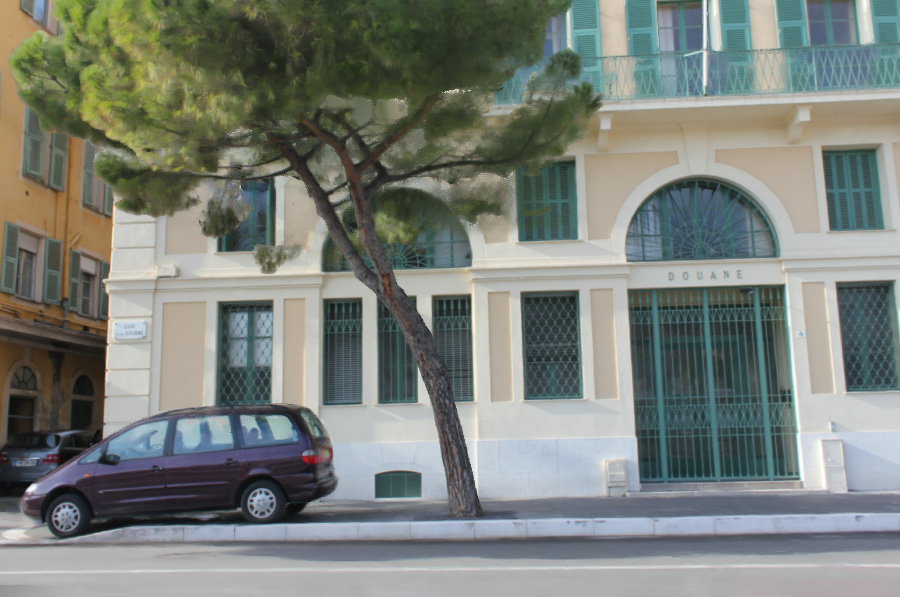} &
	\includegraphics[width=.19\linewidth, height=.126\linewidth]{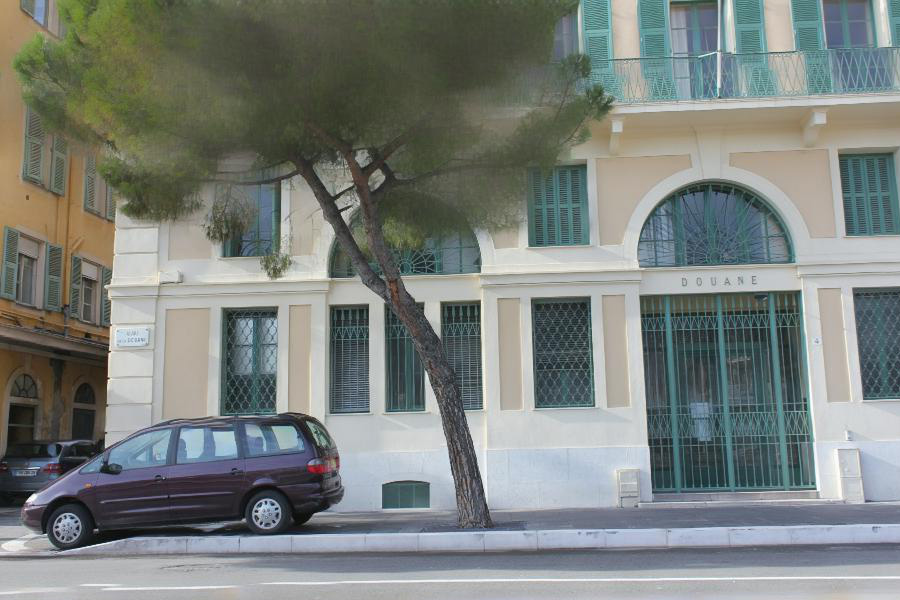} &
	\includegraphics[width=.19\linewidth, height=.126\linewidth]{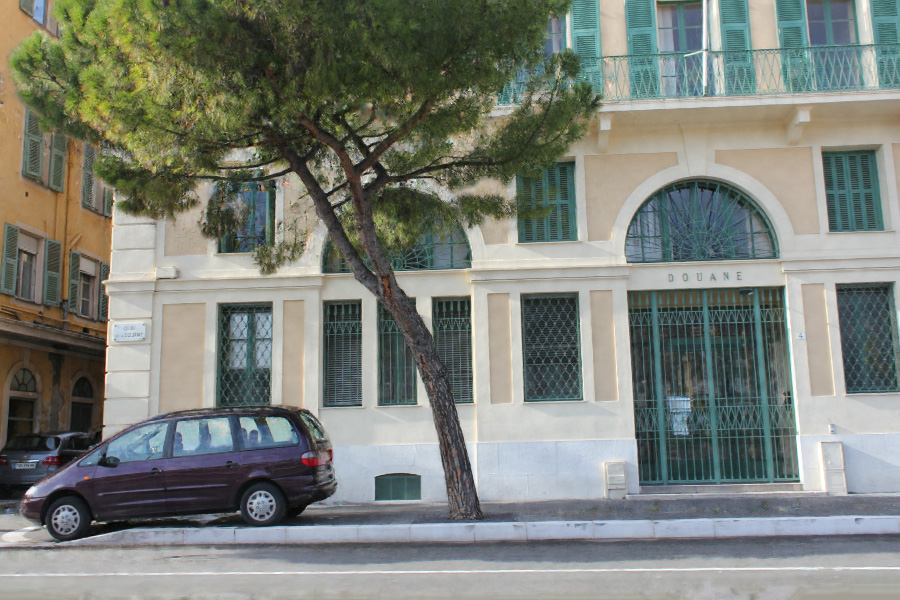}\\
	
	\includegraphics[width=.19\linewidth, height=.126\linewidth]{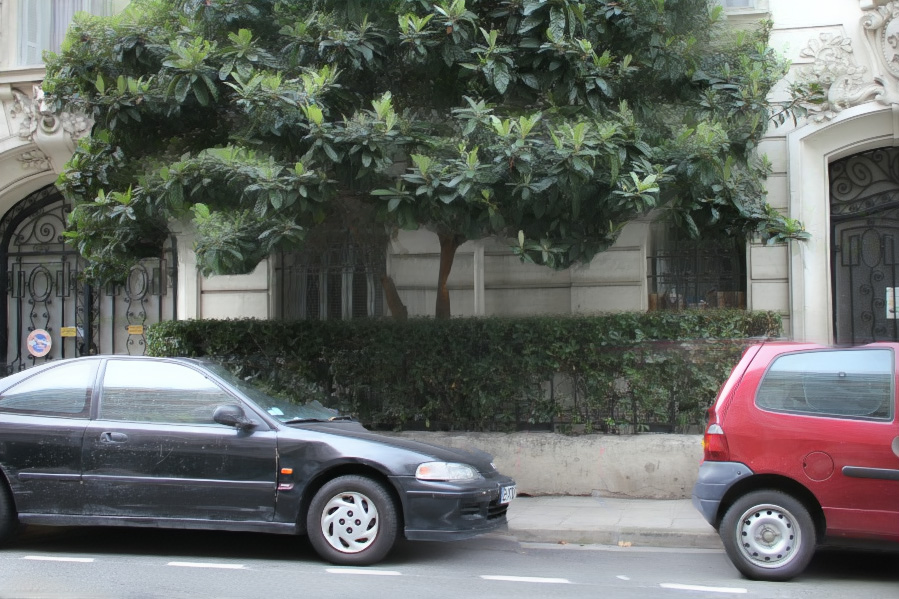} &
	\includegraphics[width=.19\linewidth, height=.126\linewidth]{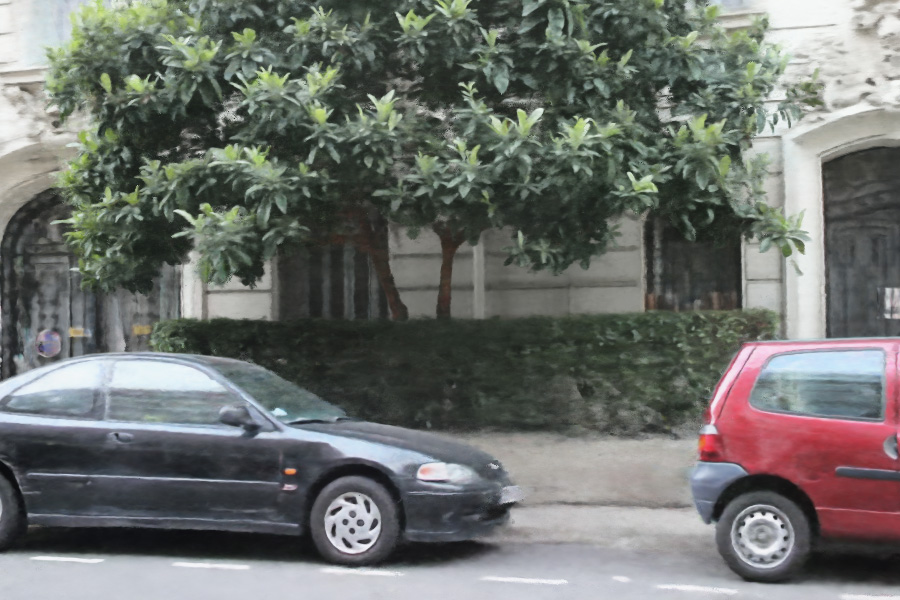} &
	\includegraphics[width=.19\linewidth, height=.126\linewidth]{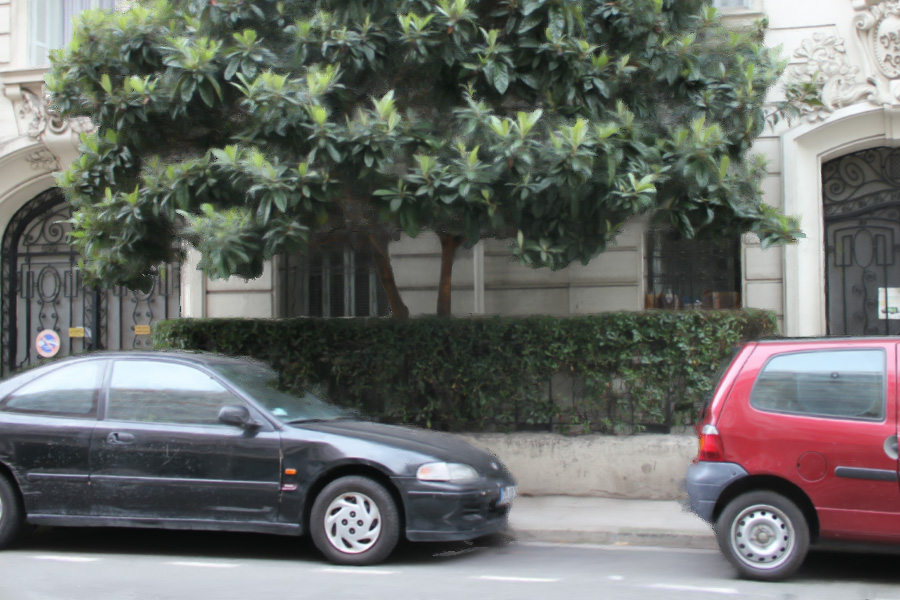} &
	\includegraphics[width=.19\linewidth, height=.126\linewidth]{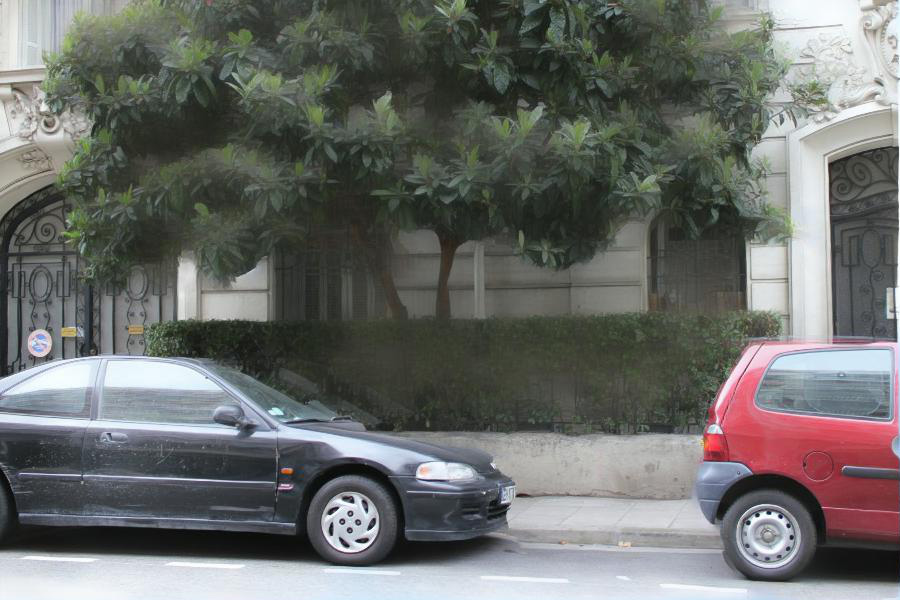} &
	\includegraphics[width=.19\linewidth, height=.126\linewidth]{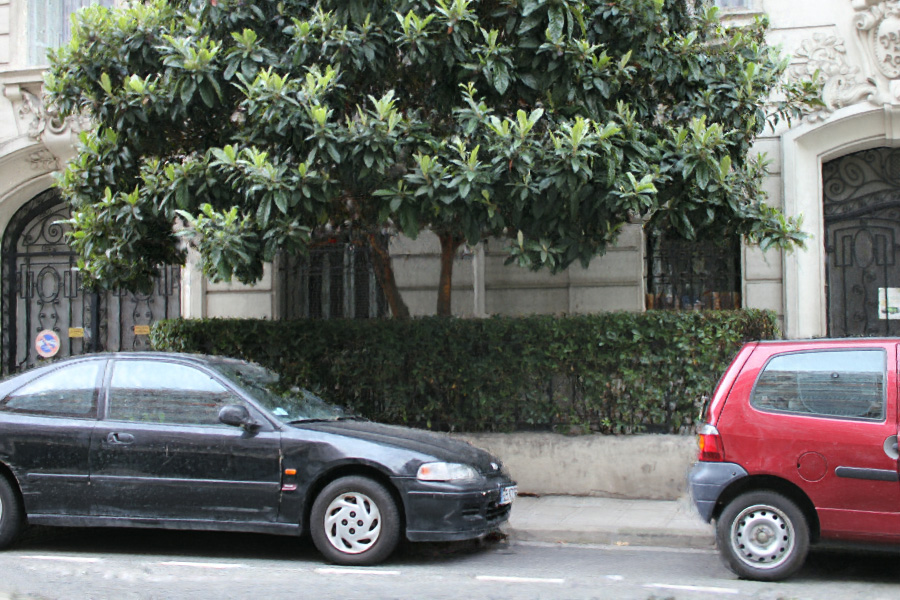}\\
	
	\includegraphics[width=.19\linewidth, height=.126\linewidth]{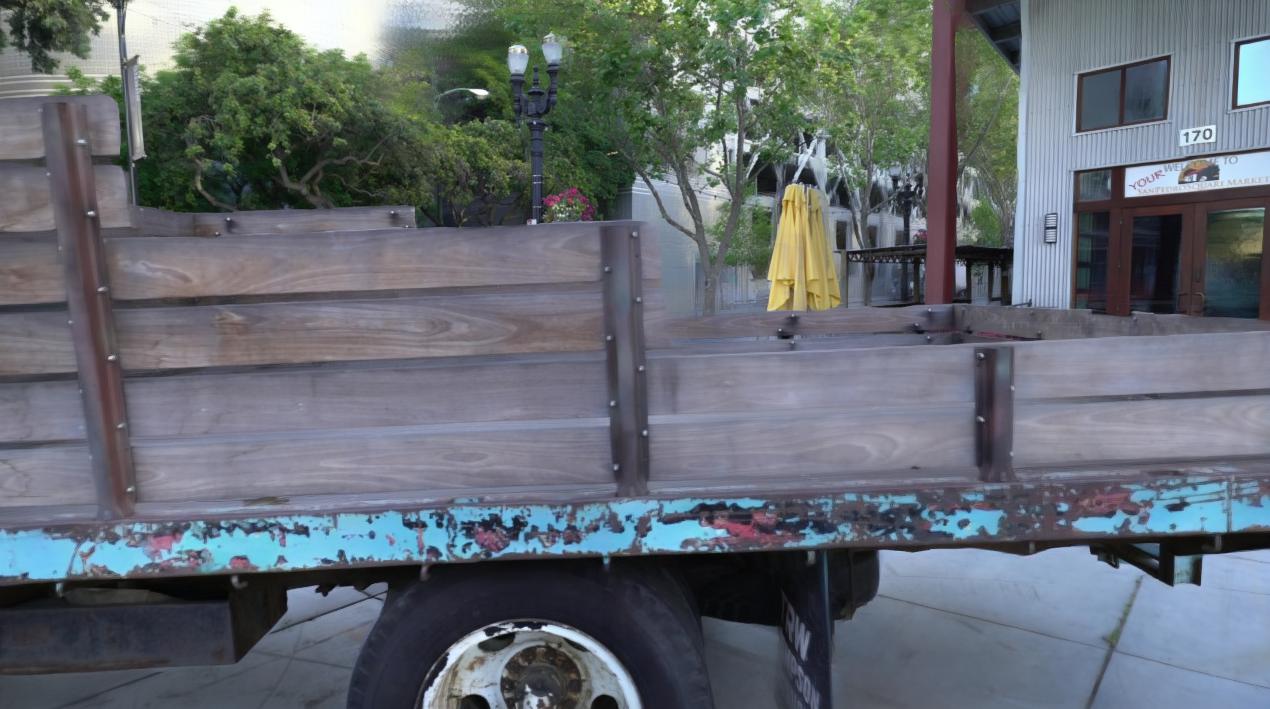} &
	\includegraphics[width=.19\linewidth, height=.126\linewidth]{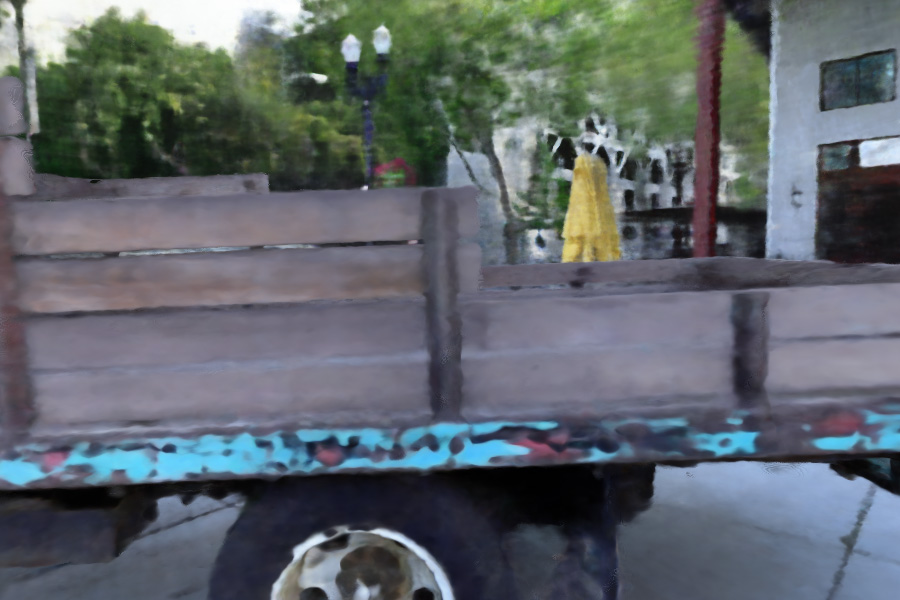} &
	\includegraphics[width=.19\linewidth, height=.126\linewidth]{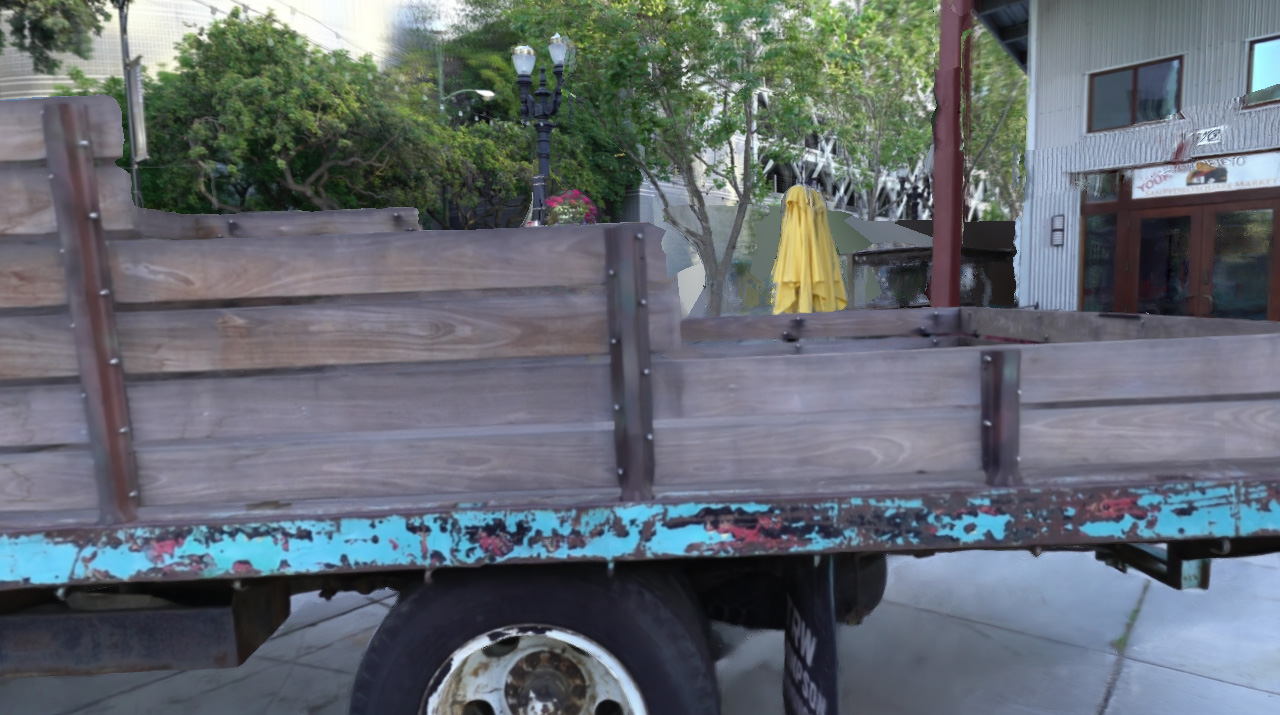} &
	\includegraphics[width=.19\linewidth, height=.126\linewidth]{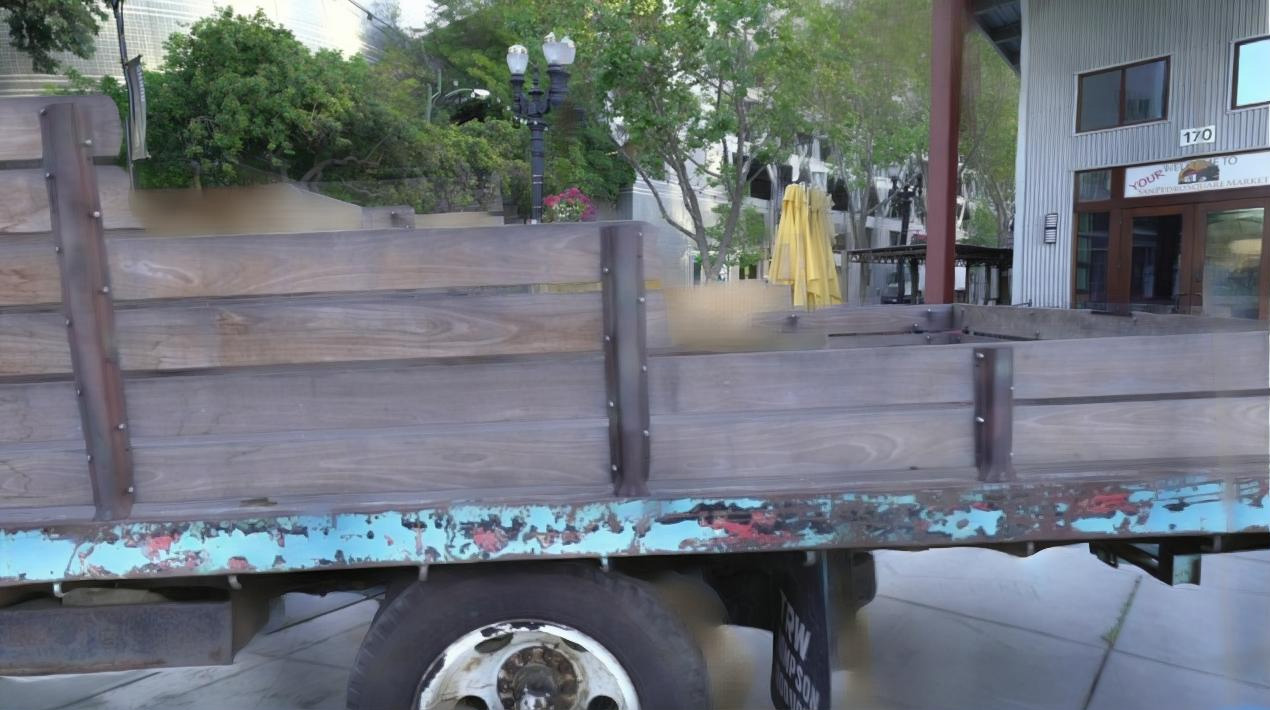} &
	\includegraphics[width=.19\linewidth, height=.126\linewidth]{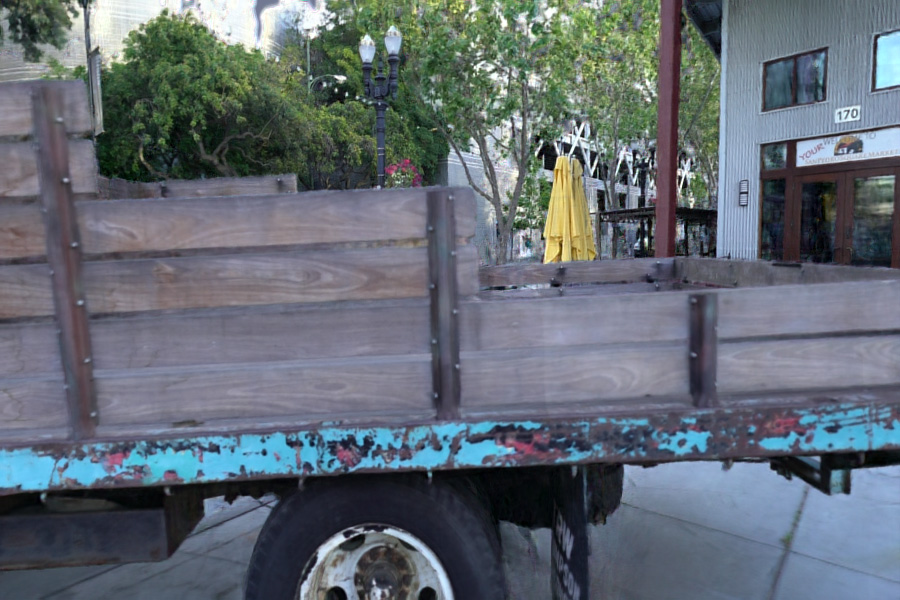}\\
\cite{riegler2020stable} &  \cite{zhang2020nerfplusplus} & \cite{hedman2018deep} & \cite{riegler2020free} &  Ours 
\end{tabular}
\caption{
\label{fig:comparisons}
\CR{Novel views that do not exist in the input dataset.} Left to right: Stable-View Synthesis, NERF++, Deep Blending, Free-view Synthesis and our method.
\vspace{-.3cm}
}
\end{figure*}

\section{Evaluation}
\label{sec:eval}

We compare our IBR method with previous IBR and recent neural rendering methods; the most meaningful comparisons are qualitative visual inspections of the videos provided in the supplemental web pages. We also provide quantitative comparisons and an ablation study analyzing the effect of each component of our method.

\subsection{IBR and Neural Rendering}
\label{sec:ibr}

We compare our algorithm with two baselines: a mesh textured directly from multi-view stereo and a per-pixel ULR algorithm ~\cite{sibr2020}. The textured mesh was generated using RealityCapture ~\cite{reality2016capture} for all scenes from the DeepBlending database and using a direct texturing method by blending using ULR weights for Stairs and Salon~\cite{sibr2020}; the figure for this comparison is presented in the supplemental.

Our main comparisons are with state-of-the-art neural renderers: Deep Blending (DB) \cite{hedman2018deep}, Free View Synthesis (FVS) \cite{riegler2020free}, Stable View Synthesis (SVS) \cite{riegler2020stable} and NeRF++~\cite{zhang2020nerfplusplus}. Standard NeRF implementations were unable to treat our scenes, for reasons explained by Zhang et al.~\cite{zhang2020nerfplusplus}.
The comparisons are best appreciated in our supplemental web page where videos can be viewed side-by-side.

Overall, our method has better overall visual quality for all scenes and all methods; only NeRF++ has comparable quality in some cases, and performs better for the thin structures in the Stairs scene. However, Fig.~\ref{fig:comparisons} shows the clear benefit of our per-view optimization in scenes with vegetation: our method produces much sharper details in such regions compared to all previous solutions (Hugo, Tree, and Street scenes). Our method also removes some of the artifacts in thin structures (rails in Stairs) and can recover them in much sharper detail when they are not reconstructed perfectly (chair and table in the Ponche scene), due to the combined effect of latent features and depth optimization. 
NeRF++ performs slightly better visually for Stairs, removing all over-reconstruction artifacts.

\begin{figure*}[!ht]
	\centering
	\setlength{\tabcolsep}{1pt}
	\begin{tabular}{ccccc}
		\includegraphics[width=.19\linewidth, height=.126\linewidth]{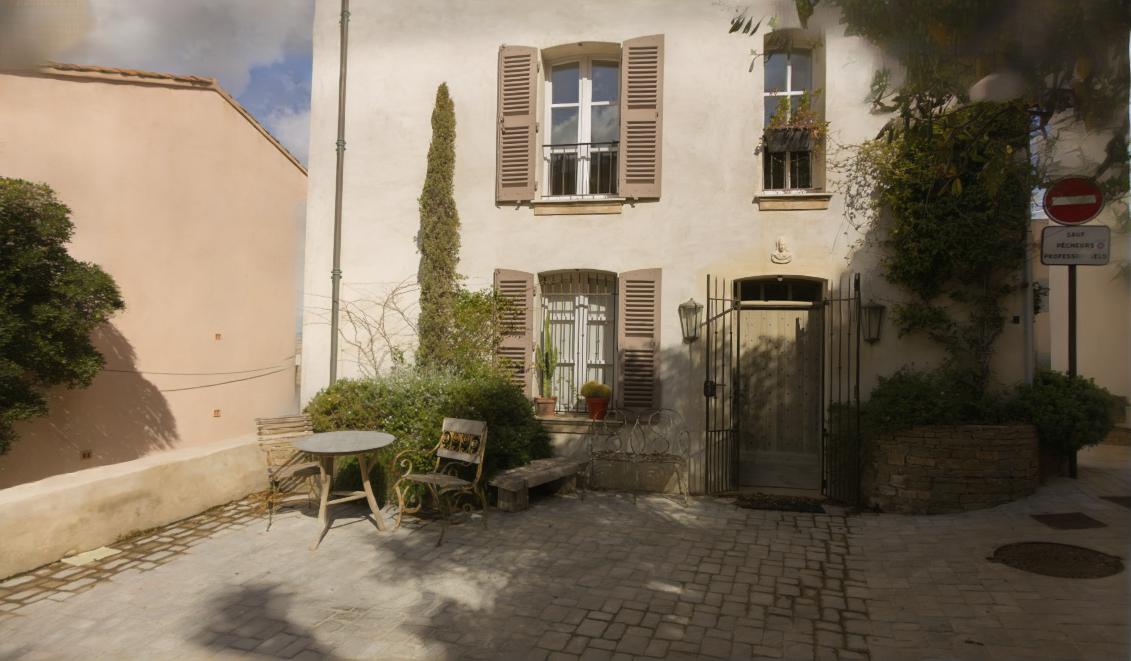} &
		\includegraphics[width=.19\linewidth, height=.126\linewidth]{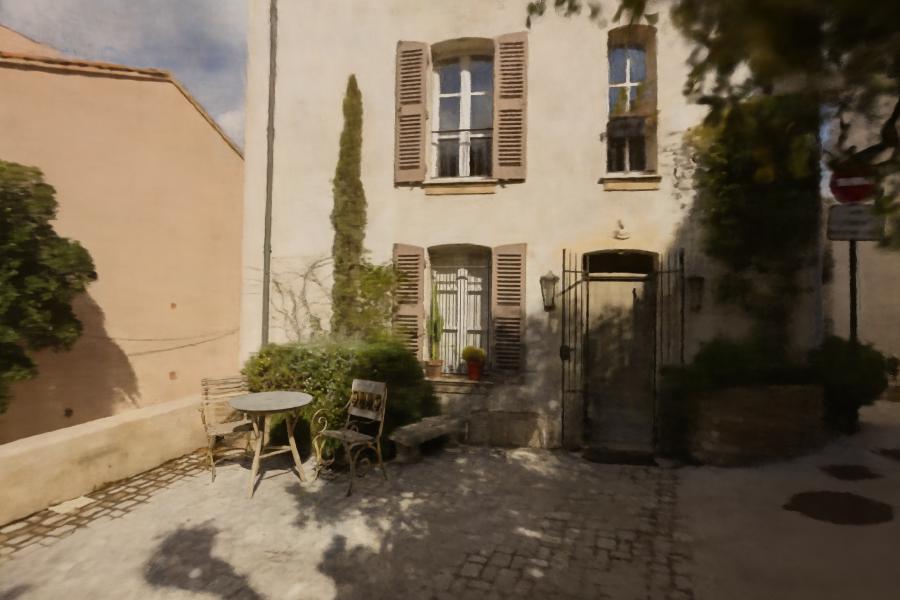}  &
		\includegraphics[width=.19\linewidth, height=.126\linewidth]{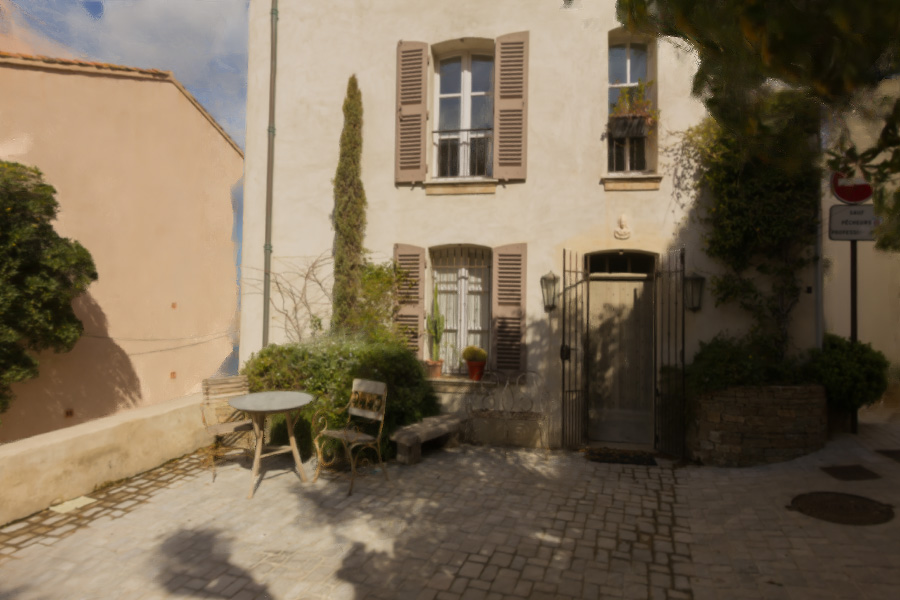} &
		\includegraphics[width=.19\linewidth, height=.126\linewidth]{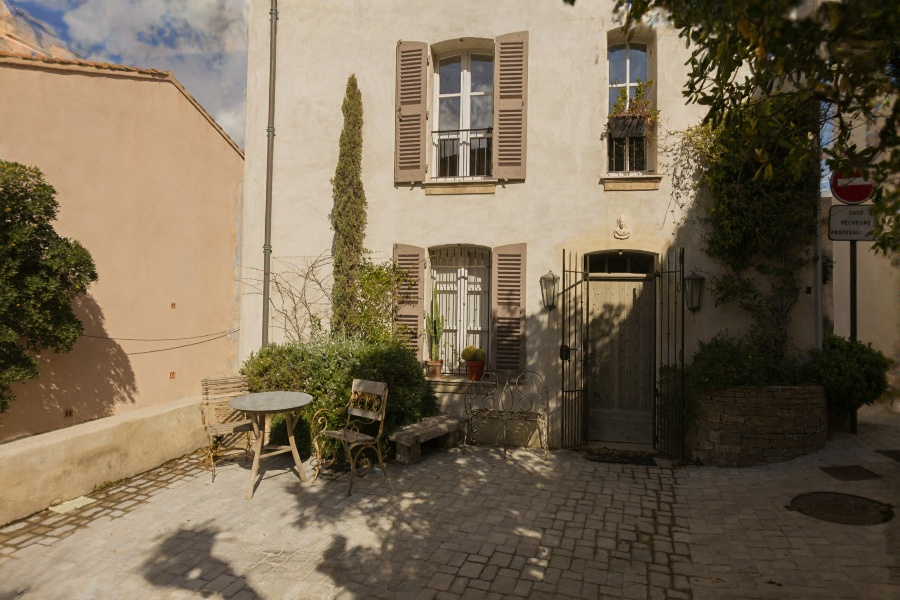}  &
		\includegraphics[width=.19\linewidth, height=.126\linewidth]{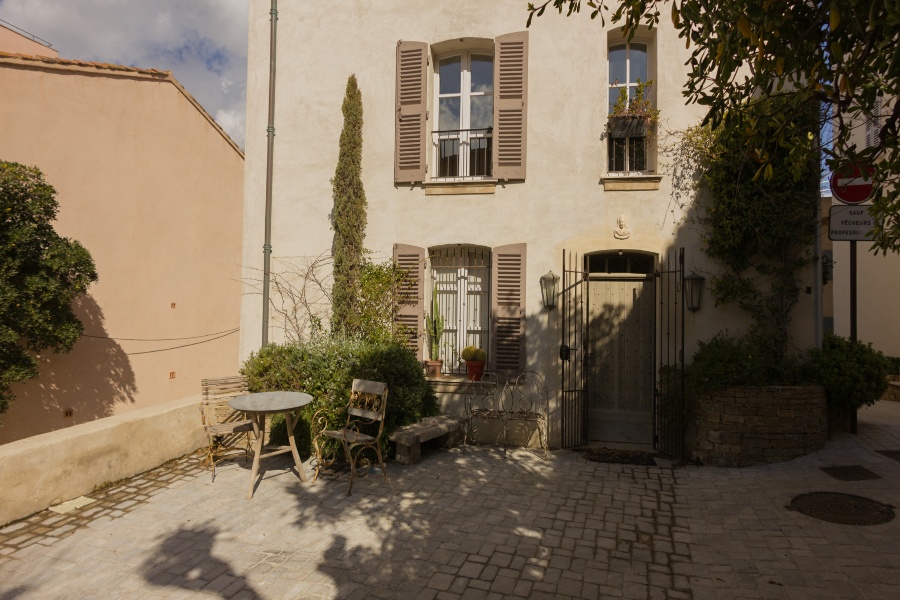} \\
		
		\includegraphics[width=.19\linewidth, height=.126\linewidth]{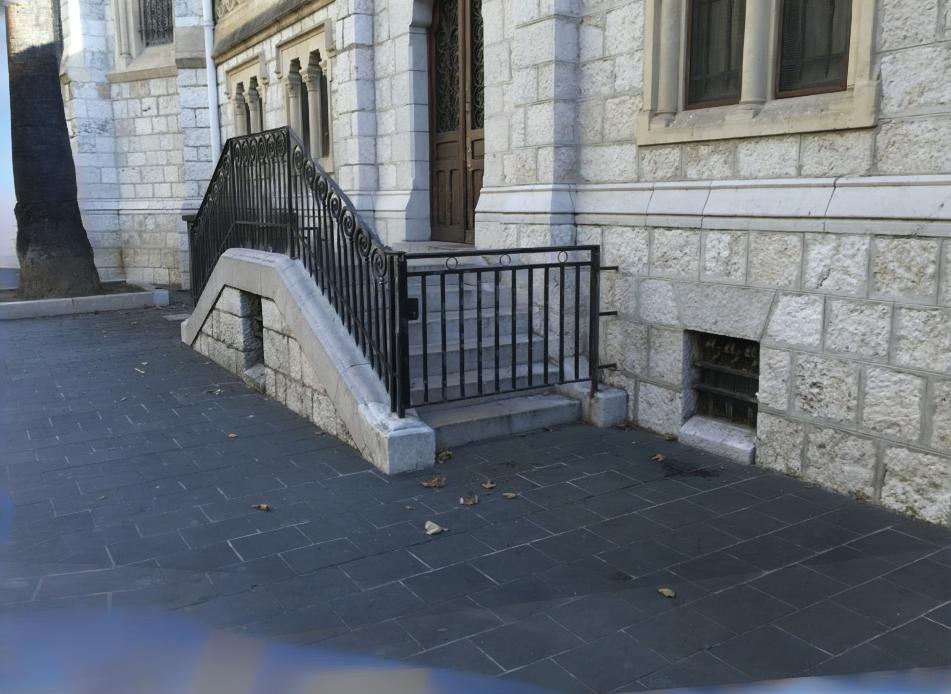} &
		\includegraphics[width=.19\linewidth, height=.126\linewidth]{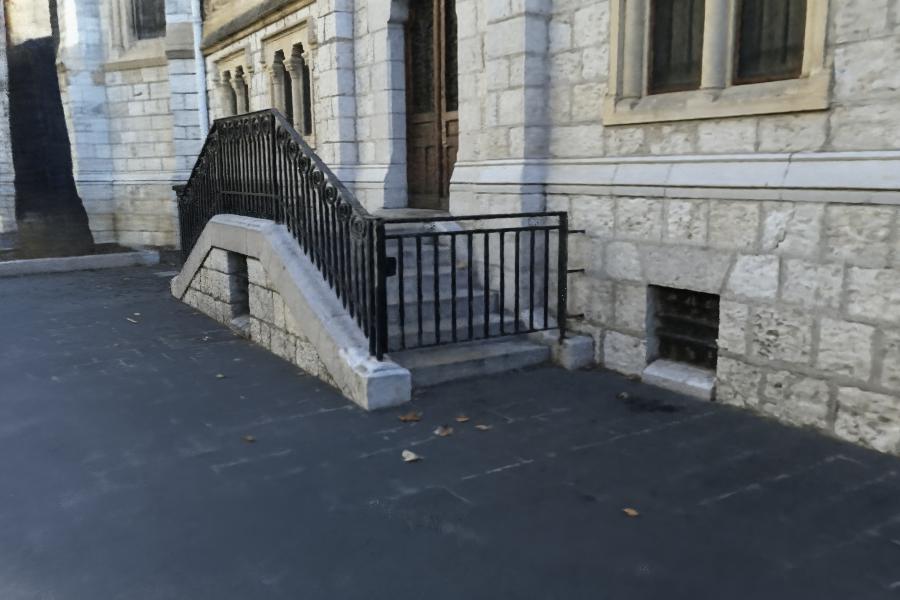}  &
		\includegraphics[width=.19\linewidth, height=.126\linewidth]{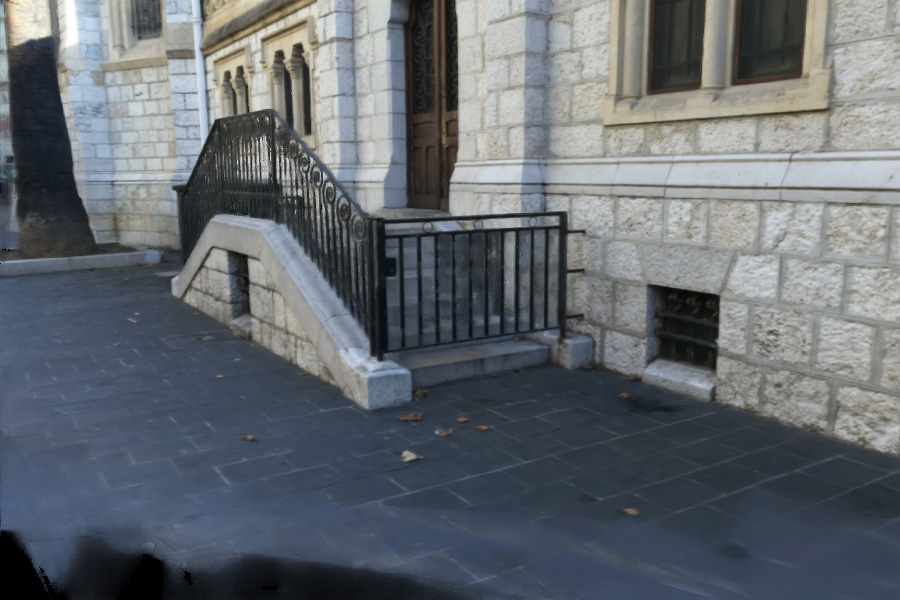}&
		\includegraphics[width=.19\linewidth, height=.126\linewidth]{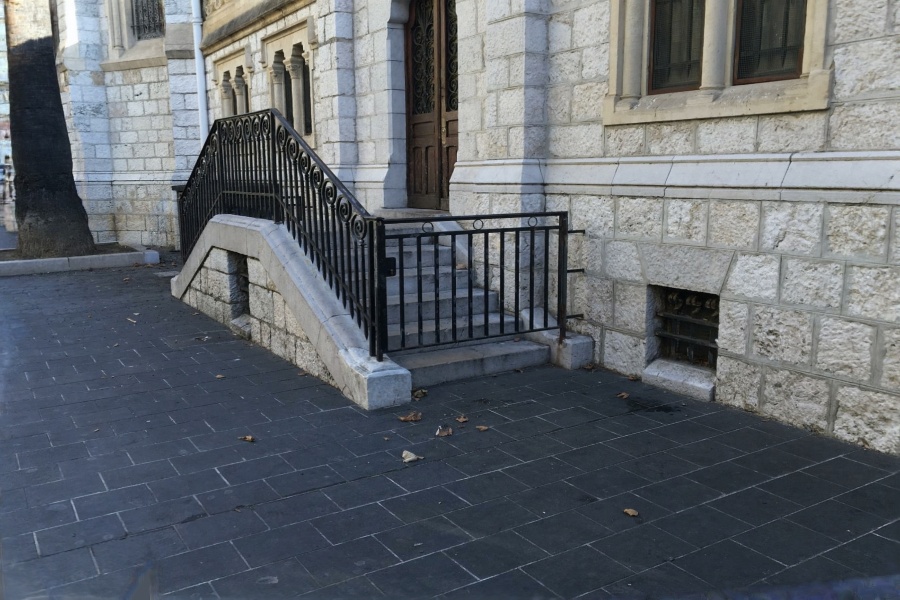} &
		\includegraphics[width=.19\linewidth, height=.126\linewidth]{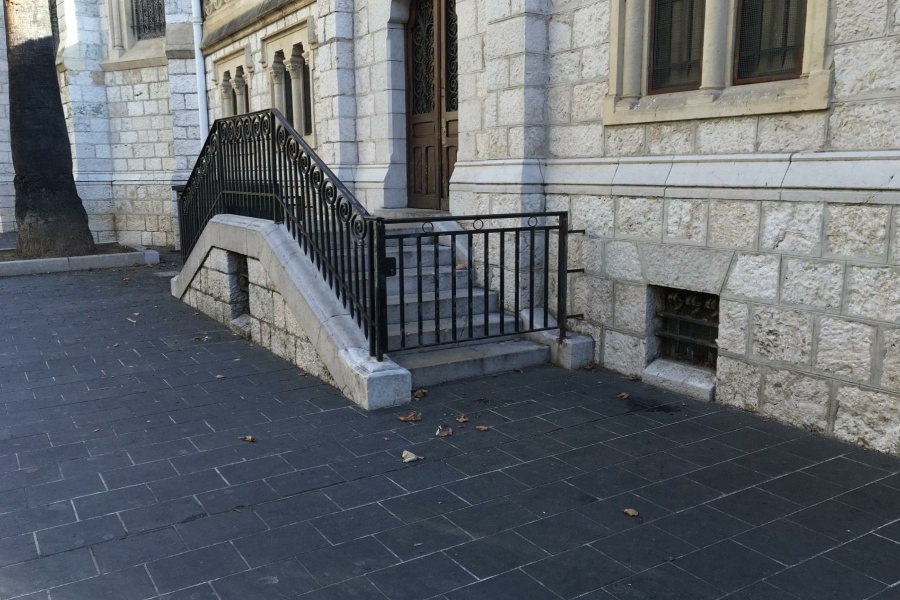} \\
		
		\includegraphics[width=.19\linewidth, height=.126\linewidth]{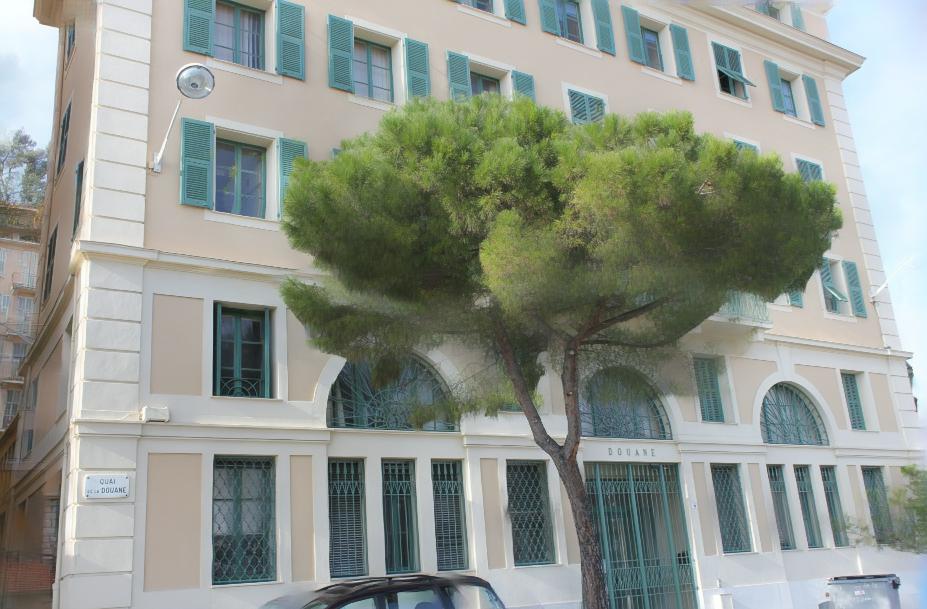} &
		\includegraphics[width=.19\linewidth, height=.126\linewidth]{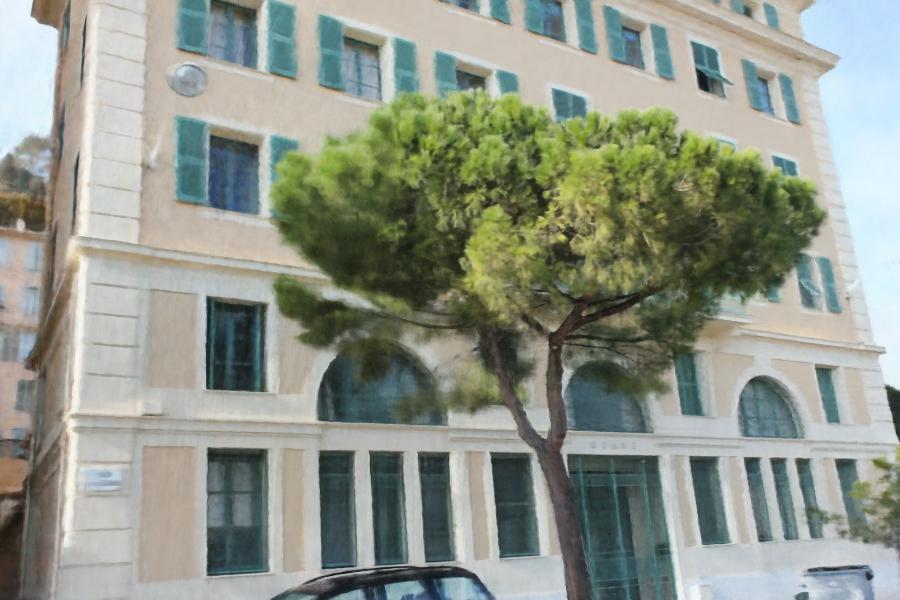}  &
		\includegraphics[width=.19\linewidth, height=.126\linewidth]{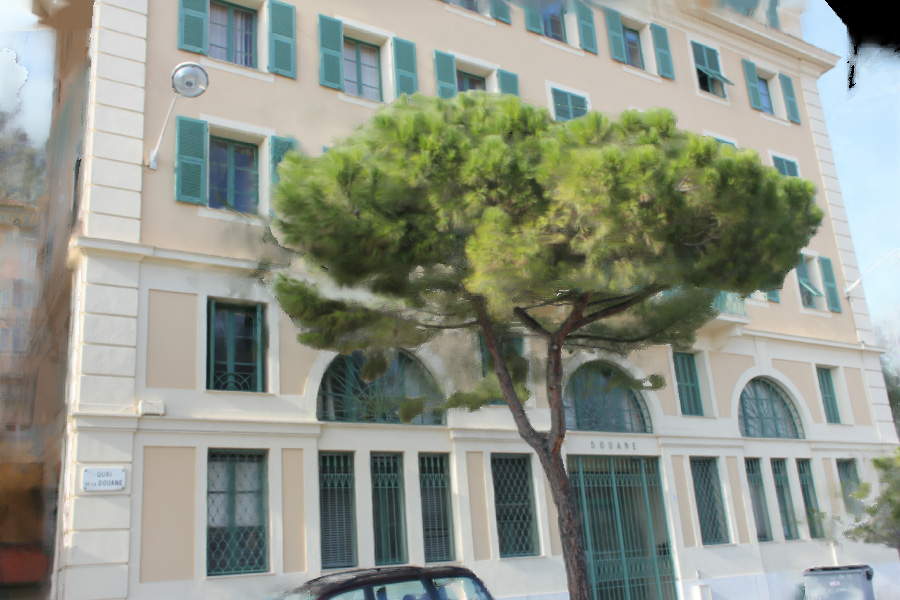} &
		\includegraphics[width=.19\linewidth, height=.126\linewidth]{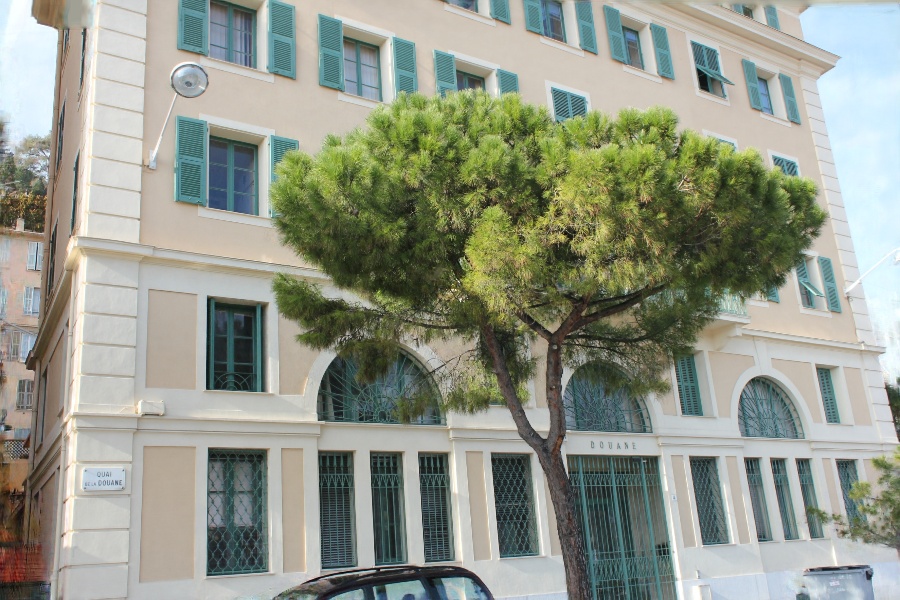} &
		\includegraphics[width=.19\linewidth, height=.126\linewidth]{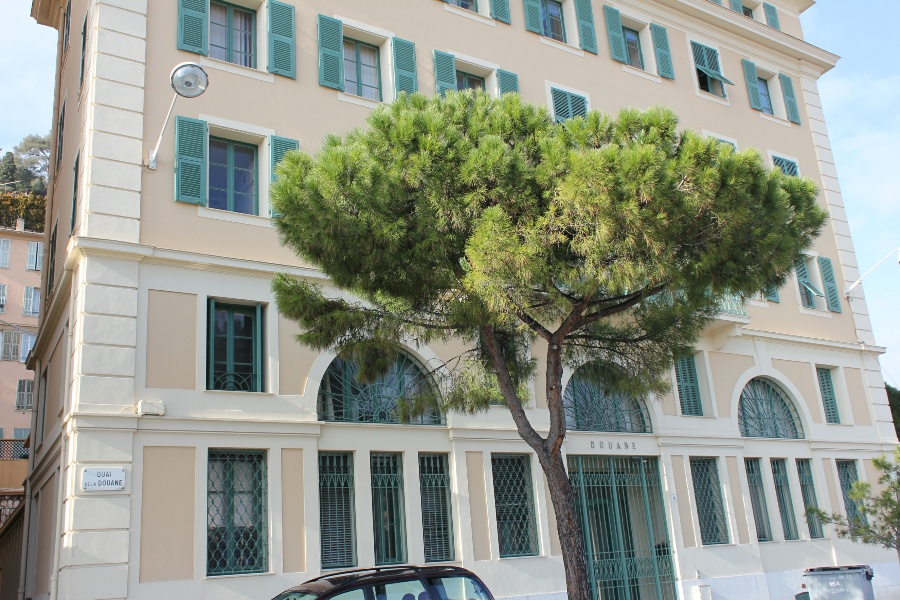}\\
		
		\includegraphics[width=.19\linewidth, height=.126\linewidth]{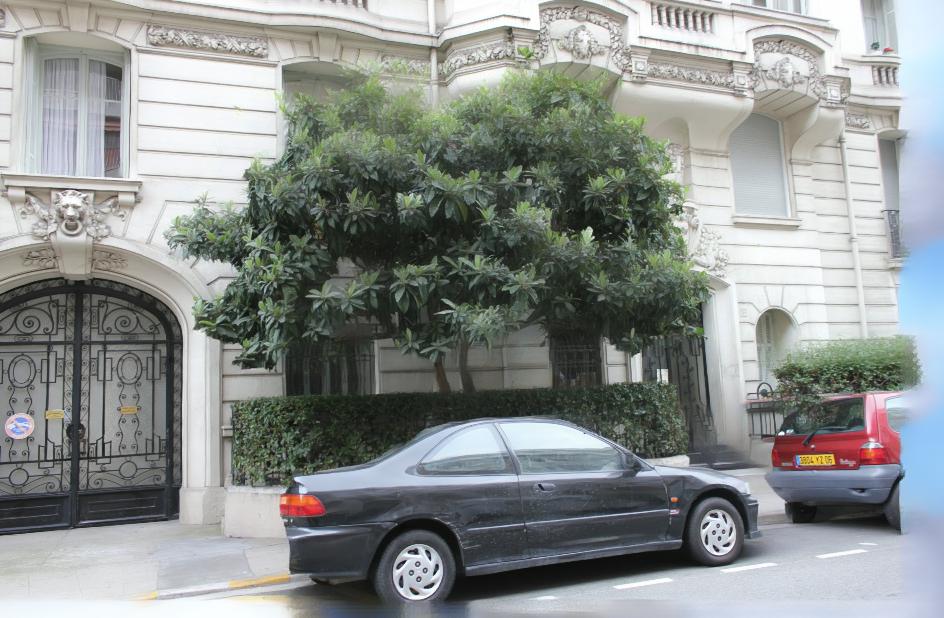} &
		\includegraphics[width=.19\linewidth, height=.126\linewidth]{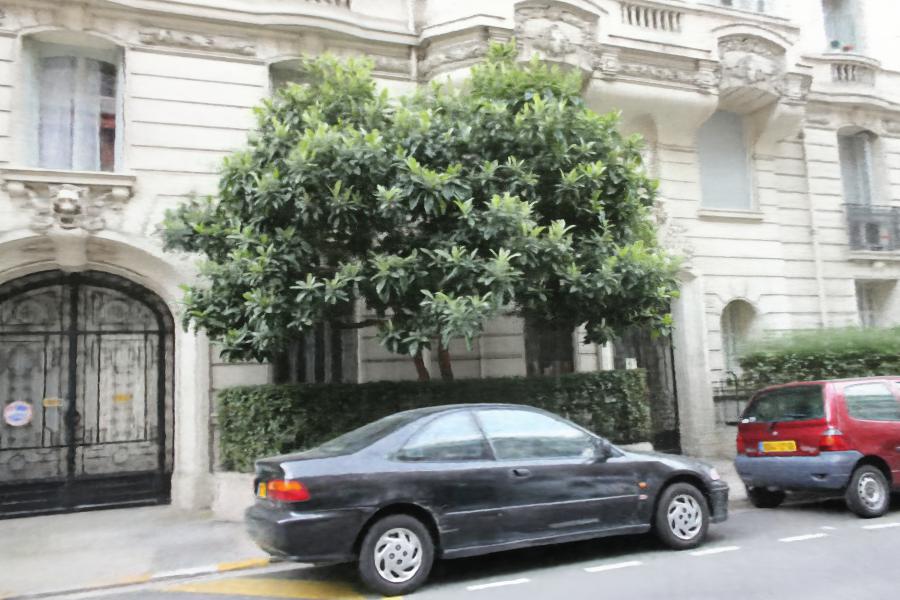} &
		\includegraphics[width=.19\linewidth, height=.126\linewidth]{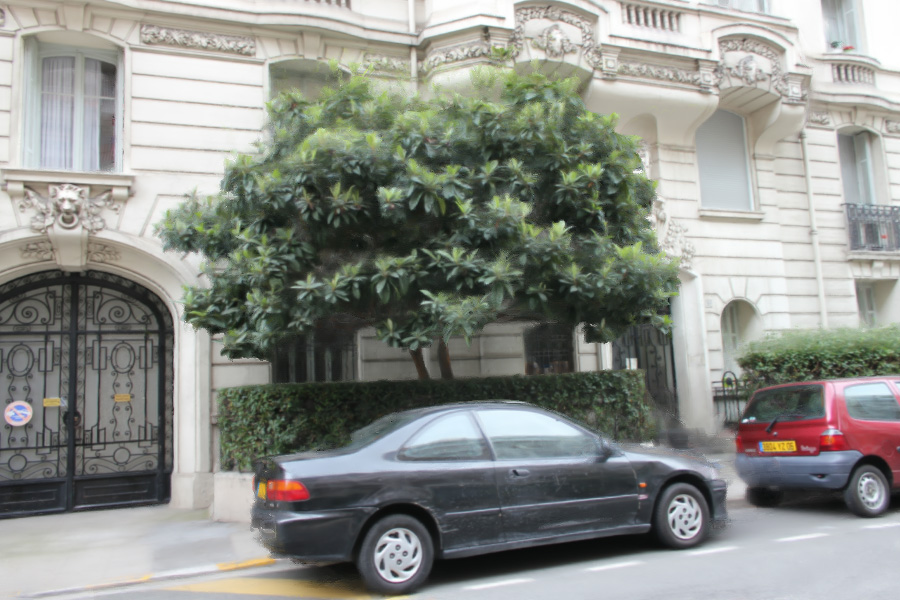} &
		\includegraphics[width=.19\linewidth, height=.126\linewidth]{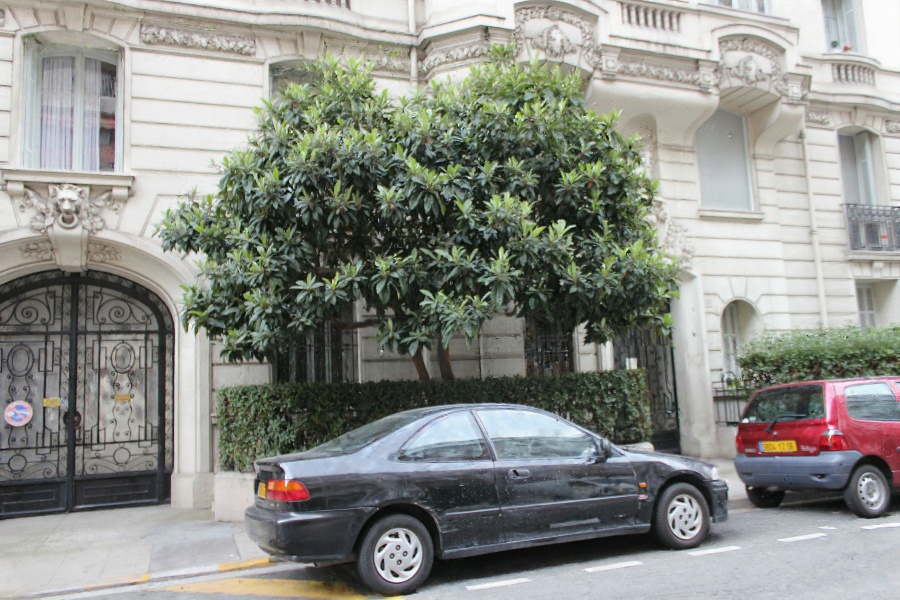} &
		\includegraphics[width=.19\linewidth, height=.126\linewidth]{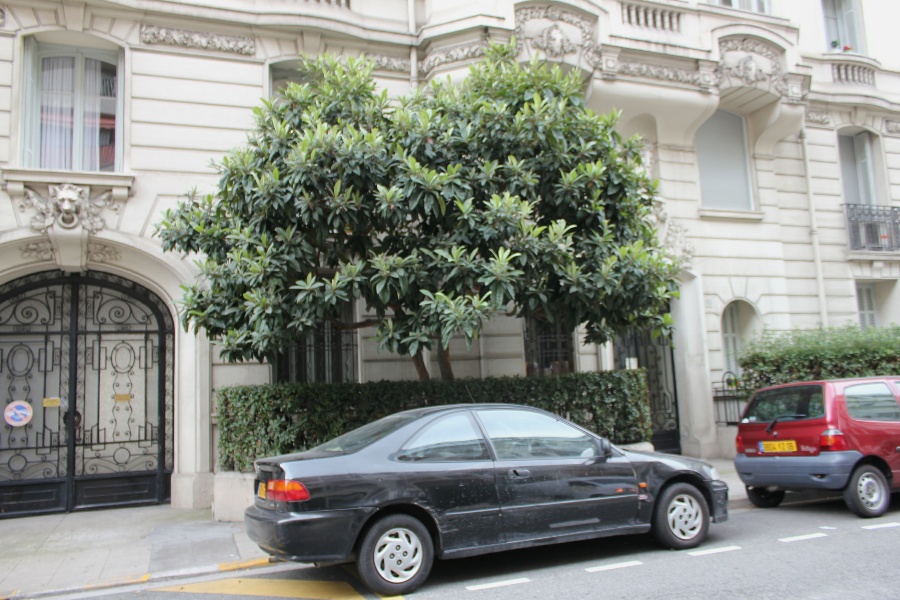}\\
		
		\includegraphics[width=.19\linewidth, height=.126\linewidth]{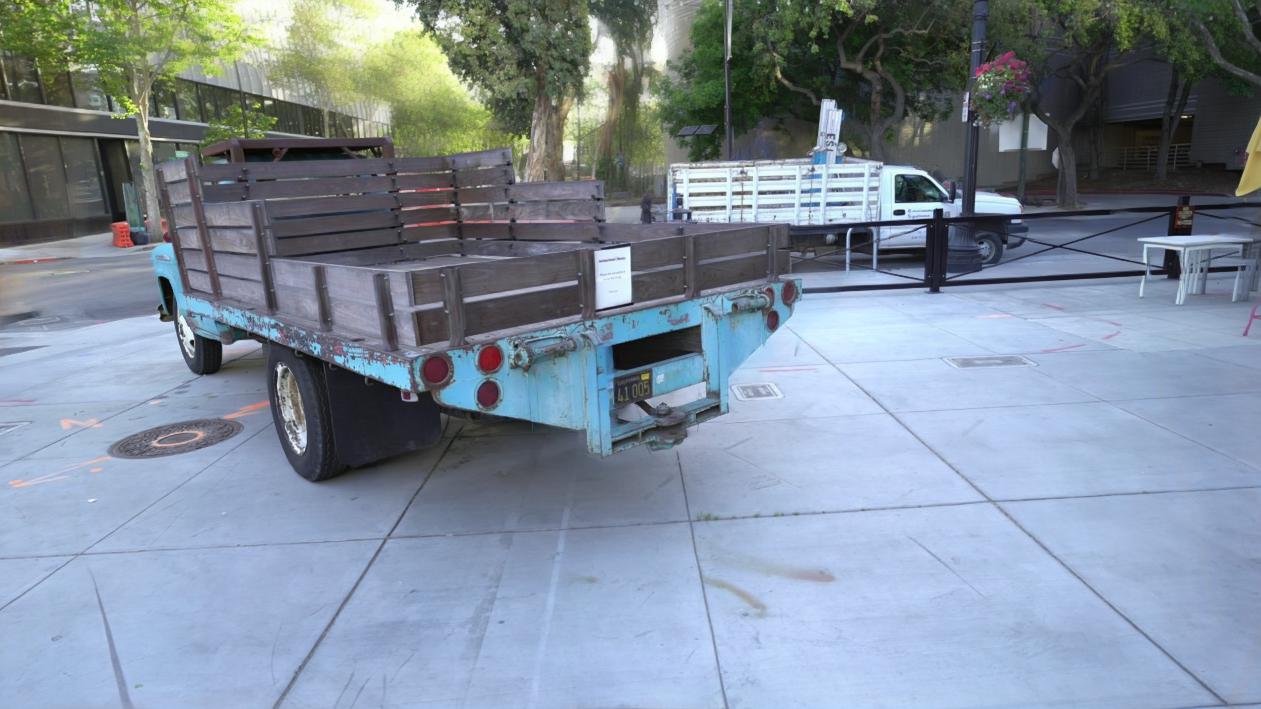} &
		\includegraphics[width=.19\linewidth, height=.126\linewidth]{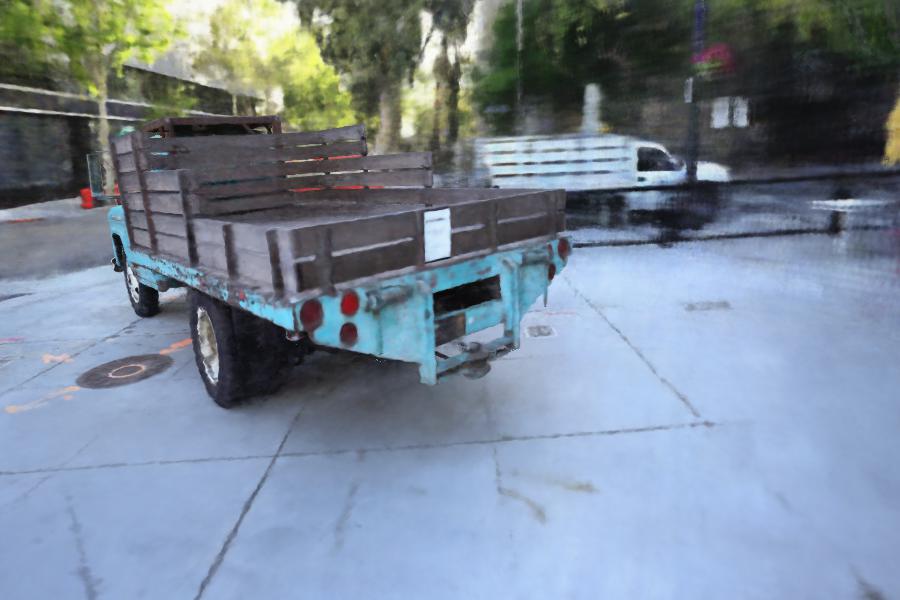} &
		\includegraphics[width=.19\linewidth, height=.126\linewidth]{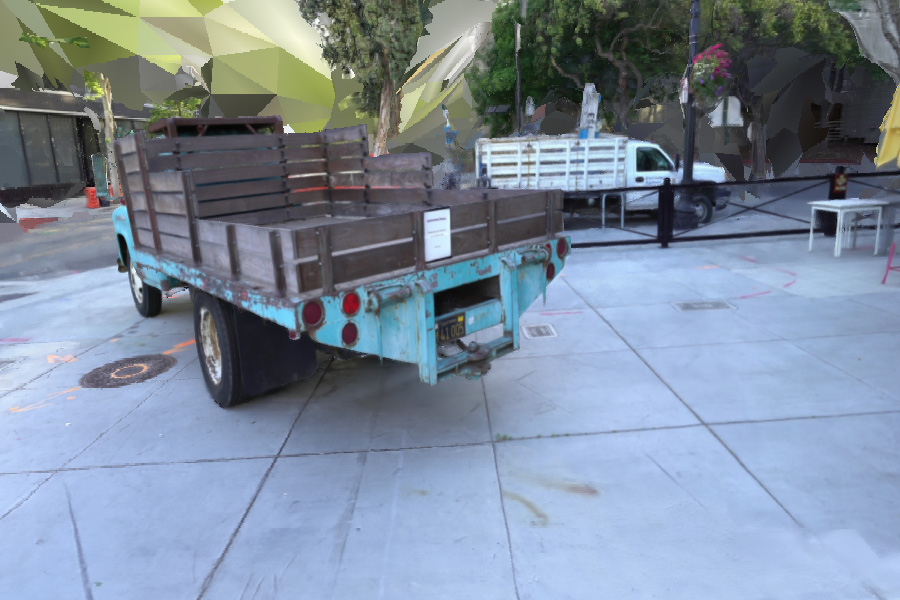} &
		\includegraphics[width=.19\linewidth, height=.126\linewidth]{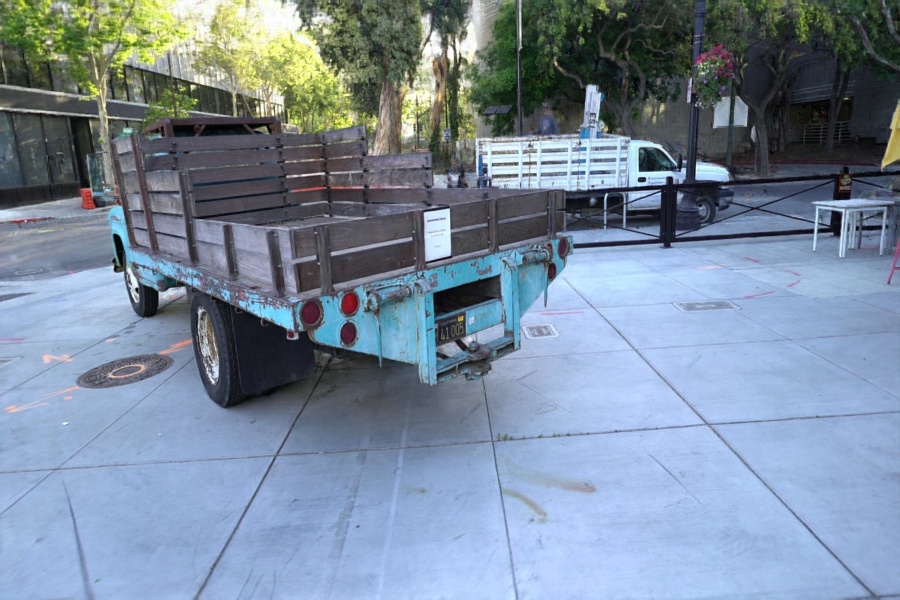} &
		\includegraphics[width=.19\linewidth, height=.126\linewidth]{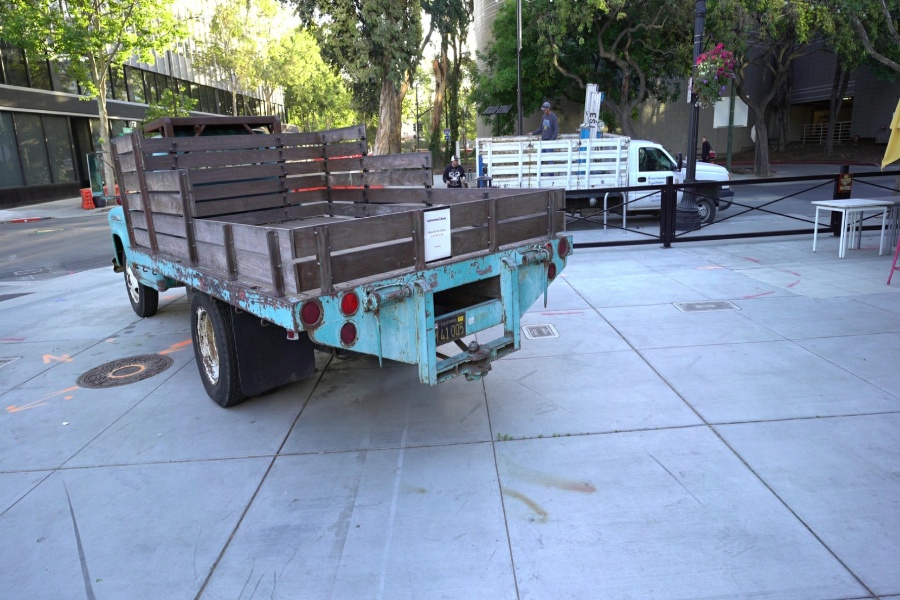}\\
		\cite{riegler2020free} &  \cite{zhang2020nerfplusplus} & \cite{hedman2018deep} & Ours &  Ground Truth 
	\end{tabular}
	\caption{
		\label{fig:comparisons-loo}
		\CR{Renderings of left-out input views. Left to right: Free-View Synthesis, NERF++, Deep Blending, Ours and Ground Truth. Our method renders input views with almost no artifacts.}
	}
\end{figure*}

We perform a quantitative leave-one-out comparison; the results are shown in Table~\ref{tab:view_synthesis_loo} \CR{and Fig.~\ref{fig:comparisons-loo}}. We used the authors' implementation of each method, using COLMAP SfM and the RealityCapture MVS mesh (except for Stairs where Colmap MVS is used); the SVS \cite{riegler2020stable} code ran out of GPU memory for this test. All views are used for SfM and MVS, and for training for NeRF++ and our method, but we leave out the target view for rendering for each method except NeRF++ where this is not possible. NeRF++ trained for 48h on average for each scene. NeRF++ thus has an ``advantage'' over other methods, since the view being rendered is not actually left out. We present results for three error metrics, PSNR, LPIPS and DSSIM. Leave-one-out quantitative comparisons are not particularly informative for IBR: The ranking of methods changes according to the metric, and can be influenced by training (e.g., FVS uses LPIPS in the training loss, and is thus better for this metric, but not for others). The fact that ULR is second best in many cases shows that the metrics do not correspond well with visual perception since ULR shows many visual artifacts compared to the others. Nonetheless, our method has the best scores for all metrics and all scenes.

\begin{table*}[!ht]
\small
	\centering
	\caption{\label{tab:view_synthesis_loo} 
		\textbf{Leave-one-out view-synthesis quantitative evaluation on real test scenes.}
	}
	\begin{tabular}{|l|ccccc|ccccc|ccccc|}
		\toprule
		\multicolumn{1}{|l|}{\bfseries Metric}   & \multicolumn{5}{c|}{\bfseries SSIM $\uparrow$} &   \multicolumn{5}{c|}{\bfseries PSNR $\uparrow$}   & \multicolumn{5}{c|}{\bfseries LPIPS $\downarrow$} \\
		Method                               & ULR & DB & FVS & N++  & Ours & ULR & DB & FVS & N++  & Ours & ULR &DB  & FVS & N++  & Ours  \\ \midrule
		Street                                   
		&                       0.67  & \cellcolor{yellow!40} 0.69  &                       0.64  & \cellcolor{orange!40} 0.77  & \cellcolor{red!40} 0.93  
		&                      17.84  & \cellcolor{yellow!40}19.91  &                      17.80  & \cellcolor{orange!40}25.23  & \cellcolor{red!40}28.67
		& \cellcolor{orange!40} 0.245 &                       0.27  & \cellcolor{yellow!40}0.249  &                       0.25  & \cellcolor{red!40} 0.06 \\
		Hugo 
		& \cellcolor{yellow!40} 0.77  &                       0.74 & \cellcolor{orange!40}0.78  &                       0.75 & \cellcolor{red!40} 0.92
		&                      21.7   & \cellcolor{yellow!40}22.9  &                      22.6  & \cellcolor{orange!40}25.2  & \cellcolor{red!40}27.25 
		& \cellcolor{yellow!40} 0.164 &                       0.23 & \cellcolor{orange!40}0.162 &                       0.29 & \cellcolor{red!40} 0.07 \\
		Tree
		&                       0.756 & \cellcolor{yellow!40} 0.76 &                       0.754 & \cellcolor{orange!40} 0.78 & \cellcolor{red!40} 0.94
		&                      20.7   & \cellcolor{yellow!40}22.5  &                      22.1   & \cellcolor{orange!40}27.3  & \cellcolor{red!40}30.4
		& \cellcolor{yellow!40} 0.18  &                       0.22 & \cellcolor{orange!40} 0.17  &                       0.25 & \cellcolor{red!40} 0.05    \\ 
		Ponche
		& \cellcolor{orange!40} 0.78 &                       0.7701 &                        0.72 & \cellcolor{yellow!40} 0.773 & \cellcolor{red!40} 0.92        
		&                      23.5  & \cellcolor{yellow!40}23.7    &                      21.5   & \cellcolor{orange!40}27.5   & \cellcolor{red!40} 29.4
		& \cellcolor{orange!40} 0.21 &                       0.27   & \cellcolor{yellow!40} 0.24  &                       0.306 & \cellcolor{red!40} 0.09\\
		Museum
		&       0.7501        &       \cellcolor{yellow!40}0.76        &      0.758     &   \cellcolor{orange!40}   0.78        &     \cellcolor{red!40}  0.95   
		&       20.9        &       \cellcolor{yellow!40}23.02       &       22.1        &     \cellcolor{orange!40}  25.7        &      \cellcolor{red!40} 30.6   
		& \cellcolor{yellow!40} 0.19           &       0.23        &     \cellcolor{orange!40}  0.16        &       0.24        &    \cellcolor{red!40}  0.04    \\
		Stairs
		&                       0.807 & \cellcolor{yellow!40} 0.821 &       0.79        & \cellcolor{orange!40}   0.83         & \cellcolor{red!40}  0.94   
		&                      20.13  & \cellcolor{yellow!40}22.04  &       21.33        & \cellcolor{orange!40} 28.88         & \cellcolor{red!40} 30.79          
		&  \cellcolor{orange!40} 0.16 &                       0.19  &     \cellcolor{yellow!40}  0.17        &       0.22         & \cellcolor{red!40}    0.05    \\
        Truck
		&                 \cellcolor{yellow!40}       0.76 &  0.64 &       \cellcolor{orange!40}0.78        &   0.703         & \cellcolor{red!40}  0.88   
		&            \cellcolor{yellow!40}           21.4  & 18.4  &       20.97        & \cellcolor{orange!40} 23.08        & \cellcolor{red!40} 25.3        
		&  \cellcolor{yellow!40} 0.18 &  0.27  &     \cellcolor{orange!40}  0.179        &       0.36         & \cellcolor{red!40}    0.11    \\  \bottomrule
	\end{tabular}
\end{table*}

\subsection{Ablations}
\label{sec:ablations}

\begin{figure}[!ht]
\centering
	\includegraphics[width=1.0\linewidth]{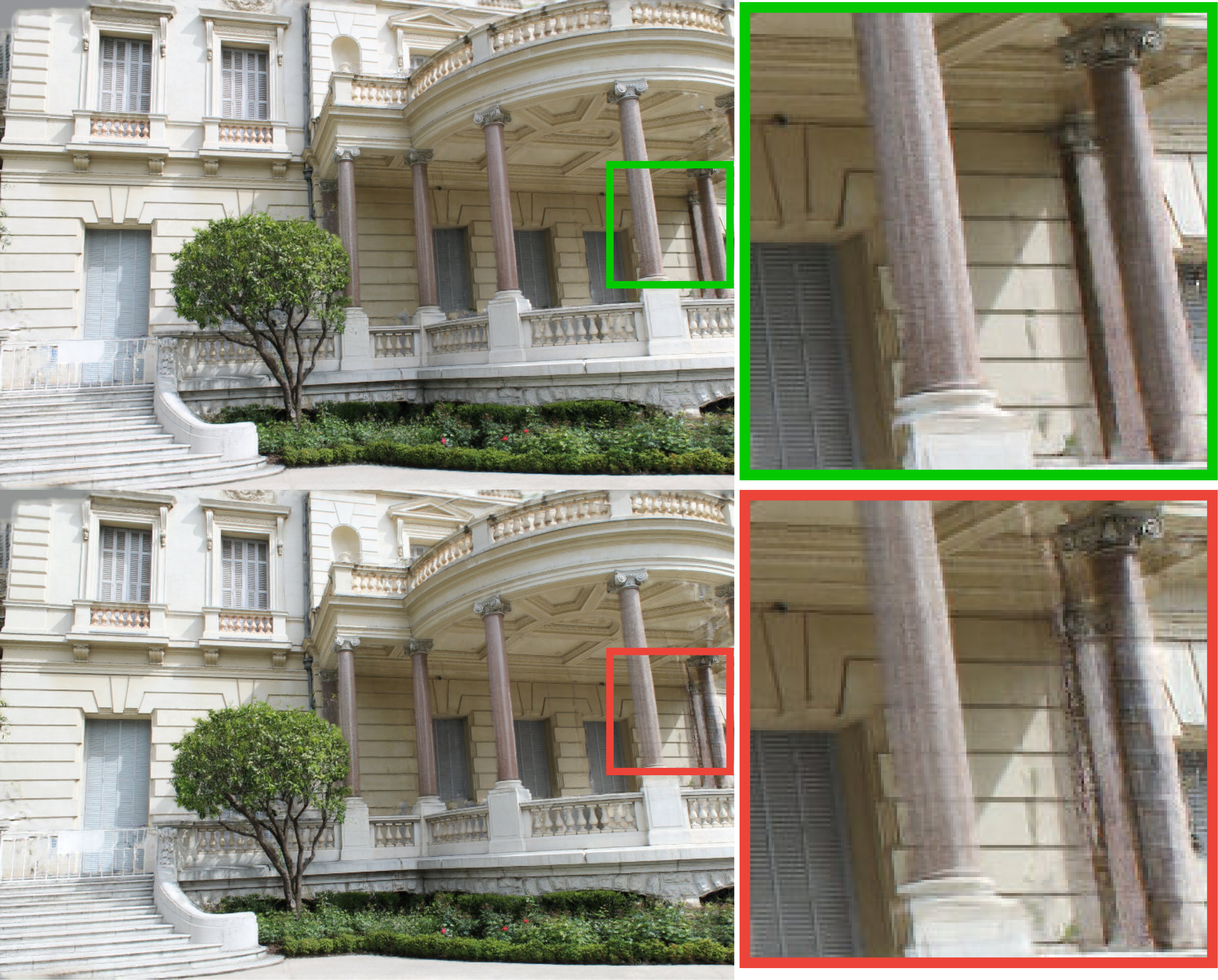}
	\caption{
		\label{fig:depth_test_results}
		The probabilistic depth test successfully resolves visibility between different viewpoints as demonstrated in the Museum scene by eliminating transparent regions. Top row: with the depth test. Bottom row: without.
	}
\end{figure}

\begin{figure}[!ht]
	\includegraphics[width=\linewidth]{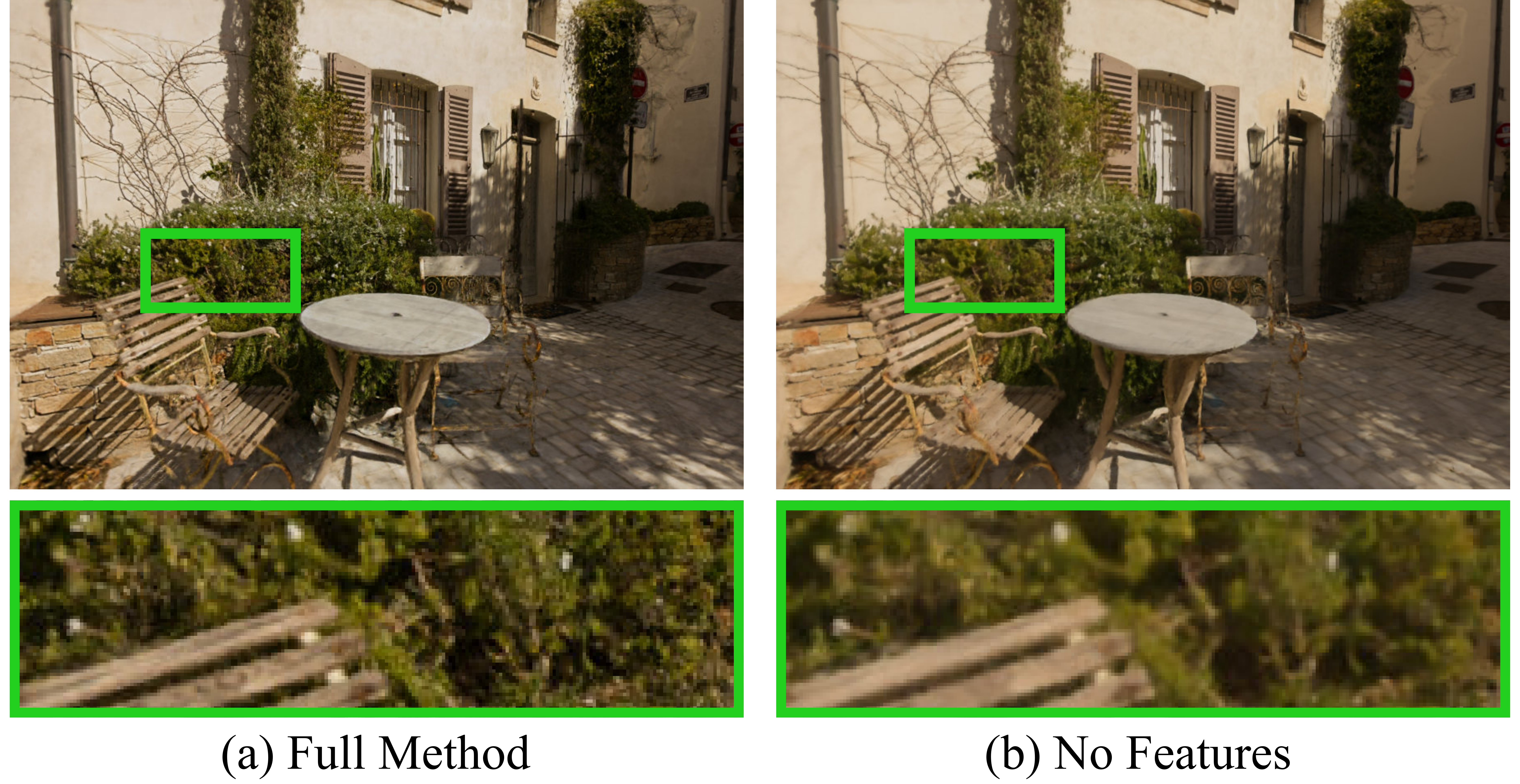}
	\caption{
		\label{fig:ablation_ponche}
Ablation for per-view latent features, which play a very important role in preserving the sharpness of vegetation.
\vspace{-.1cm}
	}
\end{figure}

\begin{figure*}[!ht]
	\includegraphics[width=\linewidth]{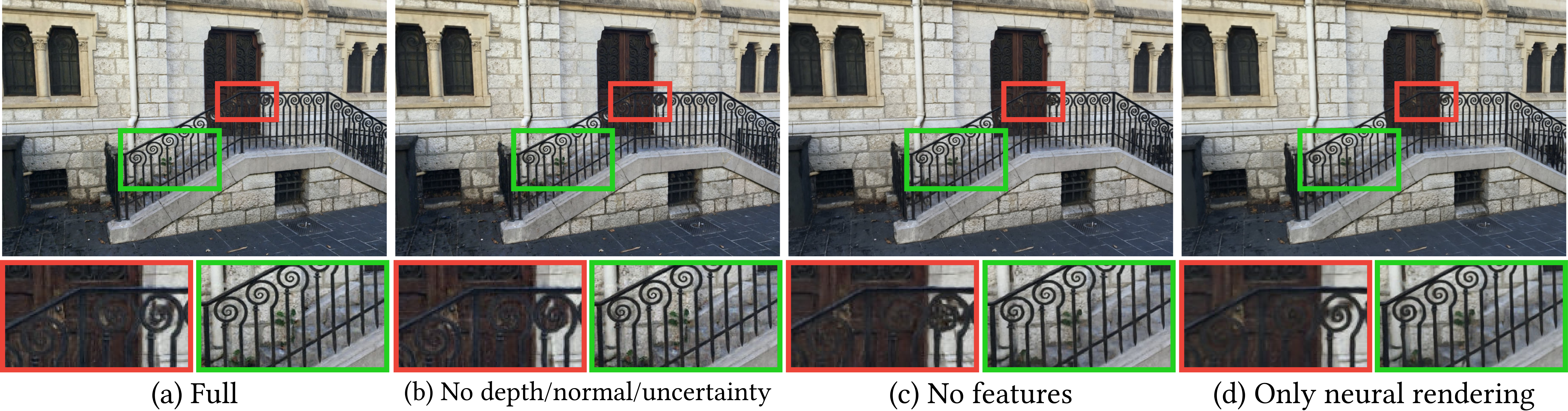}
	\caption{
		\label{fig:ablation_stairs}
Each component of our method helps correct different artifacts. E.g., without features the stairs behind the bars are blurry, without depth/normal/uncertainty the overreconstruction becomes worse. Taken all the components together results in the overall best rendering.
	}
\end{figure*}

We investigate the effect of each of our components through ablation studies in per-scene training. In particular we optimize our network in three different configurations: 1) our full pipeline, 2) without the depth test, 
3) \CR{without the 6-feature latent vector optimization}, 4) without normal/depth/uncertainty optimization and 5) only optimizing the neural renderer. We run the ablation test for 50k iterations. The results on paths of a subset of our test scenes can be seen in the supplemental webpage. All test sets comprise three views that were held-out during optimization.

We first show the importance of the depth test in Fig.~\ref{fig:depth_test_results}. If we disable the test, the soft rasterization results in incorrect ``transparent'' regions.
The reprojected features play a central role in the quality of our results, especially for vegetation. This can be seen in Fig.~\ref{fig:ablation_ponche}.

In Fig.~\ref{fig:ablation_stairs} we examine the relative importance of features and normal/depth/uncertainty in the quality of our results. 
In particular, when depth/normals/uncertainty are not used we notice that the some reconstruction artifacts are more visible in the region marked with a red box (Fig.~\ref{fig:ablation_stairs}(b)). The optimization cannot recover the background information from the over-reconstructed point splats without resizing and moving them. When this freedom is given, the optimization spreads the points so background information is visible. The neural renderer then filters out the over-reconstruction. 
Reprojected features also result in sharper rendering; see Fig.~\ref{fig:ablation_stairs}(c) green box for the railings.

Overall, we see from this study that each component plays a different role, and handles different artifacts. Our method does not ``correct'' depth \emph{per se}; the strength of our method is making these components (projected features, depth etc.) be optimized together to obtain the best overall compromise between correcting geometry reconstruction errors and obtaining sharp rendering in all regions of the synthesized novel view.

\begin{figure}[!h]
\includegraphics[width=\linewidth]{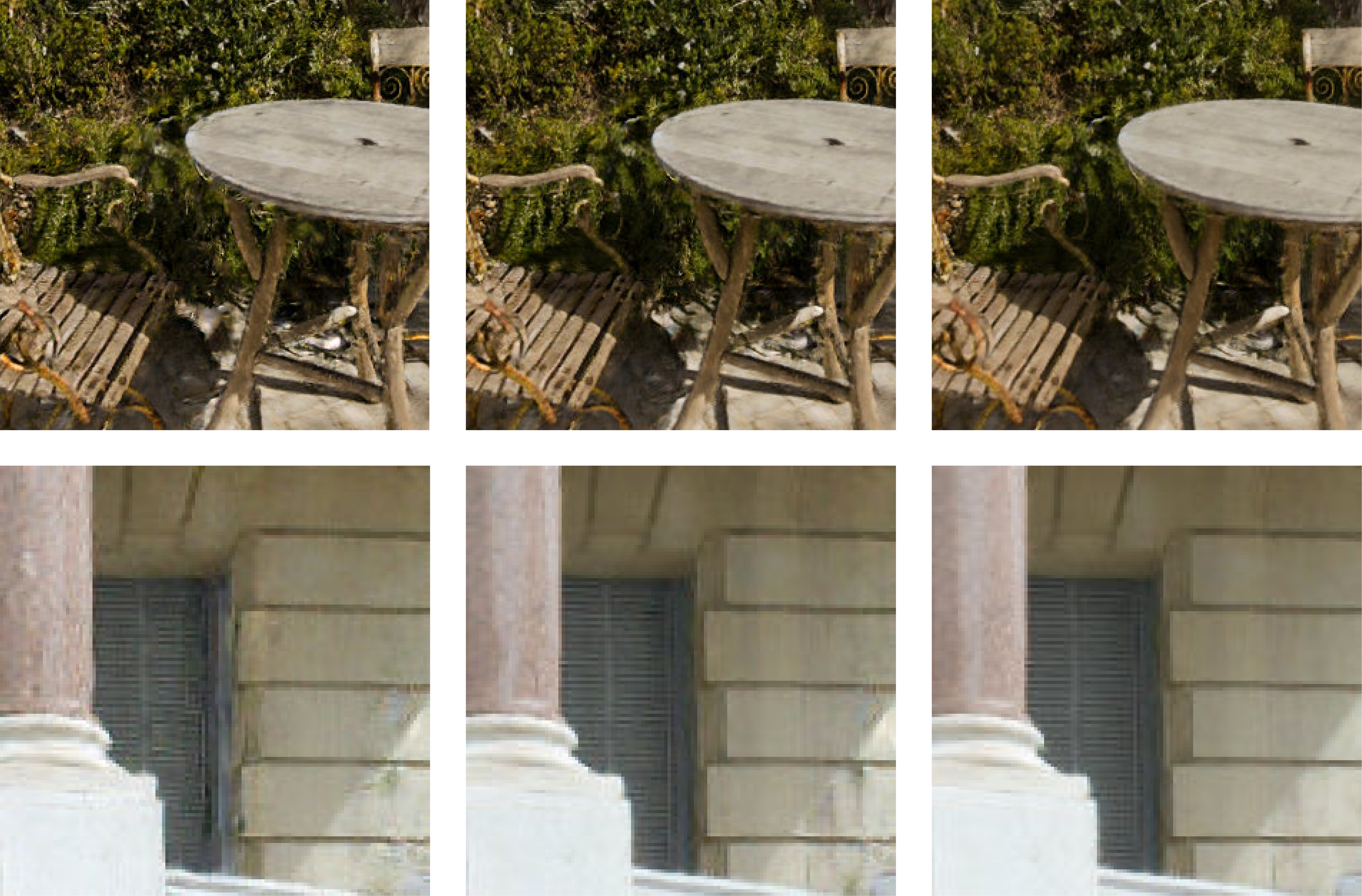}
\caption{
\label{fig:camera-ablation}
Left to right: N=4, 6 and our chosen value of N=9. For the first row $N=4$ is sufficient, but for some scenes (second row) we need 9 views.
}
\end{figure}

\begin{figure}[!h]
	\includegraphics[width=\linewidth]{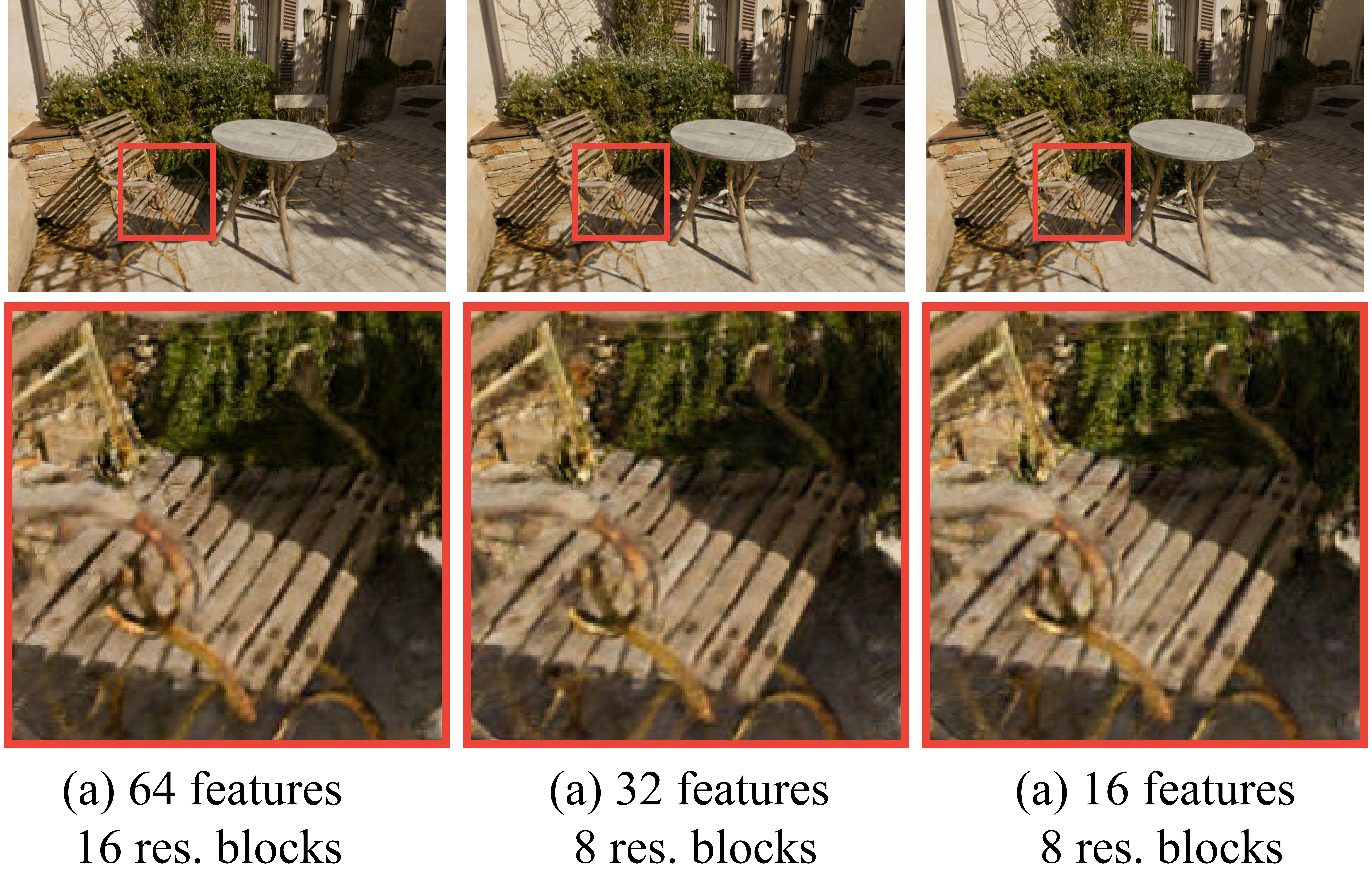}
	\caption{
		\label{fig:ablation_model_size_ponche}
We study the effect of network model size. As we can see, even with a much smaller model than that used in most of our experiments presented, the results are still of high quality.
	\vspace{-.2cm}
	}
\end{figure}
We investigate the effect of network size shown in Fig.~\ref{fig:ablation_model_size_ponche}, by reducing the number of feature and of residual blocks.
We see that even with a much smaller model than that used in most of our experiments presented, the results are still of very high quality.
We also investigate the effect of varying the number of input cameras. Here, the major limitation is memory. 
We explore a trade-off between simplifying the neural renderer as described above and the number of cameras in memory during inference. 
We test in an indoor scene ("Salon") that is challenging for camera selection. The inside-out nature means that we need to increase the number of cameras to achieve good results. The versatility of our neural rendering allows us to explore the number of cameras without the need to retrain the network. 
In the supplemental we show that decreasing the size of the model often does not affect the results substantially and a higher number of cameras used for rendering in the same time/memory budget compensates for the smaller neural renderer. 
In Fig.~\ref{fig:camera-ablation} we show how in the Museum scene quality degrades as we lower the number of cameras, although it is possible to lower the number of cameras with minimal degradation in the quality depending on the scene.

\begin{figure}[!h]
	\includegraphics[width=\linewidth]{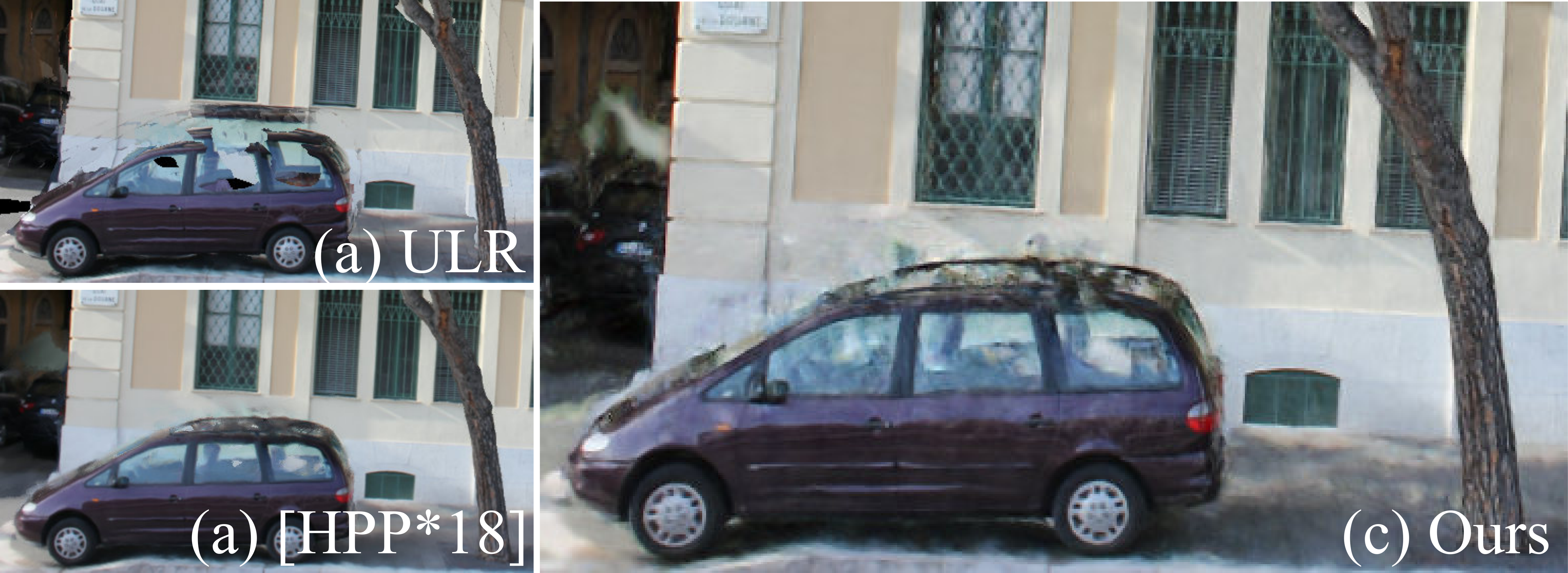}
	\caption{
		\label{fig:limitations}
The car roof is not reconstructed; our method cannot completely recover. Previous methods have similar difficulties.
	\vspace{-0.1cm}
	}
\end{figure}

\subsection{Limitations}

While our method outperforms previous solutions in the vast majority of examples we show, it is not without limitations. In some scenes  if the model overfits there is some slightly visible temporal flickering; This is visible for the Hugo scene for 40K iterations (please see supplemental \CR{website}). However, for a smaller number of iterations (20K iterations), the overall result of rendering is comparable, and the flickering  is reduced significantly. 

We currently optimize normals and depth separately; it might be beneficial to enforce a consistency loss between them, which could improve geometry optimization overall. 
Our treatment of view-dependent effects is improved using the texture stretch weight $w_{TS}$ (Sec.~\ref{sec:nr}). However, to correctly render view-dependent effects, reflection flow needs to be modeled (\cite{RPHD20}); NeRF achieves this to a certain extent in the depth/appearance optimization, but at a significant cost in training and rendering. Correctly solving this problem is an exciting future challenge.

Even though we improve over methods that cannot recover from MVS errors, if the initial reconstruction is too erroneous, our optimization cannot always recover and fails similarly to other methods. This includes cases, when a large piece of geometry is missing (Fig.~\ref{fig:limitations}), but also more complex cases such as reflections and transparency that require multiple depths. \CR{Complete failure of MVS reconstruction can also lead to temporal instability for our method e.g., "Tree" scene in the supplemental website.} These are all interesting avenues for future work; we believe our framework provides the initial tools for the development of such solutions.

\section{Conclusion}

We have presented a new differentiable point-based pipeline that enables per-input-view optimization. To achieve the rendering quality required, we introduce a differentiable splatting pipeline allowing per-view optimization, with probabilistic depth testing between input views and efficient camera selection. These components allow us to define a temporally consistent neural renderer.

This powerful pipeline allows optimization of different properties in the input views of a multi-view dataset. We present three examples: first IBR, where we optimize depth, color and reprojected features, clearly improving neural rendering quality in regions such as vegetation; second multi-view harmonization, where we are able to improve exposure related inconsistencies in multi-view datasets and finally multi-view stylization, where we achieve multi-view consistency in stylization.

Our results show that we provide a solution to our initial goal, i.e., to be able to benefit both from the rich, stable 3D information provided by multi-view stereo, while allowing neural optimization in the input views which improves novel-view synthesis. 
We believe that our generic pipeline can be applied to further improve neural rendering in the future. In particular, we are interested in investigating ways to recover from very large reconstruction errors, which, based on previous work~\cite{chaurasia2013depth,HRDB16,hedman2018deep} should potentially be easier in the space of input views. One long standing problem of IBR is the treatment of reflections and transparency~\cite{sinha2012image,RPHD20}. Our framework could be extended to handle the two depths required to treat these problems in a principled manner. Another application area could be multi-view matting.

\section{Acknowledgments}
This research was funded by the ERC Advanced grant FUNGRAPH No 788065 (\url{http://fungraph.inria.fr}). The authors are grateful to the OPAL infrastructure from Universit\'e C\^ote d'Azur for providing resources and support. The authors thank G. Riegler for help with comparisons. Thanks to A. Bousseau for proofreading earlier drafts, E. Yu for help with the figures and S.Diolatzis for thoughtful discussions throughout. Finally, the authors thank the anonymous reviewers for their valuable feedback.

\bibliographystyle{eg-alpha-doi}
\bibliography{ms.bib}
\end{document}


\title{Supplemental Material: Point-Based Neural Rendering with Per-View Optimization}
\maketitle

\section{Derivation of probability-based depth}
\label{sec:app1}

Following from the discussion in Sec. 3.3 in the main paper, we assume $D_n$ follows distribution $\df$ based on a mixture model with density $\df_{D_n}(d)$. We first assume that there exists at least one such point (thus the symbol $\exists$):
	\[ \df_{D_n}^{\exists}(d) = \sum_{i \in \mathcal{N}_n} f(d,d_{n,i})\left(\alpha_{n,i}\prod_{j=1}^{i-1}(1-\alpha_{n,j})\right)\]
\begin{equation}
	\label{eq:df}
	\df_{D_n}^{\exists}(d) = \sum_{i \in \mathcal{N}_n} f(d,d_{n,i})\beta_{n,i}
\end{equation}
\noindent
	With:
	\[\beta_{n,i}=\alpha_{n,i}\prod_{j=1}^{i-1}(1-\alpha_{n,j})\]
\noindent
where $f(d,d_{n,i},\sigma_{ni,})$ is the $\pdf$ of a distribution that describes the likelihood of point $\rho_{n,i}$  being at depth $d$. This can be an arbitrary single mode distribution such as a uniform or normal distribution.
\noindent
The distribution of Eq.~\ref{eq:df} does not integrate to one, because there is also a probability that no points exist at that pixel for that view:
	\[p_{D_n}^\emptyset = \prod_{i \in \mathcal{N}_n}(1-\alpha_{n,i}) = \beta_{n,\infty}\]
	We thus complete $\df_{D_n}^{\exists}(d)$ with a point at infinity to make it a probability density function. In order to simplify the remaining derivations we keep the notation $\mathcal{N}_n$ for the set of splats from input view $n$ \emph{completed} with that point at infinity, and use the corresponding $\beta_{n,\infty}$ for this point, leading to:
	\[ \df_{D_n}(d) = \sum_{i \in \mathcal{N}_n} f_i(d,d_{n,i})\beta_{n,i}\]
Note that implementation-wise adding the point at infinity is important to avoid corner cases.
We can then compute the probability that the (projected) depth of a view $n$ is smaller than all of the other views:
\begin{align*} 
P(D_n<\min_{m\neq n}(D_m))     		   & = \nonumber \\ 
\label{eq:smaller}
			\int_{-\infty}^{+\infty}   & P(t<\min_{m\neq n}(D_m)| D_n=t)\;\df_{D_n}(t)\;dt 
\end{align*}

\noindent
We have:
	\begin{align*} P(D_n<\min_{m\neq n}(D_m)) & = \nonumber \\ \int_{-\infty}^{+\infty}& P(t<\min_{m\neq n}(D_m)| D_n=t)\;\df_{D_n}(t)\;dt \\
	P(D_n<\min_{m\neq n}(D_m)) & = \nonumber \\ 
	\int_{-\infty}^{+\infty}& \prod_{m\neq n}P(t<D_m)\;\df_{D_n}(t)\;dt\\
	P(D_n<\min_{m\neq n}(D_m)) & = \nonumber \\ \sum_{i \in \mathcal{N}_n}\beta_{n,i} &\int_{-\infty}^{+\infty} \prod_{m\neq n}P(t<D_m)\;f_i(t,d_{n,i})\;dt
\end{align*}
\begin{align*}
	P(D_n<\min_{m\neq n}(D_m)) & = \nonumber \\ \sum_{i \in \mathcal{N}_n}\beta_{n,i}\int_{-\infty}^{+\infty} & \prod_{m\neq n}\left(\int_{t}^{\infty}\sum_{j \in \mathcal{N}_m} f_j(s,d_{m,j})\beta_{m,j}ds\right)\;f_i(t,d_{n,i})\;dt
\end{align*}
\begin{align*}
	P(D_n<\min_{m\neq n}(D_m)) & = \nonumber \\ \sum_{i \in \mathcal{N}_n}\beta_{n,i}\int_{-\infty}^{+\infty}& \prod_{m\neq n}\left(\sum_{j \in \mathcal{N}_m}\beta_{mj}\int_{t}^{\infty} f_j(s,d_{mj})ds\right)\;f_i(t,d_{n,i})\;dt
\end{align*}

\noindent
Thus:
\begin{align*}
	P(D_n<\min_{m\neq n}(D_m)) & = \nonumber \\ 
	\sum_{i \in \mathcal{N}_n}\beta_{n,i}\int_{-\infty}^{+\infty}& \prod_{m\neq n}\left(\sum_{j \in \mathcal{N}_m}\beta_{m,j}\int_{t}^{\infty} f_j(s,d_{m,j})ds\right)\;f_i(t,d_{n,i})\;dt
\end{align*}
A natural candidate for $f$ would be a normal distribution. Unfortunately this expression would be very costly to compute and does not have a closed form solution. Instead we use a simple symmetric triangle distribution of support $2\sigma$, which is easy to evaluate. Example plots of these distributions are presented in Fig.~\ref{fig:perview_depth_pdfs}. This gives $f_i$:
	\[ f_i(s,d_{n,i})ds =
	\begin{cases}
		\frac{\sigma-|t-x|}{\sigma^2} &\textrm{   if    } d_{n,i}-\sigma<t\leq d_{n,i}+\sigma \textrm{,}\\
		0 &\textrm{   otherwise}
	\end{cases}
	\]
And:
		\[ \int_{t}^{\infty}f_i(s,d_{n,i})ds = T(t,d_{n,i},\sigma)=
	\begin{cases}
		1 &\textrm{   if } t<d_{n,i}-\sigma \textrm{,}\\
		0 &\textrm{   if } t>d_{n,i}+\sigma \textrm{,}\\
		1-\frac{(x-t+\sigma)^2}{2\sigma^2} &\textrm{   if } d_{n,i}-\sigma<t\leq d_{n,i} \textrm{,}\\
		\frac{(t+\sigma-x)^2}{2\sigma^2} &\textrm{   if } d_{n,i}<t\leq d_{n,i}+\sigma \textrm{,}\\

	\end{cases}
	\]
	\begin{align*} 
		P(D_n<\min_{m\neq n}(D_m)) &= \nonumber \\ 
			\sum_{i \in \mathcal{N}_n}\beta_{n,i}\int_{d_{n,i}-\sigma}^{d_{n,i}+\sigma}& \prod_{m\neq n}\left(\sum_{j \in \mathcal{N}_m}\beta_{m,j} T(t,d_{m,j},\sigma)\right)\;f_i(t,d_{n,i})\;dt
	\end{align*}
This functional still has no simple closed form formula, so we approximate it with numerical integration:
	\begin{align} 
P(D_n<\min_{m\neq n}(D_m)) &\approx \nonumber \\ \frac{2\sigma}{S} \sum_{i \in \mathcal{N}_n}\beta_{n,i}\sum_{t=1}^{S}& \prod_{m\neq n}\left(\sum_{j \in \mathcal{N}_m}\beta_{m,j} T(s(t),d_{m,j},\sigma)\right)\;f_i(s(t),d_{n,i})\;
\label{eq:soft-depth}
	\end{align}
	With:
	\[s(t) = d_{ni}-\sigma+\frac{t}{S+1} \textrm{, and $S$ the number of samples} \]
In our experiments, we found that setting $S=1$ provided satisfactory results in all our tests. This computation is done in parallel per pixel using CUDA.

\begin{figure}[!h]
	\includegraphics[width=\linewidth]{fig/perview_depth_pdfs.pdf}
	\caption{
		\label{fig:perview_depth_pdfs}
Simplified example of two $\pdf$s of a pixel for view $I_0$ and $I_1$. Our soft depth test is the computation of the probability of the random variable $D_i$ drawn by its corresponding PDF, $\df_{D_i}$ shown above, to have a smaller depth than all other $D_j$.
	}
\end{figure}

\section{Multi-view Harmonization details}

In Fig.~\ref{fig:optimization_exp_coefs}, we illustrate the evolution of the optimization of $\mu_i$.

\begin{figure}[!h]
	\includegraphics[width=\linewidth]{fig/optimization_exp_coefs.pdf}
	\caption{
		\label{fig:optimization_exp_coefs}
		The optimization of the $\mu_i$ for each view for 10k iterations during training of "Ponche" scene. Each line with a different color corresponds to a different input view; we clearly see a group of three images that are brighter (see Fig. 7 in the main paper).
	}
\end{figure}

\section{Additional Results and Comparisons}

\begin{figure}[!h]
	\includegraphics[width=.29\linewidth]{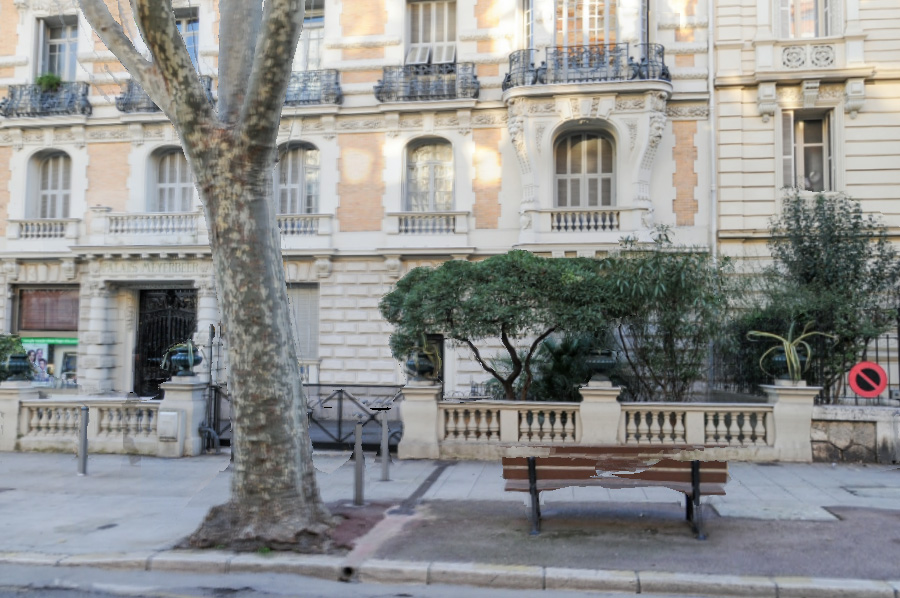}
	\includegraphics[width=.29\linewidth]{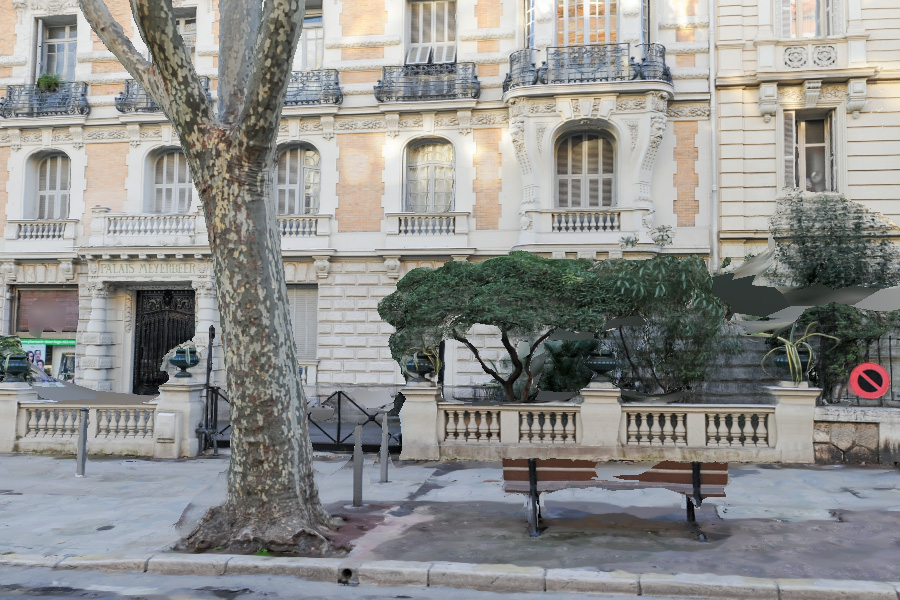} 
	\includegraphics[width=.29\linewidth]{images/results/hugo/ours/00000643.jpg}
	
	\includegraphics[width=.29\linewidth]{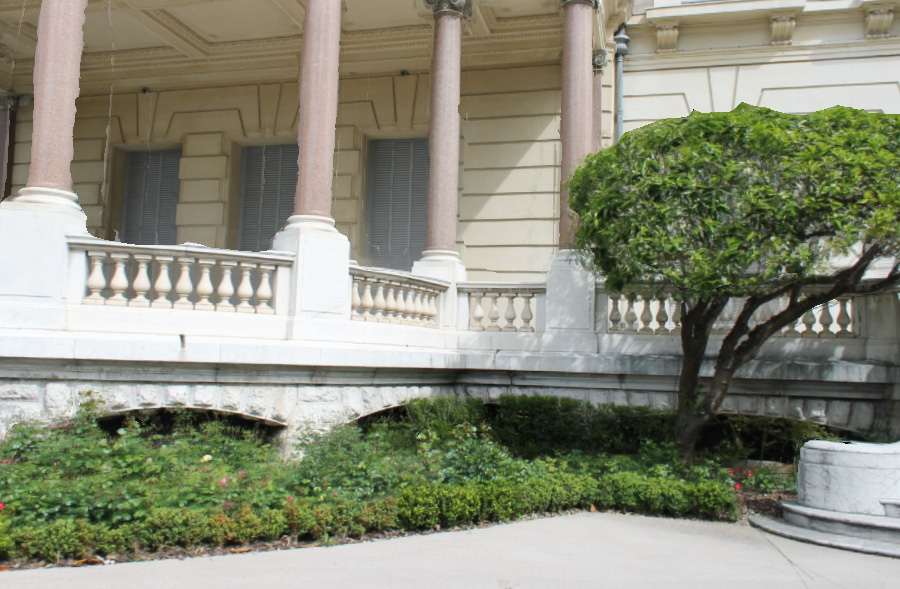}
	\includegraphics[width=.29\linewidth]{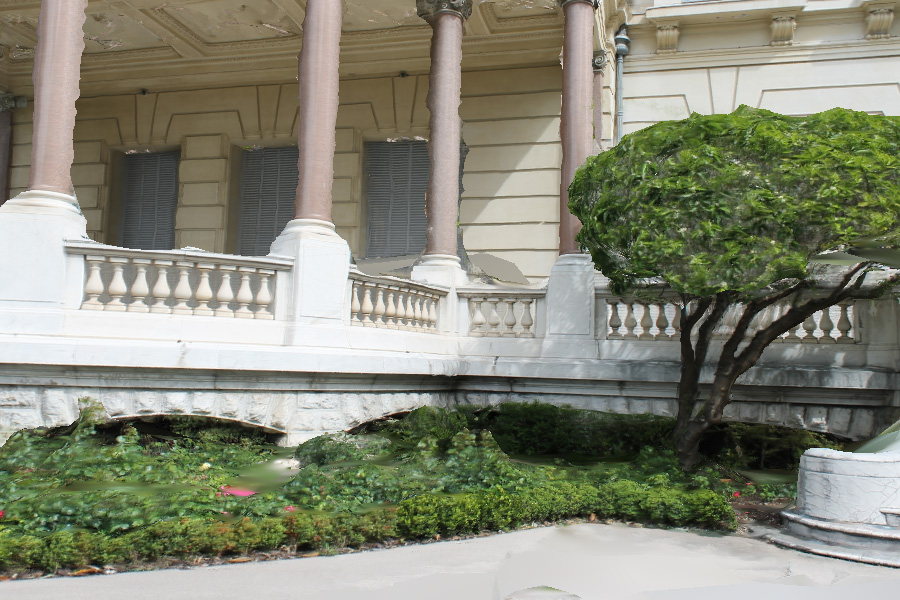} 
	\includegraphics[width=.29\linewidth]{images/results/museum/ours/00000230.jpg}
	
	\includegraphics[width=.29\linewidth]{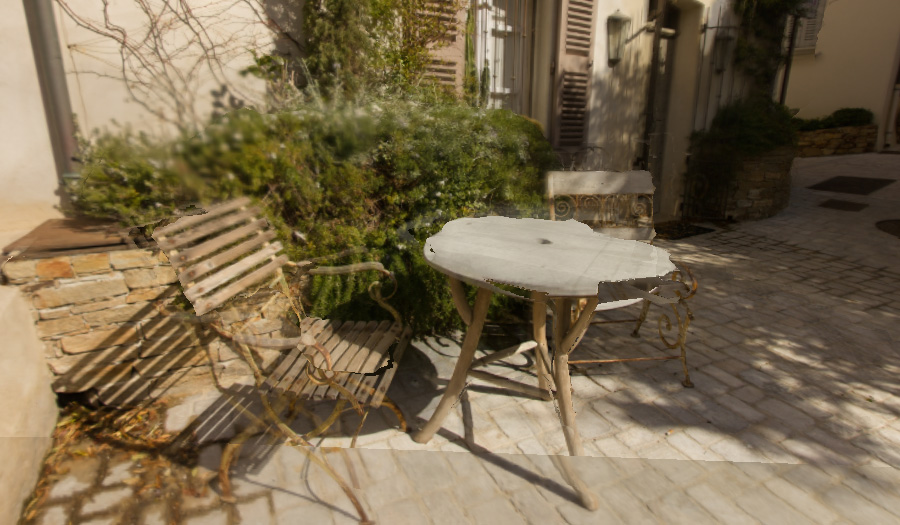}
	\includegraphics[width=.29\linewidth]{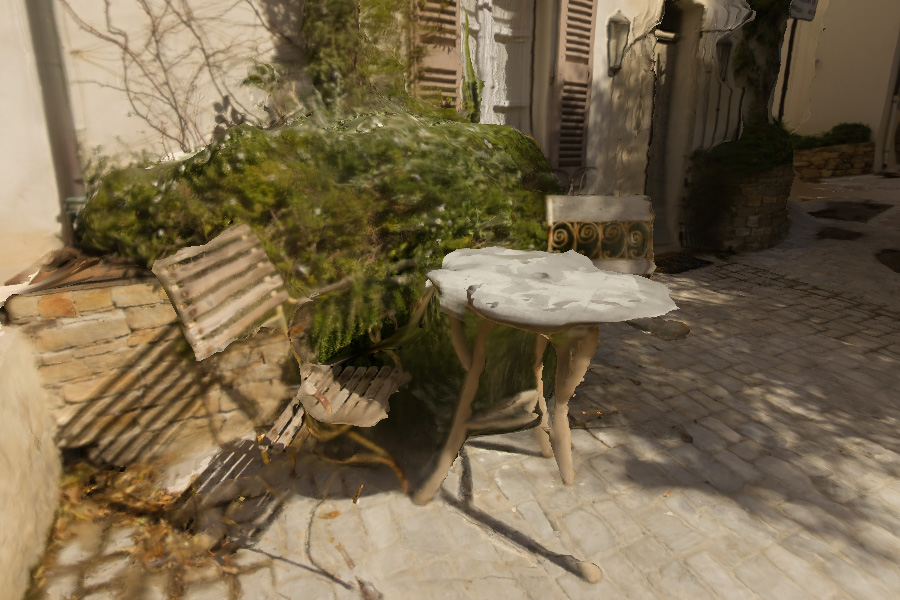} 
	\includegraphics[width=.29\linewidth]{images/results/ponche/ours/00000357.jpg}

	\includegraphics[width=.29\linewidth]{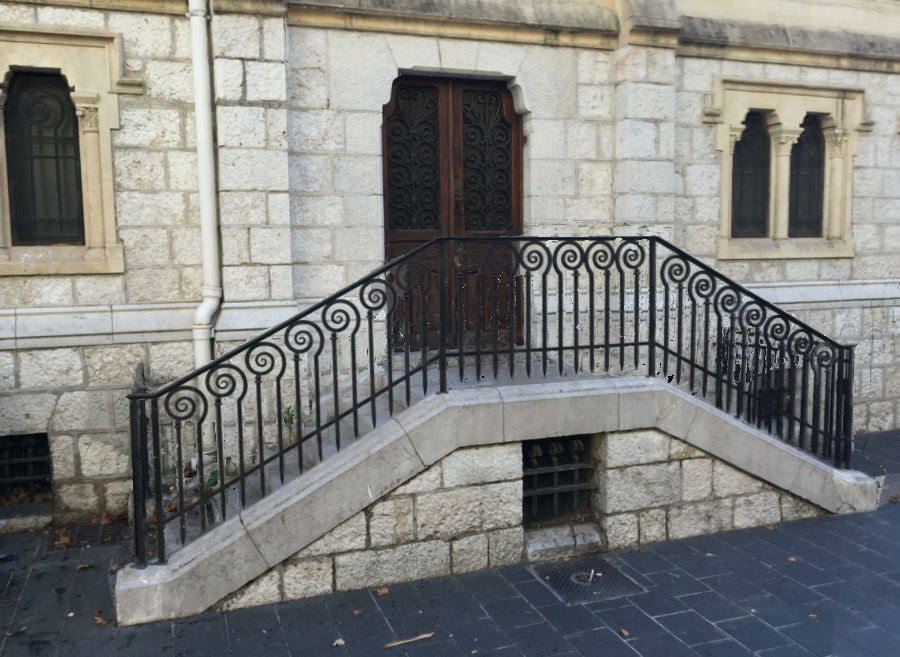}
	\includegraphics[width=.29\linewidth]{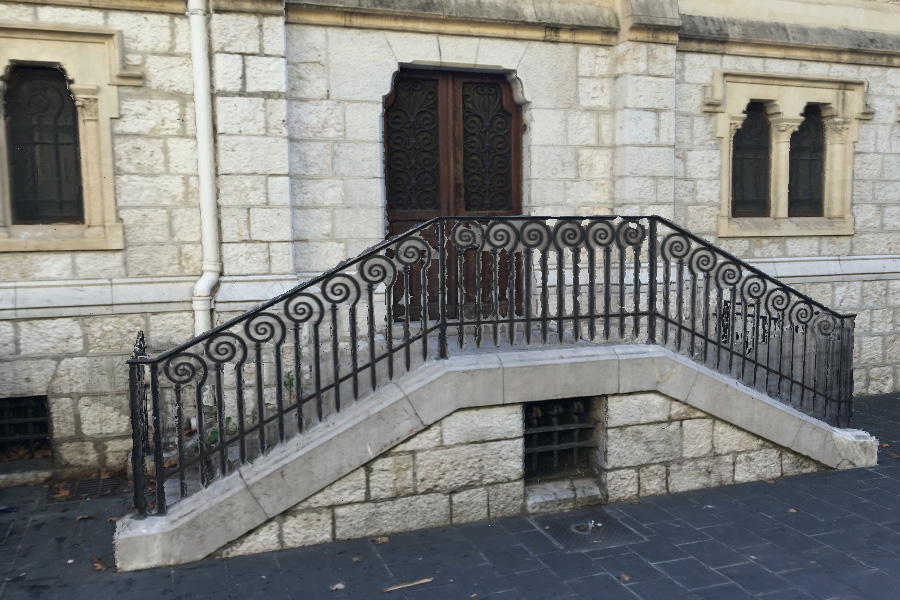} 
	\includegraphics[width=.29\linewidth]{images/results/stairs/ours/00000000.jpg}

	\includegraphics[width=.29\linewidth]{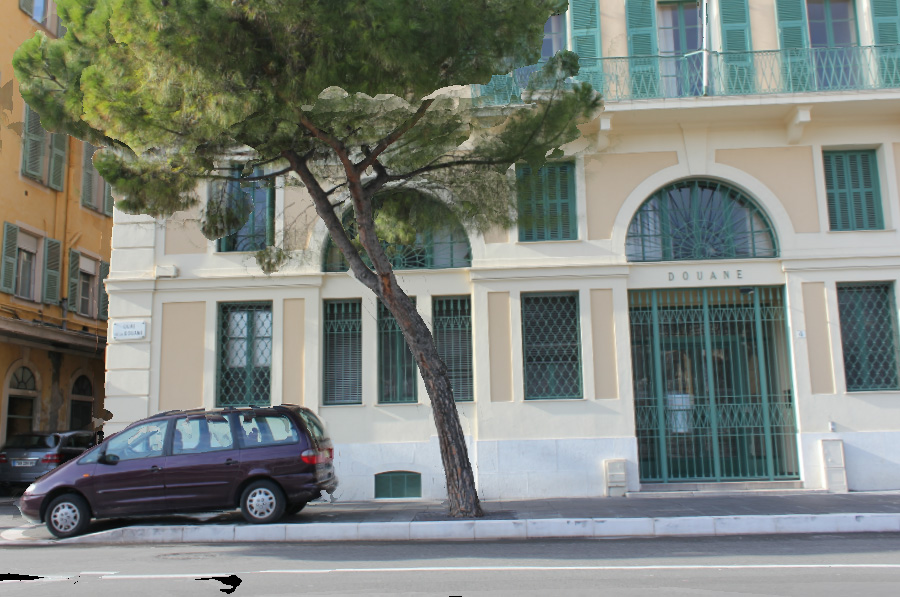}
	\includegraphics[width=.29\linewidth]{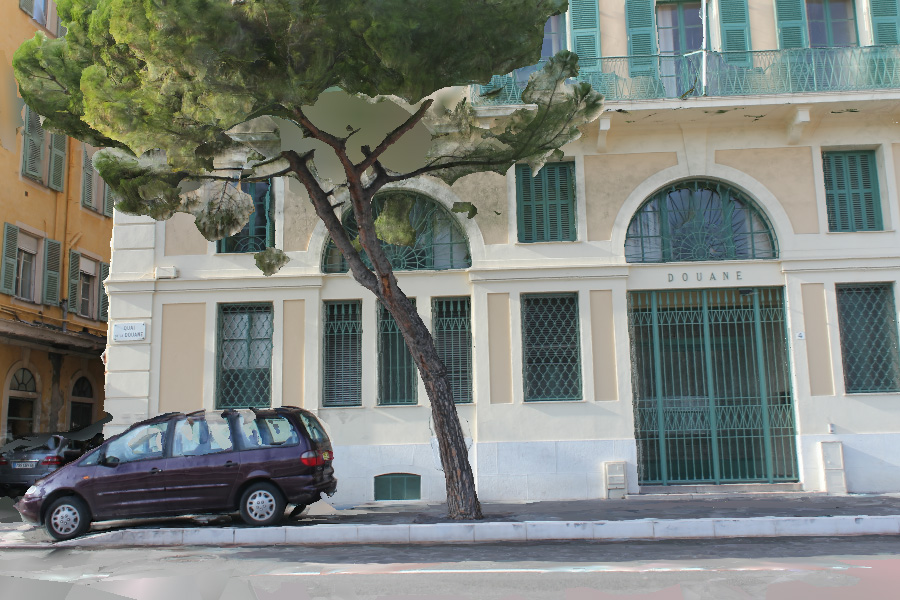} 
	\includegraphics[width=.29\linewidth]{images/results/tree/ours/00000143.jpg}
	
	\includegraphics[width=.29\linewidth]{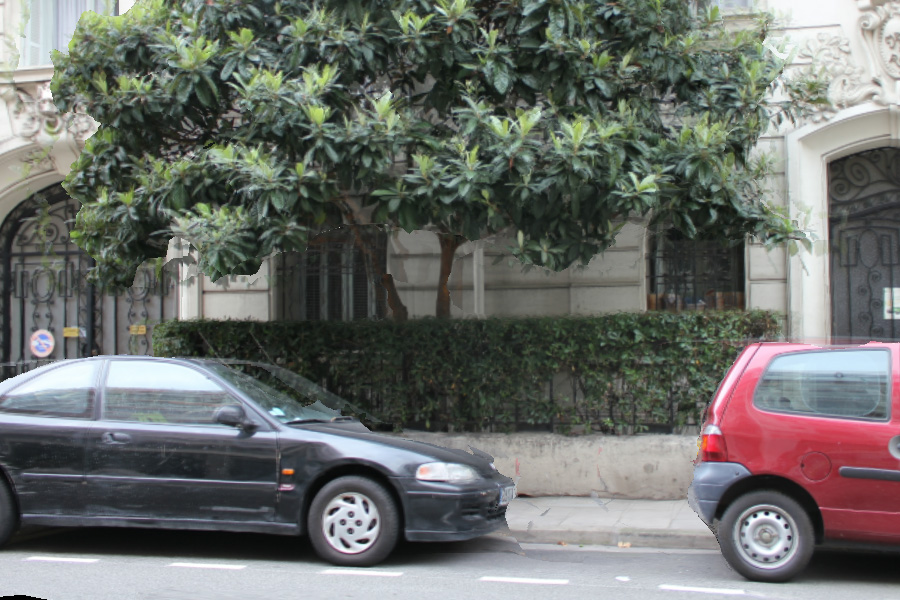}
	\includegraphics[width=.29\linewidth]{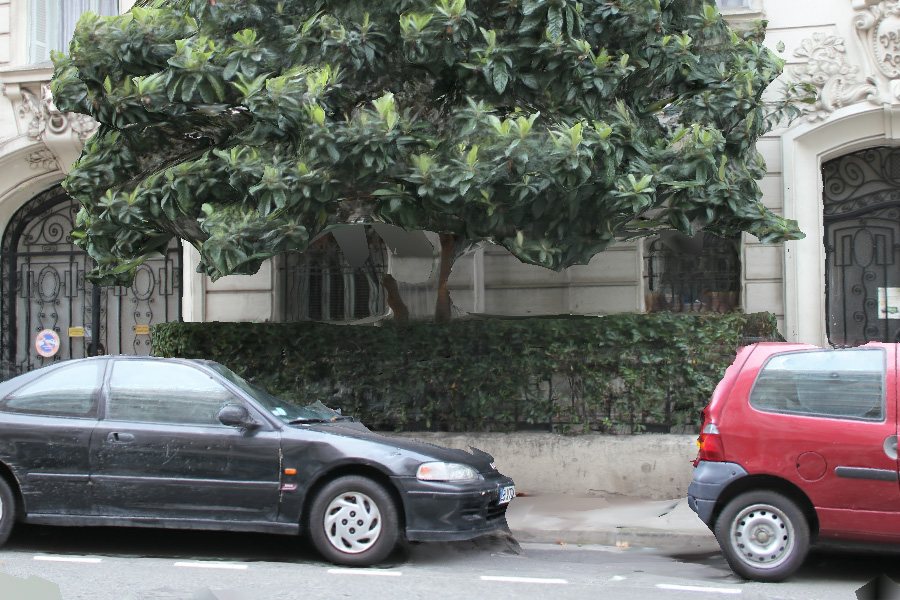} 
	\includegraphics[width=.29\linewidth]{images/results/street/ours/00000026.jpg}
	\caption{
		\label{fig:comparisons}
		Left to right: Unstructured Lumigraph Rendering, Textured Mesh, and our method.
	}
\end{figure}

In Fig.~\ref{fig:comparisons} we show comparisons of our method with two baseline methods, ULR and Textured Mesh.

In Fig.~\ref{fig:ablation_salon} we see that lowering the size of the model often does not affect the results substantially and higher number of cameras used for rendering in the same time/memory budget compensates for the smaller neural renderer. 

\begin{figure}[!h]
	\includegraphics[width=\linewidth]{fig/ablation_salon.pdf}
	\caption{
		\label{fig:ablation_salon}
In this challenging indoor scene, we use the small neural network to use a larger number of cameras.
	}
\end{figure}

\begin{table}[!ht]
	\small
	\centering
	\caption{\label{tab:mse_approximation} 
		\textbf{Quantitative analysis for the full rendering pipeline against the interactive approximation in the paper. For this experiment both methods use a smaller model to allow the latter for interactive frame-rates and isolate the impact of the point splatting approximation.}
	}
	\begin{tabular}{|l|c|}
		\toprule
		\multicolumn{1}{|c|}{}  & \multicolumn{1}{c|}{\bfseries MSE $\downarrow$} \\ \midrule
		Museum &    0.007 \\
		Ponche   &  0.004  \\
		Stairs   & 0.006    \\ 
		Street & 0.008  \\ \bottomrule
	\end{tabular}
\end{table}